\documentclass{article} 
\usepackage{iclr2025_conference,times}

\usepackage{wrapfig}
\usepackage{lipsum} 
\usepackage{wrapfig}

\usepackage{amsmath,amsfonts,bm}
\usepackage{multicol}

\def\eqref#1{equation~\ref{#1}}

\def\1{\bm{1}}

\DeclareMathAlphabet{\mathsfit}{\encodingdefault}{\sfdefault}{m}{sl}
\SetMathAlphabet{\mathsfit}{bold}{\encodingdefault}{\sfdefault}{bx}{n}

\usepackage{booktabs}      
\usepackage{multirow}      
\usepackage{amssymb}       

\usepackage{colortbl}
\usepackage{xcolor}

\usepackage{hyperref}
\usepackage{url}
\usepackage{multicol}
\usepackage{aliascnt}
\usepackage{adjustbox}
\usepackage{siunitx} 
\usepackage{booktabs}
\usepackage{multirow} 
\usepackage{enumitem}
\usepackage{booktabs} 
\usepackage{amssymb}  
\usepackage{amsmath}  
\usepackage{graphicx} 
\graphicspath{{./figs/}}
\usepackage{longtable} 
\usepackage{subcaption}
\usepackage{calc}
\usepackage[most]{tcolorbox}
\usepackage{listings}

\definecolor{uclablue}{rgb}{0.15, 0.45, 0.68}
\hypersetup{
    breaklinks,
    citecolor=uclablue,
    colorlinks=true,
}

\newtcolorbox{AIbox}[2][]{aibox,title=#2,#1}
\tcbset{
  aibox/.style={
    top=10pt,
    colframe=black,
    colbacktitle=black,
    coltitle=white,
    center,
  }
}

\lstdefinelanguage{prompt}{
    basicstyle=\scriptsize\ttfamily, 
    mathescape=true,        
    escapebegin=\color{latentcolor},  
    escapeend={},
    escapechar=@,
    stringstyle = \color{myorange},
    showstringspaces = false,
    moredelim = [s][\color{mypink}]{`}{`},
    moredelim = [s][\color{mybrown}]{```json}{```},
    moredelim = [s][\color{latentcolor}]{<StartOfLatent>}{<EndOfLatent>},
    literate = %
        {\ \ a.\ }{{\textcolor{mypurple}{\ \ a.\ }}}5
        {\ \ b.\ }{{\textcolor{mypurple}{\ \ b.\ }}}5
        {\ \ c.\ }{{\textcolor{mypurple}{\ \ c.\ }}}5
        {\ \ d.\ }{{\textcolor{mypurple}{\ \ d.\ }}}5
        {\ \ e.\ }{{\textcolor{mypurple}{\ \ e.\ }}}5
        {\ \ f.\ }{{\textcolor{mypurple}{\ \ f.\ }}}5
        {\ \ g.\ }{{\textcolor{mypurple}{\ \ g.\ }}}5
        {\ \ h.\ }{{\textcolor{mypurple}{\ \ h.\ }}}5
        {\ I.\ }{{\textcolor{mypurple}{\ I.\ }}}4
        {\ II.\ }{{\textcolor{mypurple}{\ II.\ }}}5
        {\ III.\ }{{\textcolor{mypurple}{\ III.\ }}}6
        {\ IV.\ }{{\textcolor{mypurple}{\ IV.\ }}}5
        {\ V.\ }{{\textcolor{mypurple}{\ V.\ }}}4
}

\newtcblisting[auto counter, number within=subsection]{prompt}[4][]{%
  before skip=6pt, after skip=6pt,    
  top=6pt, bottom=6pt, left=7pt, right=7pt, 
  enhanced, breakable,
  colback=white, colframe=gray!65,
  boxrule=0.6pt, arc=2pt,
  drop shadow={black!15},             
  listing only,
  listing options={
    language=prompt,
    upquote=true,
    basicstyle=\scriptsize\ttfamily \setlength{\baselineskip}{1.1\baselineskip},
    columns=fullflexible,
    breaklines=true,
    breakindent=0pt,
    xleftmargin=\ifx\empty#2\empty0pt\else#2\fi,
    xrightmargin=\ifx\empty#3\empty0pt\else#3\fi,
    aboveskip=0pt,                    
    belowskip=0pt,                    
    showstringspaces=false,
  },
  title={\ifstrempty{#4}{Prompt \thetcbcounter}{Prompt \thetcbcounter: #4}},
  colbacktitle=gray!6, coltitle=black,
  attach boxed title to top left={yshift=-1mm, xshift=6pt},
  boxed title style={colback=gray!6, boxrule=0pt, sharp corners},
  #1
}

\newtcblisting[auto counter, number within=subsection]{showcase}[4][]{%
  before=\par\vspace{\baselineskip},
  after=\par,
  width=\linewidth,
  enhanced,
  arc=0em,
  boxrule=1pt,
  listing only,
  listing options={
    language=prompt,
    upquote=true,
    basicstyle=\scriptsize\ttfamily \setlength{\baselineskip}{1.1\baselineskip},
    breaklines=true,
    breakindent=0pt,
    xleftmargin=\ifx\empty#2\empty-12pt\else#2\fi,
    xrightmargin=\ifx\empty#3\empty-5pt\else#3\fi,
    aboveskip=-4pt,
    belowskip=-4pt,
    columns=fullflexible,
    },
  colback=white,
  colframe=gray,
  colbacktitle=gray!5,
  coltitle=black,
  attach boxed title to top center={yshift=-3mm},
  box align=center,
  parbox=false,
  title={\ifstrempty{#4}{Example \thetcbcounter}{Example \thetcbcounter: #4}},
  #1
}

\definecolor{linkColor}{rgb}{0.2,0.4,0.6}
\definecolor{myblue}{HTML}{0379AC}
\definecolor{myred}{HTML}{A50E50}
\definecolor{myorange}{RGB}{238, 133, 74}
\definecolor{latentcolor}{named}{cyan}
\definecolor{normalcolor}{RGB}{0, 0, 0}
\usepackage{marvosym}

\usepackage{tabularx}
\usepackage{bbm}
\usepackage{makecell}

\usepackage{cleveref}
\newcommand{\minisection}[1]{\noindent{\textbf{#1}}.}

\usepackage{caption}
\title{Diversity or Precision? A Deep Dive into Next Token Prediction}

\author{
Haoyuan Wu$^{1, 2}$, Hai Wang$^{2, \dagger}$, Jiajia Wu$^{2}$, Jinxiang Ou$^{2}$, Keyao Wang$^{2}$,\\  
Weile Chen$^{2}$, Zihao Zheng$^{2}$,
Bei Yu$^{1}$ \\
\textbf{$^1$The Chinese University of Hong Kong} \quad
\textbf{$^2$LLM Department, Tencent} 
}

\begin{document}
\maketitle
\let\oldthefootnote\thefootnote
\let\thefootnote\relax\footnotetext{$^\dagger$ Project Lead.}
\let\thefootnote\oldthefootnote

\begin{abstract}

Recent advancements have shown that reinforcement learning (RL) can substantially improve the reasoning abilities of large language models (LLMs). 
The effectiveness of such RL training, however, depends critically on the exploration space defined by the pre-trained model's token-output distribution.
In this paper, we revisit the standard cross-entropy loss, interpreting it as a specific instance of policy gradient optimization applied within a single-step episode. 
To systematically study how the pre-trained distribution shapes the exploration potential for subsequent RL, we propose a generalized pre-training objective that adapts on-policy RL principles to supervised learning. 
By framing next-token prediction as a stochastic decision process, we introduce a reward-shaping strategy that explicitly balances diversity and precision. 
Our method employs a positive reward scaling factor to control probability concentration on ground-truth tokens and a rank-aware mechanism that treats high-ranking and low-ranking negative tokens asymmetrically. 
This allows us to reshape the pre-trained token-output distribution and investigate how to provide a more favorable exploration space for RL, ultimately enhancing end-to-end reasoning performance. 
Contrary to the intuition that higher distribution entropy facilitates effective exploration, we find that imposing a precision-oriented prior yields a superior exploration space for RL.

\end{abstract}
\section{Introduction}
Recent advancements have demonstrated that reinforcement learning (RL)~\citep{bai2022rlhf,guo2025deepseekr1} can significantly enhance the reasoning capabilities of large language models (LLMs)~\citep{deepmind2025gemini,guo2025deepseekr1,anthropic2025claude,kimi2025kimik2}.
By utilizing verifiable rewards, such as passing unit tests or deriving correct mathematical solutions, LLMs evolve from merely mimicking human data to actively searching for optimal reasoning paths~\citep{guo2025deepseekr1}.
On-policy training paradigms have proven effective in unlocking the potential of pre-trained LLMs, prompting researchers to investigate how token output distributions influence RL. 
Recent studies~\citep{wang2025beyond82rule,zhu2025negrl,cui2025entropyrl,gandhi2025cogbeh} indicate that uncertainty in chain-of-thought reasoning is concentrated within a small subset of high-entropy forking tokens that govern pivotal decisions, while the majority of tokens exhibit low entropy.  
This observation underscores the critical impact of the pre-trained model's output distribution on subsequent RL outcomes. 

Concurrently, researchers have explored next-token and next-segment reasoning objectives to derive self-supervised signals from massive unlabeled pre-training corpora~\citep{zelikman2024quietstar,dong2025rpt,li2025rlpt,xing2025pretrainzero}.
Applying RL to the pre-training corpus suggests a theoretical bridge connecting pre-training and RL.
Specifically, next-token prediction can be reformulated as a reasoning task optimized via RL algorithms, where the model receives verifiable rewards for accurately predicting the subsequent token according to a given context. 
Notably, if the intermediate reasoning process is omitted, resulting in the direct generation of the answer, this procedure becomes analogous to standard pre-training.
From the perspective of policy optimization, next-token prediction serves a foundational role by defining the initial policy distribution for subsequent RL. 
This distribution establishes the model’s behavioral trajectory and implicitly constrains its exploration space, thereby determining which reasoning paths the model prioritizes during RL.

Motivated by this connection, we revisit the cross-entropy loss for next token prediction. 
Although traditionally viewed as a supervised metric, cross-entropy can be interpreted as a specific instance of policy gradient optimization within a single-step episode~\citep{wu2025dft,ming2025otr}.
This interpretation suggests that next-token prediction inherently permits an on-policy perspective, even though standard teacher forcing utilizes off-policy samples drawn directly from the training corpus distribution. 
From an entropy perspective, cross-entropy implicitly assigns maximal reward to the single ground-truth token while uniformly suppressing all negative tokens. 
Building on this insight, we aim to establish a unified pre-training objective that subsumes cross-entropy as a special case, enabling a systematic study of how reward configurations during pre-training influence subsequent RL dynamics.

In this paper, we propose a generalized objective that integrates on-policy training principles into supervised learning. 
By formulating next-token prediction as a stochastic decision process, we expose the intrinsic reward mechanism of cross-entropy and introduce a reward-shaping strategy. 
This approach explicitly regulates the trade-off between diversity and precision during pre-training, rather than deferring this balance to subsequent RL stages. 
Specifically, we introduce a positive reward scaling factor to control the concentration of probability mass on ground-truth tokens, and we differentiate between high-ranking and low-ranking negative tokens to modulate suppression asymmetrically. 
This strategy allows us to reshape the token output distribution and systematically analyze the relationship between pre-training objectives and RL exploration. 
Contrary to the conventional intuition that higher distribution entropy facilitates effective exploration, our findings reveal that imposing a precision-oriented prior yields a superior exploration space for RL, ultimately enhancing end-to-end reasoning performance.

Our main contributions are summarized as follows:
\begin{itemize}[leftmargin=2em]
    \item We propose a generalized pre-training objective for next-token prediction that incorporates a reward-shaping strategy, utilizing a positive reward scaling factor and rank-aware negative suppression.
    \item We investigate how reshaping the token output distribution during pre-training modulates the exploration space for subsequent RL, thereby impacting end-to-end reasoning performance.
    \item We demonstrate that a precision-oriented pre-training prior provides a more effective initialization for RL than high-entropy distributions, leading to improved reasoning capabilities.
\end{itemize}
\section{Method}

\subsection{Next Token Prediction}

Autoregressive LLMs are typically trained using a next-token prediction objective. 
This process can be formulated as a sequential decision-making problem where the LLM functions as a stochastic policy $\pi_\theta$.

Let $X = \{x_1, x_2, \cdots, x_n\}$ denote a sequence of $n$ tokens. 
At step $t$, the state $s_t$ is defined by the prefix $X_{<t} = \{x_1, x_2, \cdots, x_{t-1}\}$. 
The action $a_t$ corresponds to the next token, sampled from the vocabulary $V$ according to the policy $\pi_\theta(\cdot \mid s_t)$. 
The training objective optimizes the parameters $\theta$ to maximize the expected cumulative reward:
\begin{equation}
J(\theta) = \mathbbm{E}_{\tau \sim \pi_\theta} \Big[\sum_{t=1}^{n} r(s_t, a_t)\Big],
\end{equation}
where $\tau =(s_1, a_1, s_2, a_2, \cdots)$ represents a trajectory sampled from $\pi_\theta$, and $r(s_t, a_t)$ is the scalar reward received for taking action $a_t$ in state $s_t$. 
The policy gradient can be derived as:
\begin{equation}
\nabla_\theta J(\theta)
= \mathbbm{E}_{\tau \sim \pi_\theta}\Big[\sum_{t=1}^{n} R(\tau) \nabla_\theta \log \pi_\theta(a_t \mid s_t)\Big],
\label{eq:pg}
\end{equation}
where $R(\tau)=\sum_{t'=1}^n r(s_{t'},a_{t'})$.
To reduce variance without introducing bias, the total return $R(\tau)$ is typically replaced by the return-to-go $G_t=\sum_{t' = t}^n r(s_{t'},a_{t'})$, often incorporating a baseline $b(s_t)$ for variance reduction:
\begin{equation}
\nabla_\theta J(\theta)
= \mathbbm{E}_{\tau \sim \pi_\theta}\Big[\sum_{t=1}^{n} (G_t - b(s_t)) \nabla_\theta \log \pi_\theta(a_t \mid s_t) \Big].
\label{eq:pg_unbias}
\end{equation}
Building upon \Cref{eq:pg_unbias}, we treat the generation of a single token as a complete episode~\citep{ming2025otr}. 
The objective for a fixed state $s_t$ simplifies to:
\begin{equation}
J_t(\theta \mid s_t) = \mathbbm{E}_{a_t \sim \pi_\theta(\cdot \mid s_t)}[ r(s_t, a_t)],
\label{eq:otr_obj}
\end{equation}
yielding the gradient:
\begin{equation}
\nabla_\theta J_t(\theta \mid s_t)
= \mathbbm{E}_{a_t \sim \pi_\theta(\cdot \mid s_t)}\big[r(s_t, a_t) \nabla_\theta \log \pi_\theta(a_t \mid s_t) \big].
\label{eq:otr_pg}
\end{equation}
Crucially, for \Cref{eq:otr_pg} to remain consistent with the cumulative reward structure of \Cref{eq:pg_unbias}, the reward $r(s_t, a_t)$ must depend solely on the immediate state-action pair.

\subsection{Revisiting Cross-Entropy}

LLM pre-training is generally cast as a supervised learning process designed to maximize the log-likelihood of the ground-truth token $x_t$ given the context $s_t = X_{<t}$:
\begin{equation}
    J_{\text{CE}}(\theta) = \log \pi_\theta(x_t \mid s_t).
\end{equation}
The gradient of this objective explicitly maximizes the probability of the ground-truth token:
\begin{equation}
    \nabla_\theta J_{\text{CE}}(\theta) = \nabla_\theta \log \pi_\theta(x_t \mid s_t).
\end{equation}
We can express this gradient as an expectation over the full policy distribution $\pi_\theta(\cdot \mid s_t)$, encompassing both positive ($a_t = x_t$) and negative ($a_t \neq x_t$) tokens. 
By invoking the log-derivative identity $\nabla_\theta \log \pi_\theta(x) = \frac{\nabla_\theta \pi_\theta(x)}{\pi_\theta(x)}$ and introducing the indicator function $\mathbbm{1}(a_t = x_t)$, we expand the gradient into a summation over the vocabulary $V$:
\begin{align}
    \nabla_\theta J_{\text{CE}}(\theta) 
    &= \frac{1}{\pi_\theta(x_t \mid s_t)} \nabla_\theta \pi_\theta(x_t \mid s_t) \nonumber \\
    &= \frac{1}{\pi_\theta(x_t \mid s_t)} \sum_{a_t \in V} \mathbbm{1}(a_t = x_t) \nabla_\theta \pi_\theta(a_t \mid s_t).
\end{align}
We recover the probability density using the substitution $\nabla_\theta \pi_\theta(a_t \mid s_t) = \pi_\theta(a_t \mid s_t) \nabla_\theta \log \pi_\theta(a_t \mid s_t)$, and then form an expectation:
\begin{align}
    \nabla_\theta J_{\text{CE}}(\theta) 
    &= \sum_{a_t \in V} \pi_\theta(a_t \mid s_t) \left[\frac{\mathbbm{1}(a_t = x_t)}{\pi_\theta(a_t \mid s_t)} \nabla_\theta \log \pi_\theta(a_t \mid s_t) \right] \nonumber \\
    &= \mathbbm{E}_{a_t \sim \pi_\theta(\cdot \mid s_t)} \left[r_{\text{CE}}(s_t, a_t) \nabla_\theta \log \pi_\theta(a_t \mid s_t) \right].
    \label{eq:ce_as_pg}
\end{align}
In supervised training, the ground-truth token $x_t$ is deterministically defined by the dataset. 
Consequently, the indicator $\mathbbm{1}(a_t = x_t)$ evaluates the action $a_t$ against a static property of $s_t$, ensuring that the derived intrinsic reward depends exclusively on information available at step $t$. 
Comparing \Cref{eq:ce_as_pg} with \Cref{eq:otr_pg} reveals the intrinsic reward function of cross-entropy:
\begin{equation}
    r_{\text{CE}}(s_t, a_t) = \text{sg}(\frac{\mathbbm{1}(a_t = x_t)}{\pi_\theta(a_t \mid s_t)}), \label{eq:ce_reward}
\end{equation}
where $\text{sg}(\cdot)$ denotes the stop-gradient operator.
\Cref{eq:ce_reward} demonstrates that when the sampled action matches the ground truth ($a_t = x_t$), the reward is scaled by the inverse probability $\frac{1}{\pi_\theta(x_t \mid s_t)}$.
On the contrary, for all negative tokens, the intrinsic reward is exactly 0. 
Unlike RL scenarios where negative actions are often explicitly penalized, cross-entropy achieves suppression of negative tokens implicitly through the Softmax normalization constraint $\sum\limits_{a_t \in V} \pi_\theta(a_t \mid s_t) = 1$.
By increasing the probability of the positive tokens via positive rewards, the probabilities of competing tokens are forced to decrease.

\subsection{Diversity or Precision}

As derived in \Cref{eq:ce_reward}, the intrinsic reward of the cross-entropy objective implicitly balances diversity and precision. 
To explicitly regulate the trade-off between these two objectives, we propose a generalized reward function designed to independently control the influence of positive and negative tokens.

First, we introduce a modulating factor to scale the reward associated with the ground-truth token.
Let $a_t$ denote the generated token and $x_t$ the ground truth, we define the modified positive reward as:
\begin{equation}
    \bar{r}_{\text{pos}}(s_t, a_t) = \text{sg}((\frac{1}{\pi_\theta(a_t \mid s_t)})^{(1-\pi_\theta(a_t \mid s_t))^{\beta}}),
    \label{eq:pos_reward}
\end{equation}
where $(1-\pi_\theta(a_t \mid s_t))^{\beta}$ serves as a positive reward scaling factor. 
\Cref{eq:pos_reward} facilitates the control of global entropy. 
Specifically, when $\beta < 0$, the reward is amplified relative to the baseline ($\beta = 0$).
This produces large gradient updates that aggressively concentrate probability mass onto the ground truth, collapsing the distribution and minimizing global entropy. 
Conversely, $\beta > 0$ attenuates the reward signal.
In this regime, the model is less penalized for assigning a lower probability to the ground truth, allowing the policy to maintain a flatter distribution with higher entropy.

Second, while standard cross-entropy assigns zero reward to all negative tokens, we propose shaping the negative distribution to control local entropy. 
Let $\mathcal{K}_t = \text{TopK}(\pi_\theta(\cdot \mid s_t), k)$ denote the set of the top-$k$ predicted tokens, we define the negative reward as:
\begin{align}
    \bar{r}_{\text{neg}}(s_t, a_t) = &\tilde{\lambda} \cdot \mathbbm{1}(a_t \in \mathcal{K}_t \land a_t \neq x_t) + \hat{\lambda} \cdot \mathbbm{1}(a_t \notin \mathcal{K}_t \land a_t \neq x_t).
    \label{eq:neg_reward}	
\end{align}
As shown in \Cref{eq:neg_reward}, we assign a reward $\tilde{\lambda}$ to high-ranking negative tokens to prevent the model from becoming overly confident in the ground truth alone, thereby reserving probability mass for plausible alternatives. 
Meanwhile, to suppress low-probability tail tokens, we apply a reward $\hat{\lambda}$ to tokens falling outside $\mathcal{K}_t$, forcing the distribution to concentrate on the head.

Finally, the generalized reward function for the single-step objective is defined as:
\begin{align}
    \bar{r}(s_t, a_t) &= \bar{r}_{\text{pos}}(s_t, a_t) \cdot \mathbbm{1}(a_t = x_t) + \bar{r}_{\text{neg}}(s_t, a_t) \cdot \mathbbm{1}(a_t \neq x_t).
    \label{eq:dorp_reward}
\end{align}
Notably, the setting $\beta = 0, \tilde{\lambda} = 0, \hat{\lambda} = 0$ recovers standard cross-entropy.
\section{Experiments}

\subsection{Training Settings}

The training pipeline proceeds in three stages: pre-training, mid-training, and RLVR.
Adhering to the Qwen3~\citep{yang2025qwen3}, we develop LLMs using both dense and MoE architectures. 
Specifically, we develop a series of LLMs, which include 1B and 4B dense models, as well as 5B-A0.3B, 10B-A0.5B and 20B-A1B MoE models.
Moreover, we conduct the complete training pipeline on the 4B, 10B-A0.5B and 20B-A1B models, while the 1B and 5B-A0.3B models undergo the pre-training stage only.
More training details are provided in \Cref{appendix:pretrain} and \Cref{appendix:rl}.

\minisection{Training Data}
For pre-training, we curate a corpus of 500B tokens primarily focused on general knowledge. 
This is followed by a mid-training stage comprising 100B tokens, which incorporates approximately 5\% synthetic data and significantly increases the proportion of reasoning-oriented content. 
Crucially, we deliberately exclude the synthetic long-reasoning data from all training stages to accurately observe the activation trends of the model's long-CoT reasoning capabilities.
The RL stage prioritizes mathematical reasoning tasks, as the emergence of long-reasoning capabilities is typically associated with these domains.

\minisection{Hyperparameters}
Hyperparameters are maintained across the pre-training and mid-training stages. 
Our goal is to investigate how different reward shaping strategies influence end-to-end performance. 
Consequently, we perform specific reward configurations for positive tokens ($\beta=-0.25$ and $\beta=0.5$) and negative tokens ($\hat{\lambda}=-0.1, \tilde{\lambda}=0, k=100$ and $\hat{\lambda}=0, \tilde{\lambda}=0.1, k=100$). 
Employing these distinct hyperparameter configurations allows us to isolate the specific effects of positive and negative reward signals.

\subsection{Evaluation Settings}

\minisection{Evaluation of Base Models}
Our comprehensive evaluation of base models assesses five core capabilities: general knowledge, logic reasoning, commonsense reasoning, mathematics, and coding. 
The evaluation is conducted using 19 distinct benchmarks:
\begin{itemize}[itemsep=0pt, topsep=0pt, parsep=0pt]
    \item \textbf{General Knowledge:} MMLU~\citep{hendrycks2020mmlu}(4-shot, CoT), MMLU-Pro~\citep{wang2024mmlupro}(5-shot, CoT), TriviaQA~\citep{joshi2017triviaqa}(5-shot), and NaturalQuestions~\citep{kwiatkowski2019naturalquestions}(5-shot).
    \item \textbf{Commonsense Reasoning:} Hellaswag~\citep{zellers2019hellaswag}(0-shot), SIQA~\citep{sap2019socialiqa}(0-shot), PIQA~\citep{bisk2020piqa}(0-shot), WinoGrande~\citep{sakaguchi2021winogrande}(0-shot), OpenBookQA~\citep{mihaylov2018openbookqa}(5-shot), and CommonsenseQA~\citep{talmor2018commonsenseqa}(5-shot)
    \item \textbf{Logic Reasoning:} ARC-Easy~\citep{clark2018arc}(0-shot), ARC-Challenge~\citep{clark2018arc}(0-shot), and BBH~\citep{suzgun2022bbh}(3-shot, CoT)
    \item \textbf{Mathematics:} GSM8K~\citep{cobbe2021gsm8k}(4-shot, CoT), MATH-500~\citep{lightman2023math500}(4-shot, CoT), Minerva~\citep{lewkowycz2022minerva}(4-shot, CoT), and OlympiadBench~\citep{he2024olympiadbench}(0-shot).
    \item \textbf{Coding:} HumanEval+~\citep{liu2023evalplus}(0-shot) and MBPP+~\citep{liu2023evalplus}(3-shot).
\end{itemize}
Specifically, general knowledge and commonsense reasoning evaluate the model’s knowledge-base capabilities, whereas logical reasoning, mathematics, and coding probe its reasoning-base capabilities.
Moreover, we employ the $\text{Pass@}k$ metric to evaluate the model's upper-bound capability for tasks requiring mathematical reasoning and code generation. 
Pass@k measures the probability that at least one correct solution is present within $k$ independent attempts.  
We utilize the unbiased estimator of $\text{Pass@}k$~\citep{chen2021passk}, which is defined as:
\begin{equation}
	\text{Pass@}k = 1 - \frac{\binom{m-c}{k}}{\binom{m}{k}},
\end{equation}
where $m$ represents the total number of sampled responses generated per prompt, and $c$ denotes the count of correct responses among those $m$ samples.
We sample $m = 128$ responses with temperature 0.7 and top-p 0.95 and report $\text{Pass@}$64 metric.
Notably, we configure the maximum output length to 4K for pre-trained models and 16K for mid-trained models.

\minisection{Evaluation of RL Models}
For RL models evaluation, we employ various mathematics benchmarks, including AMC23~\citep{maa2025amc}, AIME~\citep{maa2025aime}, MATH-500~\citep{lightman2023math500}, Minerva~\citep{lewkowycz2022minerva}, and OlympiadBench~\citep{he2024olympiadbench}. 
We sample 128 responses per problem and report $\text{Avg@}$128, $\text{Cons@}$128, and $\text{Pass@}$64 metrics.
Specifically, $\text{Avg@}128$ represents the average accuracy across all 128 samples, while $\text{Cons@}128$ refers to the majority voting accuracy. 
Similarly, we configure the maximum output length to 16K for RL models.

\begin{figure*}[!tb]
    \centering
    \begin{subfigure}{0.230\linewidth}
        \centering
        \includegraphics[width=\linewidth]{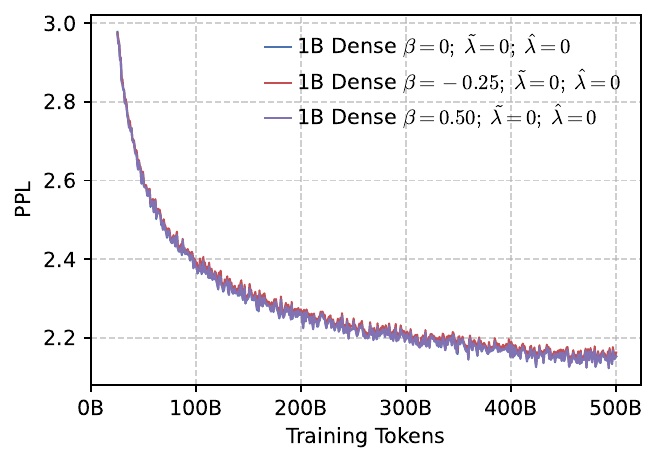}
    \end{subfigure}
    \hfill
    \begin{subfigure}{0.230\linewidth}
        \centering
        \includegraphics[width=\linewidth]{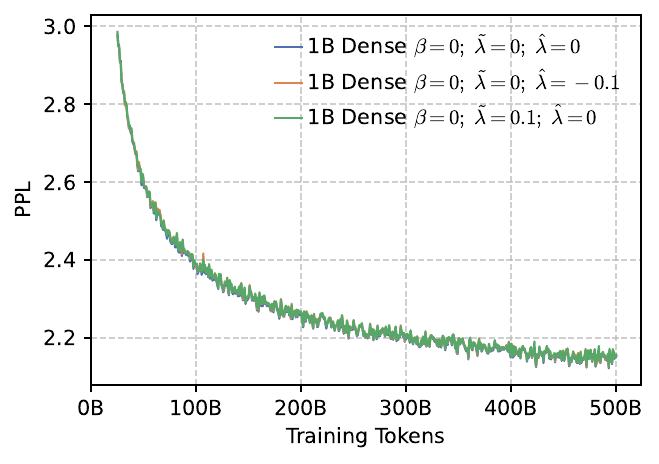}
    \end{subfigure}
    \hfill
    \begin{subfigure}{0.230\linewidth}
        \centering
        \includegraphics[width=\linewidth]{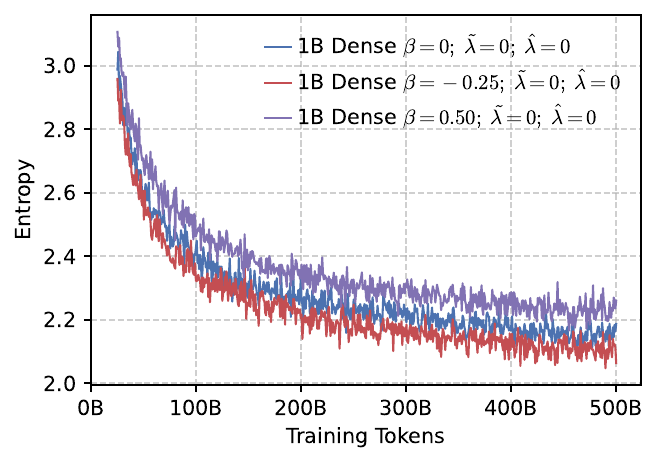}
    \end{subfigure}
    \hfill
    \begin{subfigure}{0.230\linewidth}
        \centering
        \includegraphics[width=\linewidth]{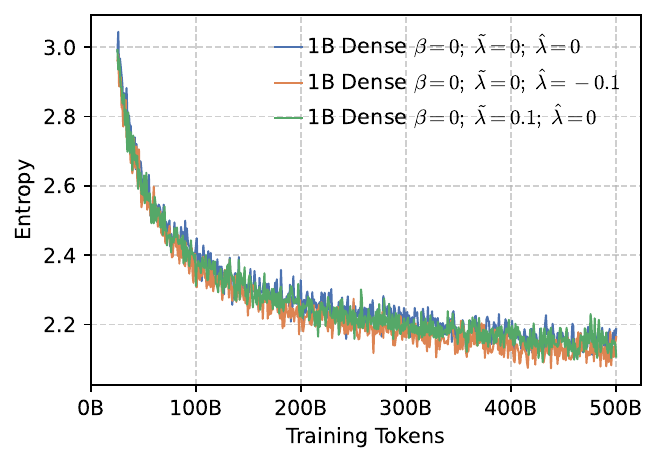}
    \end{subfigure}

    \begin{subfigure}{0.230\linewidth}
        \centering
        \includegraphics[width=\linewidth]{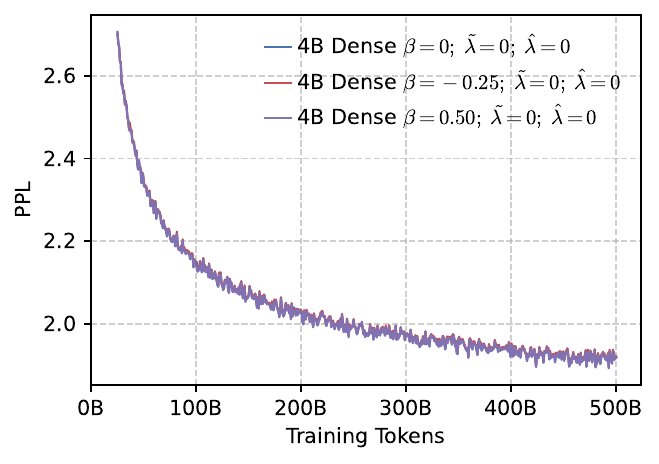}
    \end{subfigure}
    \hfill
    \begin{subfigure}{0.230\linewidth}
        \centering
        \includegraphics[width=\linewidth]{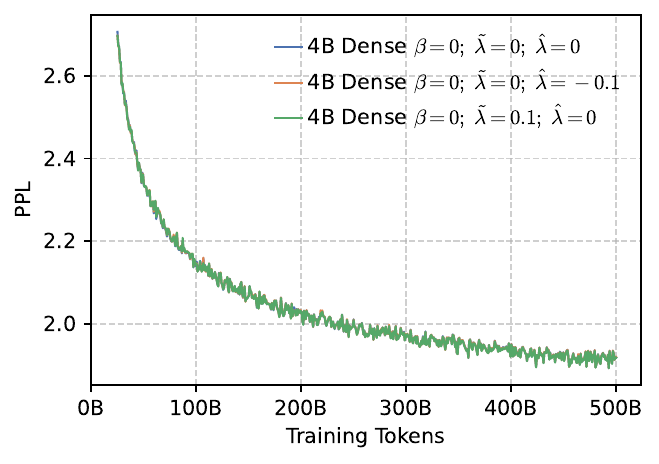}
    \end{subfigure}
    \hfill
    \begin{subfigure}{0.230\linewidth}
        \centering
        \includegraphics[width=\linewidth]{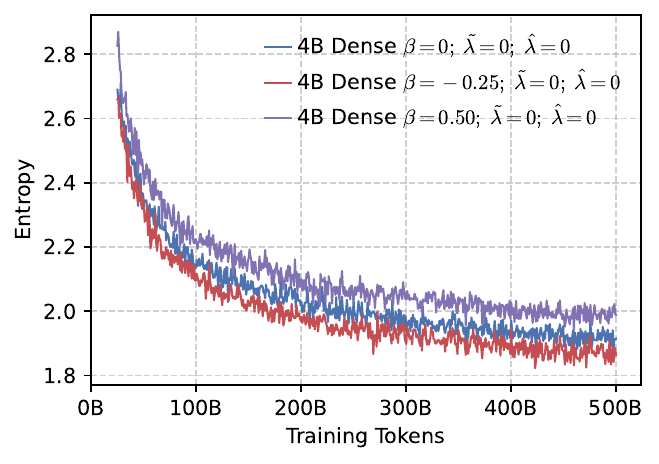}
    \end{subfigure}
    \hfill
    \begin{subfigure}{0.230\linewidth}
        \centering
        \includegraphics[width=\linewidth]{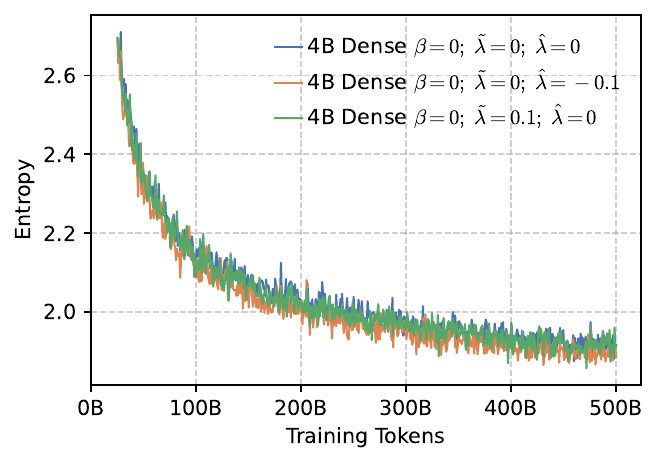}
    \end{subfigure}
    \caption{Changes of PPL and entropy during pre-training across 1B and 4B dense models, developed based on different configurations.}
    \label{fig:dense_pt_exp_supp}
\end{figure*}

\begin{figure*}[!tb]
    \centering
    \begin{subfigure}{0.230\linewidth}
        \centering
        \includegraphics[width=\linewidth]{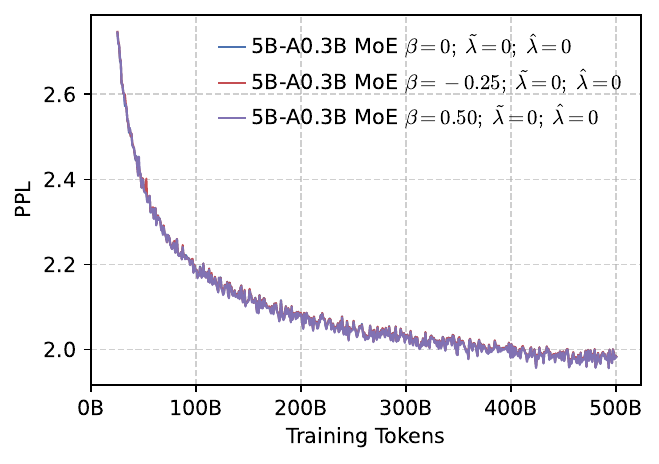}
    \end{subfigure}
    \hfill
    \begin{subfigure}{0.230\linewidth}
        \centering
        \includegraphics[width=\linewidth]{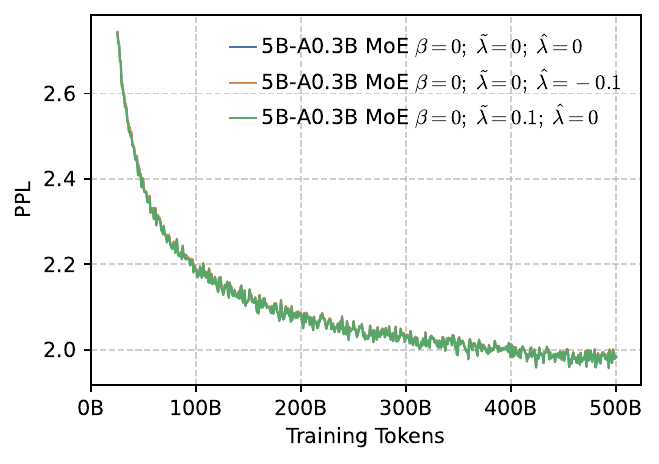}
    \end{subfigure}
    \hfill
    \begin{subfigure}{0.230\linewidth}
        \centering
        \includegraphics[width=\linewidth]{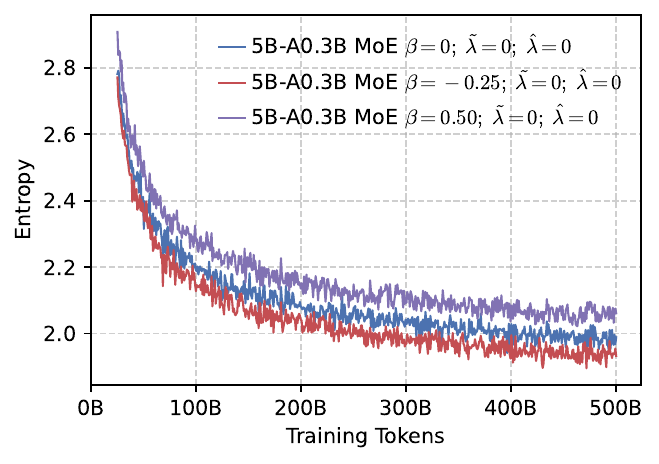}
    \end{subfigure}
    \hfill
    \begin{subfigure}{0.230\linewidth}
        \centering
        \includegraphics[width=\linewidth]{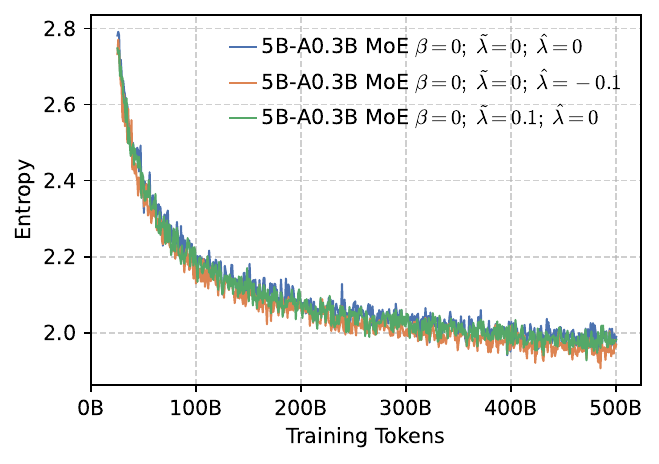}
    \end{subfigure}
    
    \begin{subfigure}{0.230\linewidth}
        \centering
        \includegraphics[width=\linewidth]{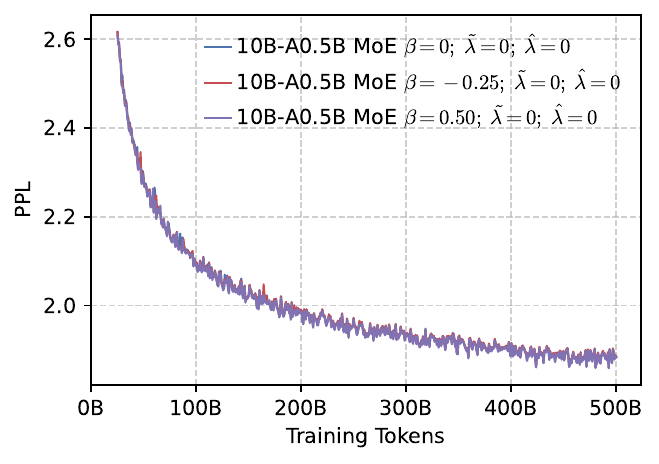}
    \end{subfigure}
    \hfill
    \begin{subfigure}{0.230\linewidth}
        \centering
        \includegraphics[width=\linewidth]{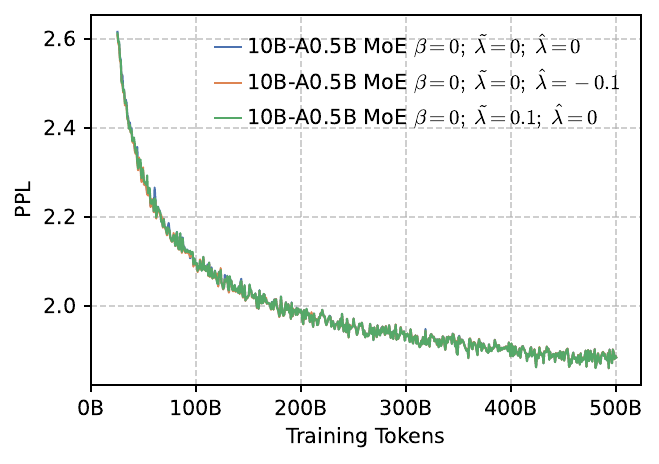}
    \end{subfigure}
    \hfill
    \begin{subfigure}{0.230\linewidth}
        \centering
        \includegraphics[width=\linewidth]{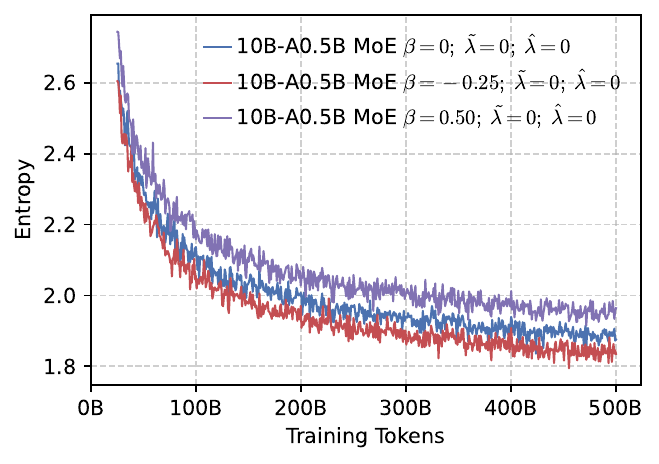}
    \end{subfigure}
    \hfill
    \begin{subfigure}{0.230\linewidth}
        \centering
        \includegraphics[width=\linewidth]{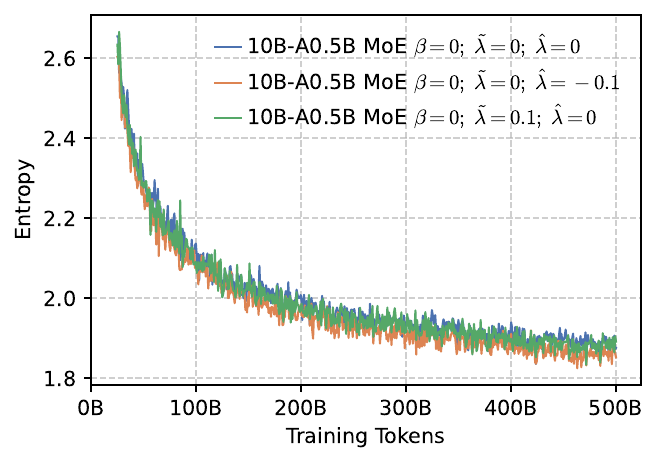}
    \end{subfigure}

    \begin{subfigure}{0.230\linewidth}
        \centering
        \includegraphics[width=\linewidth]{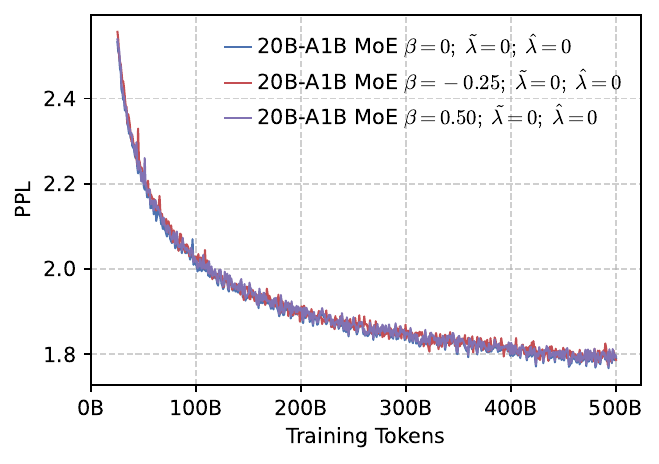}
    \end{subfigure}
    \hfill
    \begin{subfigure}{0.230\linewidth}
        \centering
        \includegraphics[width=\linewidth]{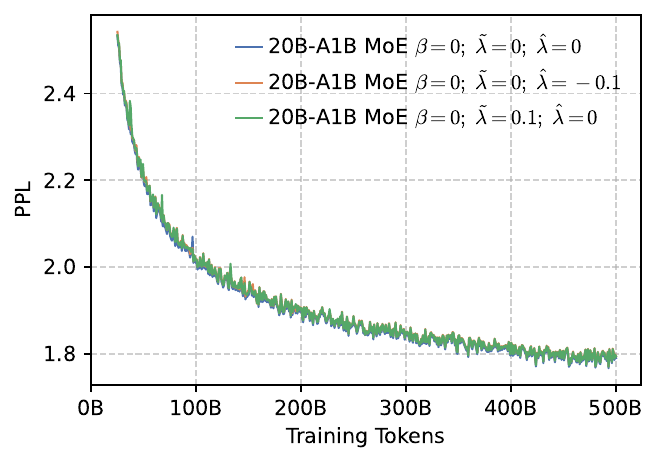}
    \end{subfigure}
    \hfill
    \begin{subfigure}{0.230\linewidth}
        \centering
        \includegraphics[width=\linewidth]{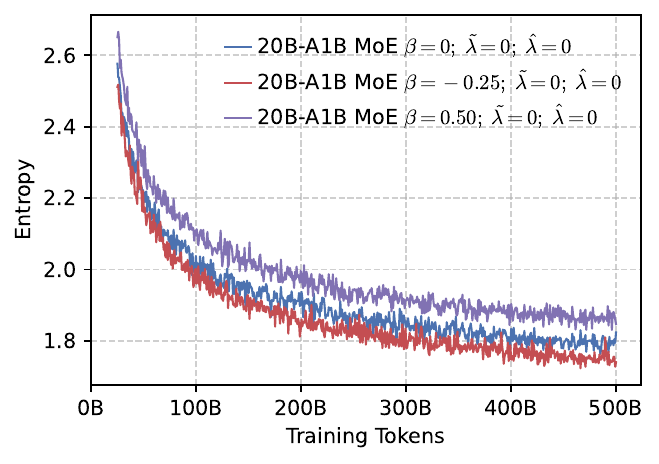}
    \end{subfigure}
    \hfill
    \begin{subfigure}{0.230\linewidth}
        \centering
        \includegraphics[width=\linewidth]{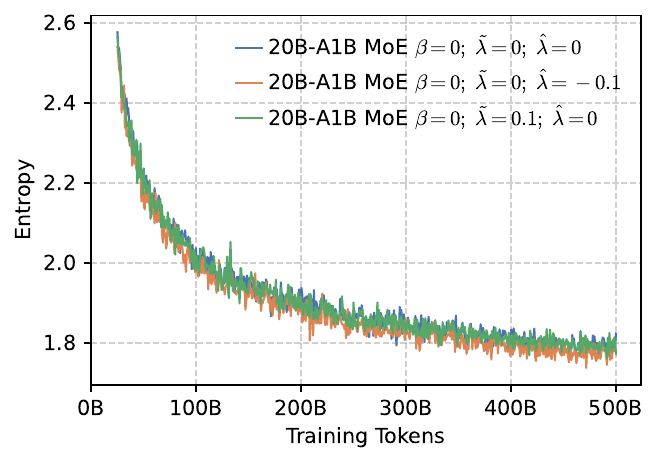}
    \end{subfigure}
    \caption{Changes of PPL and entropy during pre-training across 5B-A0.3B, 10B-A0.5B  and 20B-A1B MoE models, developed based on different configurations.}
    \label{fig:moe_pt_exp_supp}
\end{figure*}

\subsection{Pre-Training}

Our analysis of the proposed generalized training objective reveals that it effectively regulates the trade-off between diversity and precision by strategically varying reward configurations. 
As illustrated in \Cref{fig:dense_pt_exp_supp} and \Cref{fig:moe_pt_exp_supp}, perplexity (PPL) consistently converges to comparable low values across both dense (1B, 4B) and MoE (5B-A0.3B, 10B-A0.5B, 20B-A1B) architectures.
This demonstrates that, within a specific range, modifying the reward function modulates training dynamics without compromising final predictive accuracy.
The parameter $\beta$ serves as a potent global entropy regulator. 
Specifically, setting $\beta < 0$ significantly reduces entropy, resulting in a more peaked and confident token distribution by amplifying rewards for ground turth tokens. 
Conversely, $\beta > 0$ maintains higher entropy and a flatter distribution, thereby promoting diversity in the generated output. 
Meanwhile, the parameters $\hat{\lambda}$ and $\tilde{\lambda}$ facilitate local entropy fine-tuning. 
These parameters shape the token distribution by either rewarding ($\hat{\lambda}=0, \tilde{\lambda}=0.1, k=100$) or penalizing ($\hat{\lambda}=-0.1, \tilde{\lambda}=0, k=100$) negative tokens, enabling granular control over the training process.

\begin{figure*}[!tb]
    \centering
    \begin{subfigure}{0.230\linewidth}
        \centering
        \includegraphics[width=\linewidth]{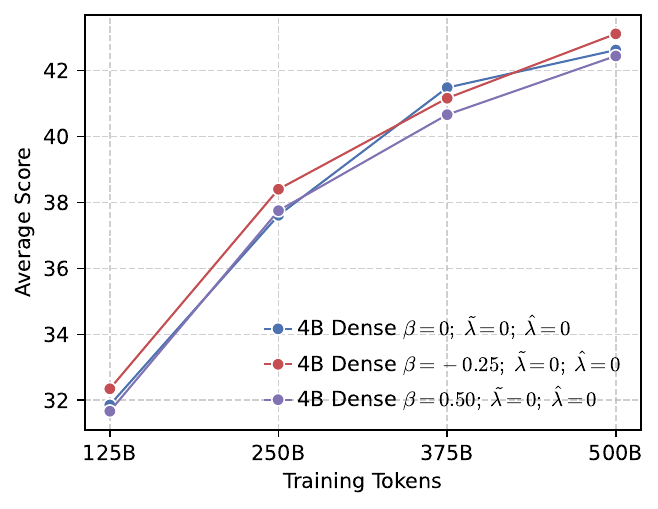}
    \end{subfigure}
    \hfill
    \begin{subfigure}{0.230\linewidth}
        \centering
        \includegraphics[width=\linewidth]{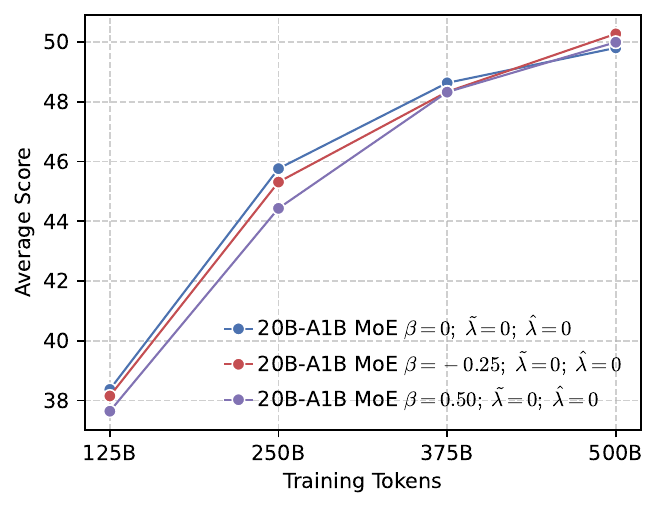}
    \end{subfigure}
    \hfill
    \begin{subfigure}{0.230\linewidth}
        \centering
        \includegraphics[width=\linewidth]{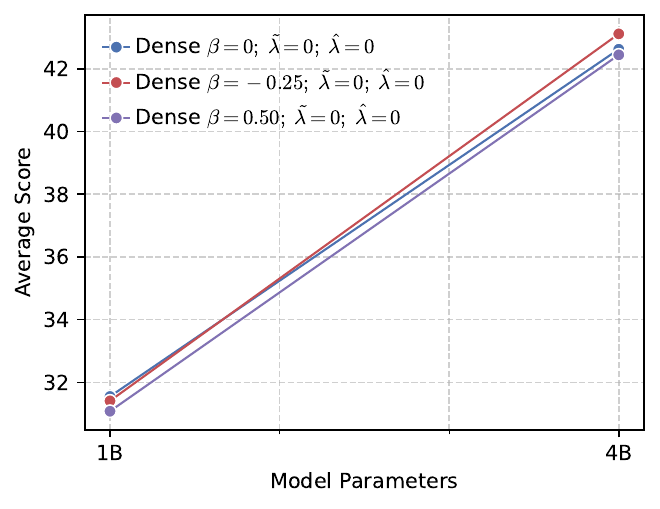}
    \end{subfigure}
    \hfill
    \begin{subfigure}{0.230\linewidth}
        \centering
        \includegraphics[width=\linewidth]{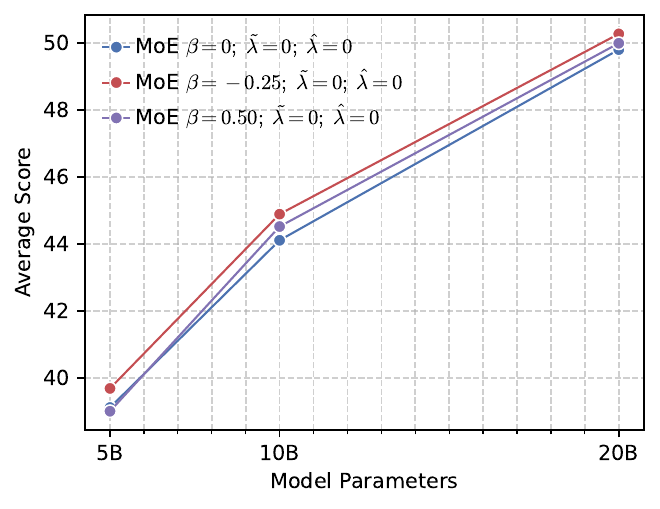}
    \end{subfigure}
    
    \begin{subfigure}{0.230\linewidth}
        \centering
        \includegraphics[width=\linewidth]{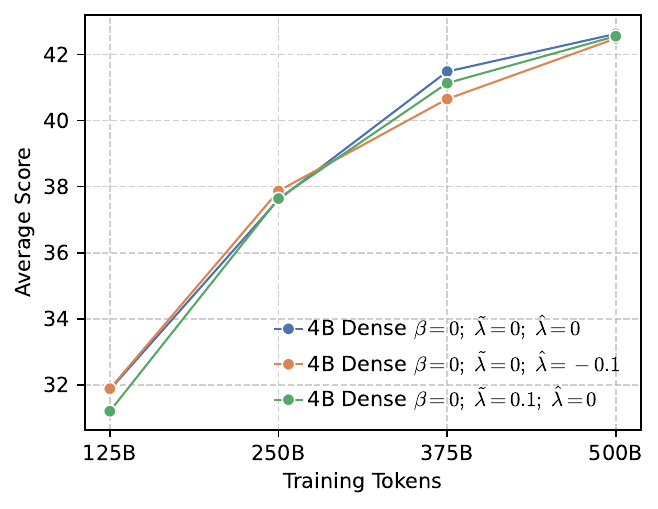}
    \end{subfigure}
    \hfill
    \begin{subfigure}{0.230\linewidth}
        \centering
        \includegraphics[width=\linewidth]{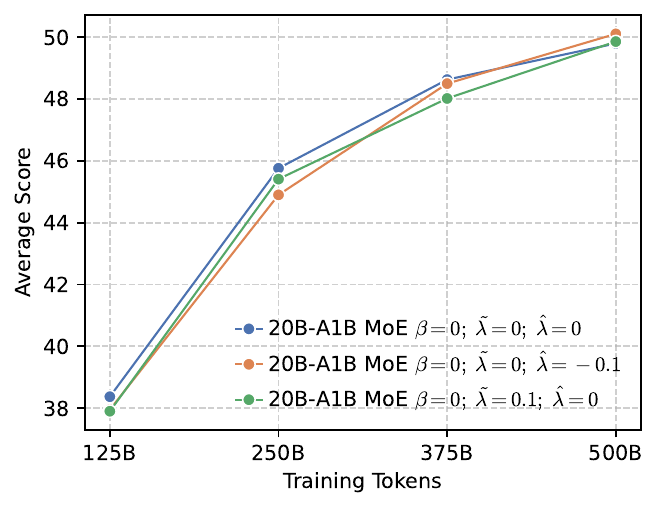}
    \end{subfigure}
    \hfill
    \begin{subfigure}{0.230\linewidth}
        \centering
        \includegraphics[width=\linewidth]{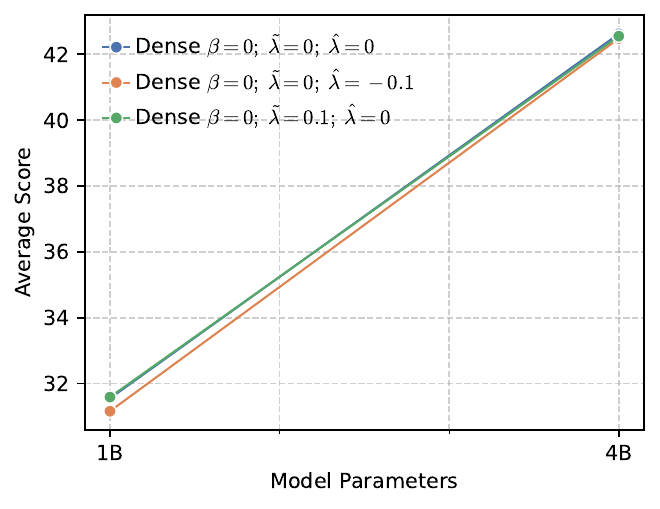}
    \end{subfigure}
    \hfill
    \begin{subfigure}{0.230\linewidth}
        \centering
        \includegraphics[width=\linewidth]{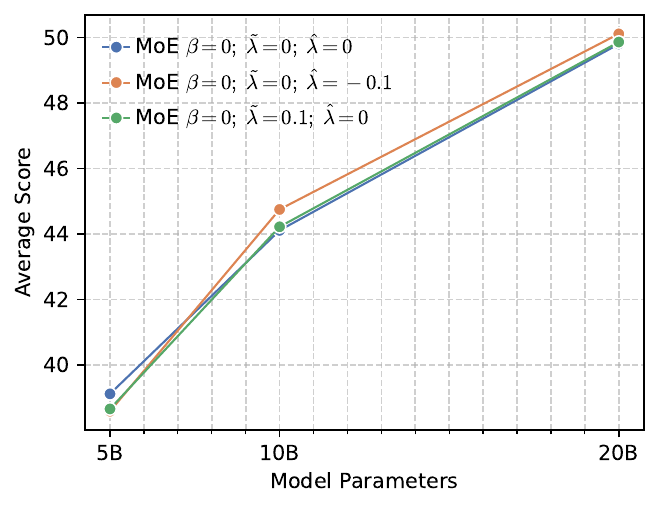}
    \end{subfigure}
    
    \caption{Changes of performance during pre-training across models with various model parameters, developed based on dense and MoE architectures under different configurations.}
    \label{fig:pt_exp}
\end{figure*}

Furthermore, we analyze the evolution of model performance during pre-training to investigate the dynamics and specific impact of the proposed reward function. 
As depicted in \Cref{fig:pt_exp}, larger models consistently achieve substantially higher final performance than smaller models after processing an equivalent number of training tokens. 
This confirms that explicitly regulating the diversity-precision trade-off is an orthogonal mechanism that does not interfere with the fundamental scaling properties of language models. 
Crucially, configurations that prioritize lowering global entropy ($\beta < 0$) or maintaining high local entropy ($\hat{\lambda}=-0.1, \tilde{\lambda}=0, k=100$) demonstrate superior performance and scaling behavior. 
Although these settings may not yield optimal initial performance in smaller models, they exhibit enhanced growth potential as model size increases. 
This suggests that with greater model capacity, strategies that promote precision, either globally via generously rewarding positive tokens or locally by aggressively penalizing tail negative tokens, lead to better performance growth compared to the baseline.

\begin{figure*}[!tb]
    \centering
    \begin{subfigure}{0.230\linewidth}
        \centering
        \includegraphics[width=\linewidth]{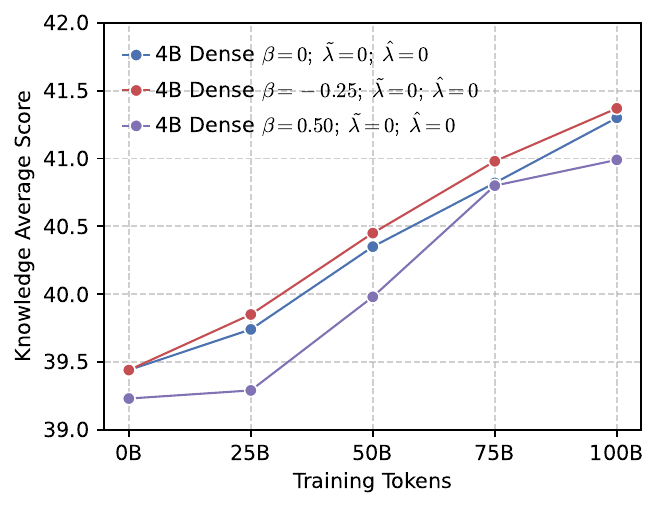}
    \end{subfigure}
    \hfill
    \begin{subfigure}{0.230\linewidth}
        \centering
        \includegraphics[width=\linewidth]{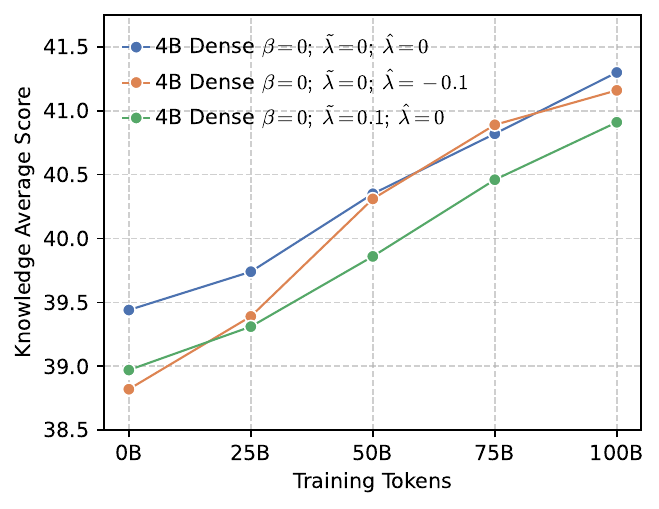}
    \end{subfigure}
    \hfill
    \begin{subfigure}{0.230\linewidth}
        \centering
        \includegraphics[width=\linewidth]{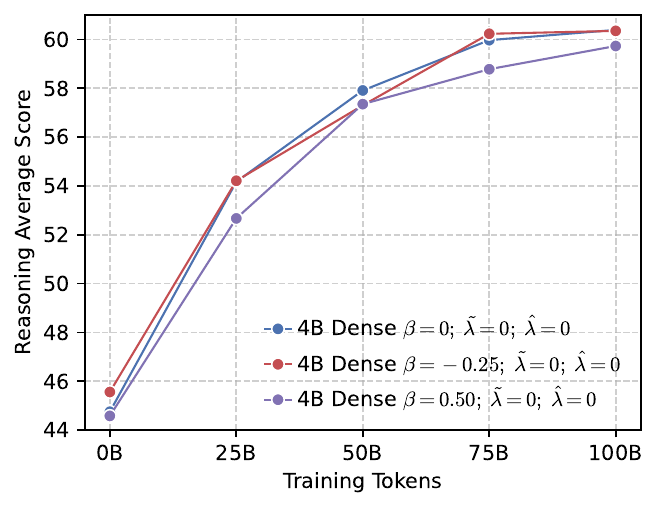}
    \end{subfigure}
    \hfill
    \begin{subfigure}{0.230\linewidth}
        \centering
        \includegraphics[width=\linewidth]{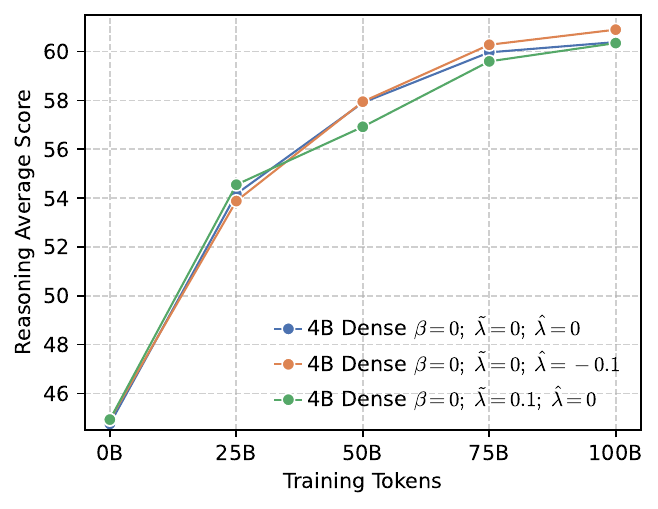}
    \end{subfigure}
    
    \begin{subfigure}{0.230\linewidth}
        \centering
        \includegraphics[width=\linewidth]{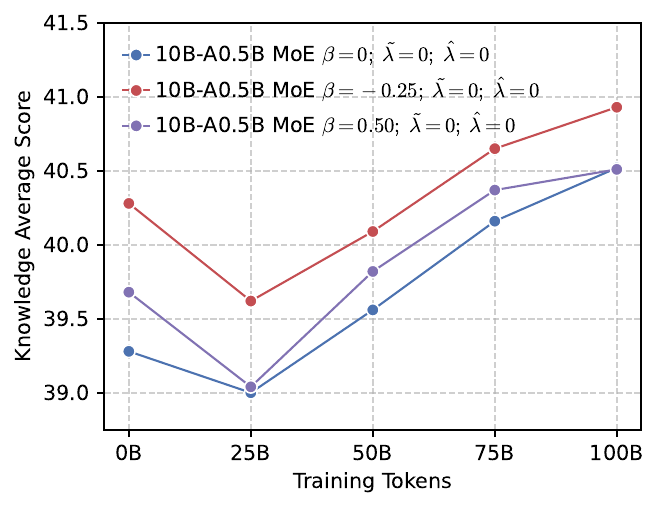}
    \end{subfigure}
    \hfill
    \begin{subfigure}{0.230\linewidth}
        \centering
        \includegraphics[width=\linewidth]{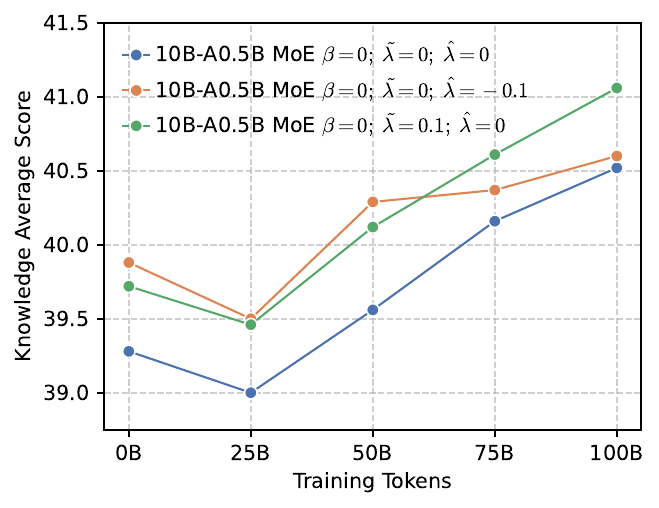}
    \end{subfigure}
    \hfill
    \begin{subfigure}{0.230\linewidth}
        \centering
        \includegraphics[width=\linewidth]{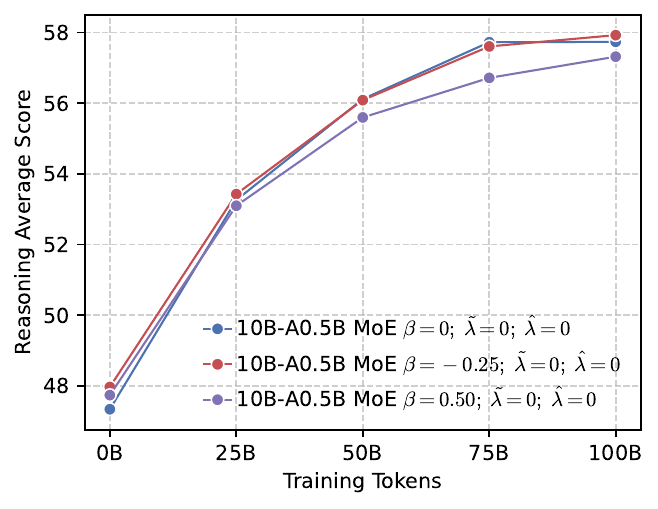}
    \end{subfigure}
    \hfill
    \begin{subfigure}{0.230\linewidth}
        \centering
        \includegraphics[width=\linewidth]{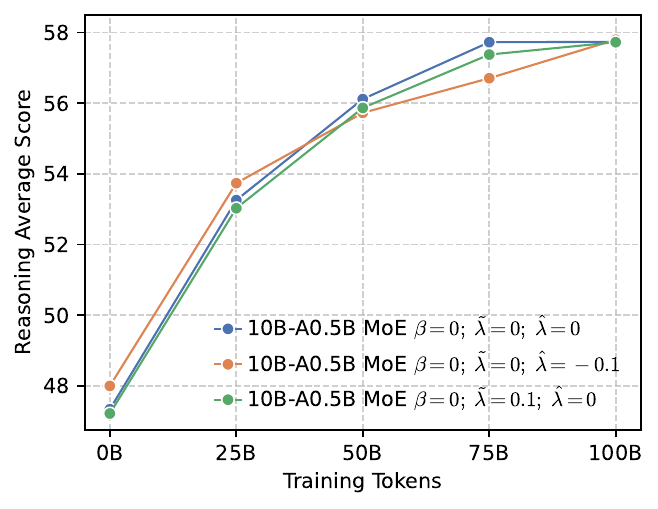}
    \end{subfigure}

    \begin{subfigure}{0.230\linewidth}
        \centering
        \includegraphics[width=\linewidth]{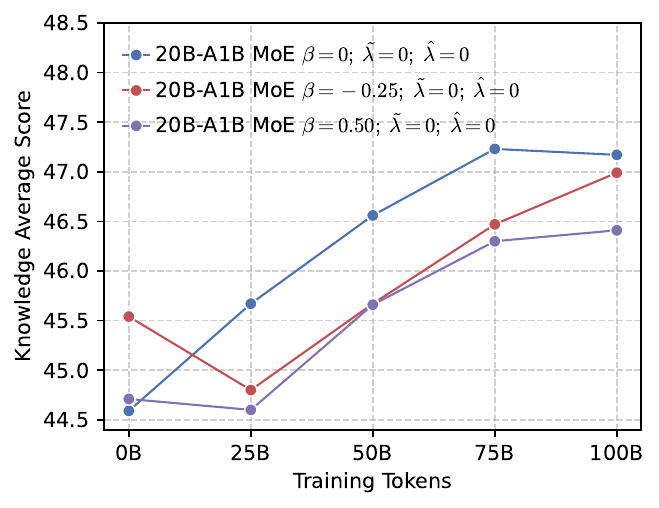}
    \end{subfigure}
    \hfill
    \begin{subfigure}{0.230\linewidth}
        \centering
        \includegraphics[width=\linewidth]{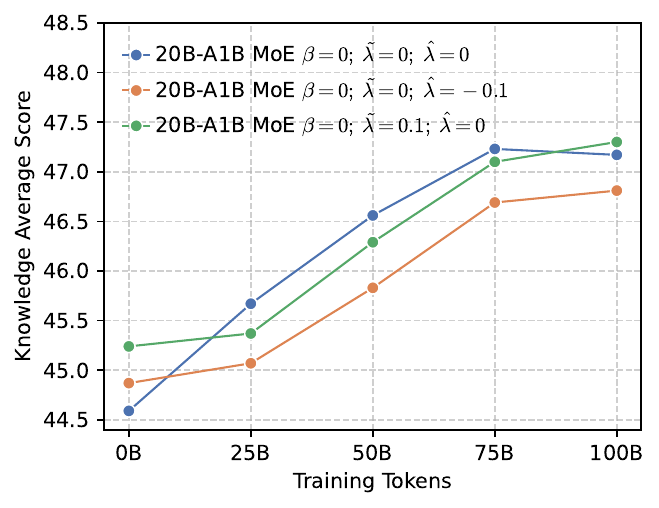}
    \end{subfigure}
    \hfill
    \begin{subfigure}{0.230\linewidth}
        \centering
        \includegraphics[width=\linewidth]{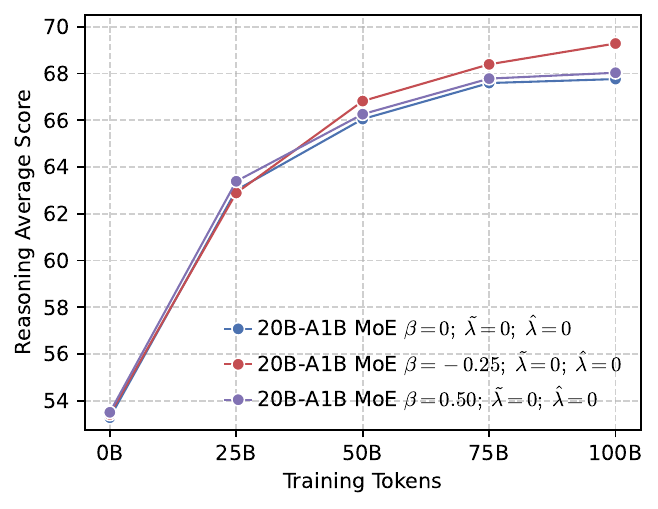}
    \end{subfigure}
    \hfill
    \begin{subfigure}{0.230\linewidth}
        \centering
        \includegraphics[width=\linewidth]{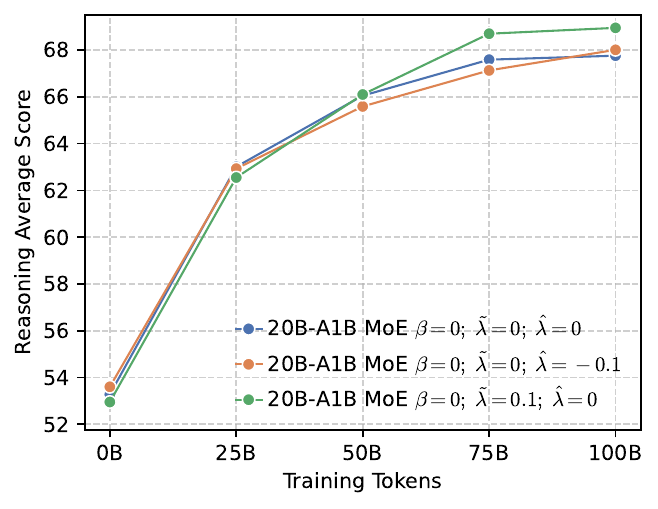}
    \end{subfigure}
    \caption{Changes of performance during mid-training across 4B dense, 10B-A0.5B MoE and 20B-A1B MoE  models, developed based on different configurations.}
    \label{fig:mt_exp}
\end{figure*}

\subsection{Mid-Training}

Subsequently, we evaluate the evolution of model performance during the mid-training stage, spanning from 0B to 100B tokens.
As depicted in \Cref{fig:mt_exp}, the choice of $\beta$ significantly influences training dynamics. 
We observe a consistent trend where a negative value, specifically $\beta = -0.25$, yields the best results. 
This configuration consistently outperforms the baseline ($\beta = 0$) across both dense and MoE models in knowledge and reasoning tasks. 
Conversely, a positive setting ($\beta = 0.50$) does not demonstrate consistent superior performance comparing to the baseline.
Similar to the observations with $\beta$, a slight negative adjustment appears beneficial. 
The configuration $\hat{\lambda}=-0.1, \tilde{\lambda}=0, k=100$ generally matches or slightly surpasses the performance of the standard CE baseline.
However, when shifting to $\hat{\lambda}=0.1, \tilde{\lambda}=0, k=100$, performance exhibited uncertainty in knowledge-intensive scenarios. 
In reasoning tasks, the performance remained comparable to the standard CE baseline.

\begin{figure*}[!tb]
    \centering
    \begin{subfigure}{0.315\linewidth}
        \centering
        \includegraphics[width=\linewidth]{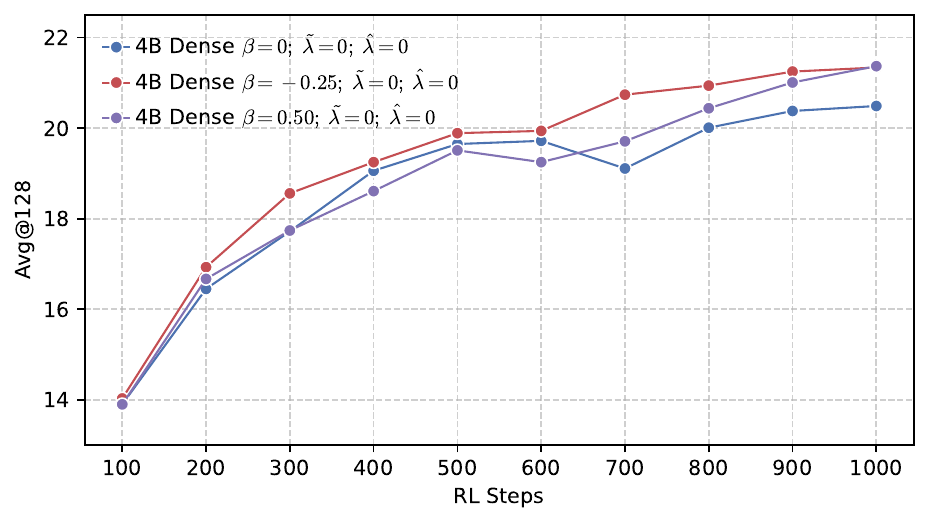}
    \end{subfigure}
    \hfill
    \begin{subfigure}{0.315\linewidth}
        \centering
        \includegraphics[width=\linewidth]{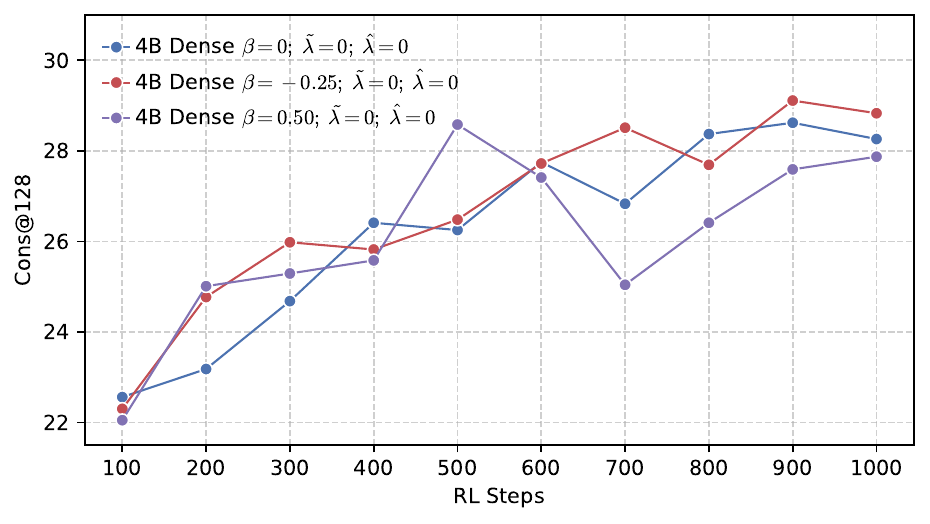}
    \end{subfigure}
    \hfill
    \begin{subfigure}{0.315\linewidth}
        \centering
        \includegraphics[width=\linewidth]{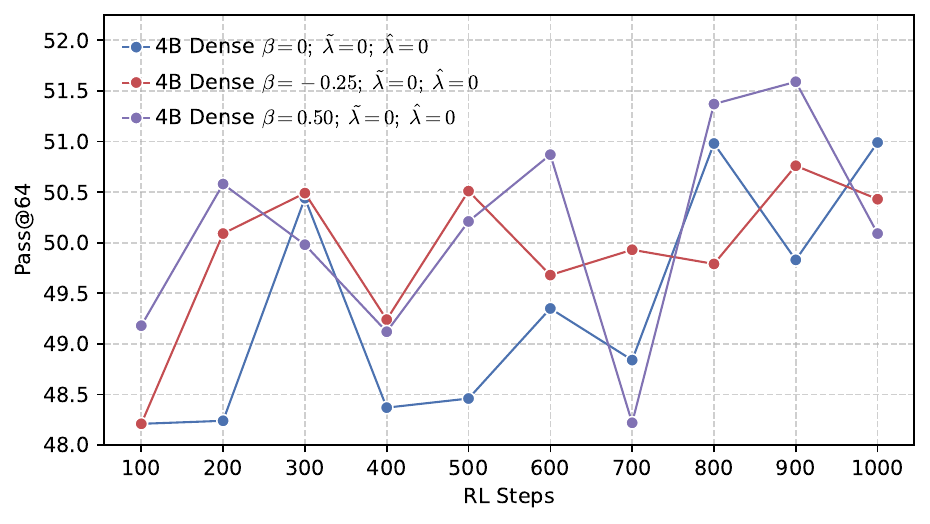}
    \end{subfigure}
    
    \begin{subfigure}{0.315\linewidth}
        \centering
        \includegraphics[width=\linewidth]{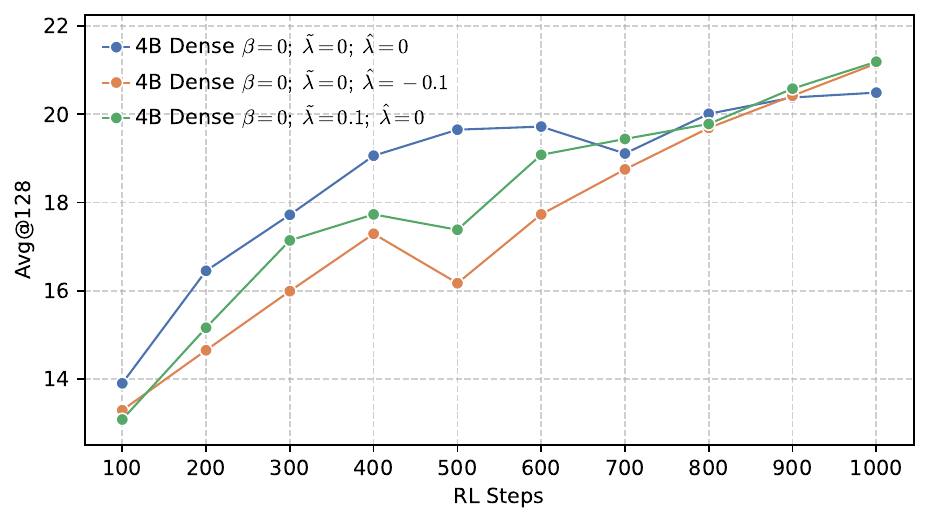}
    \end{subfigure}
    \hfill
    \begin{subfigure}{0.315\linewidth}
        \centering
        \includegraphics[width=\linewidth]{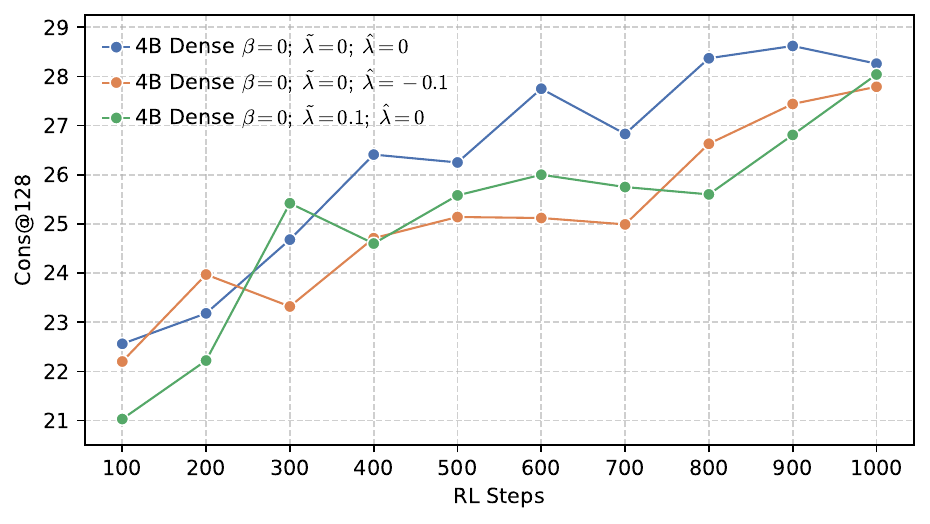}
    \end{subfigure}
    \hfill
    \begin{subfigure}{0.315\linewidth}
        \centering
        \includegraphics[width=\linewidth]{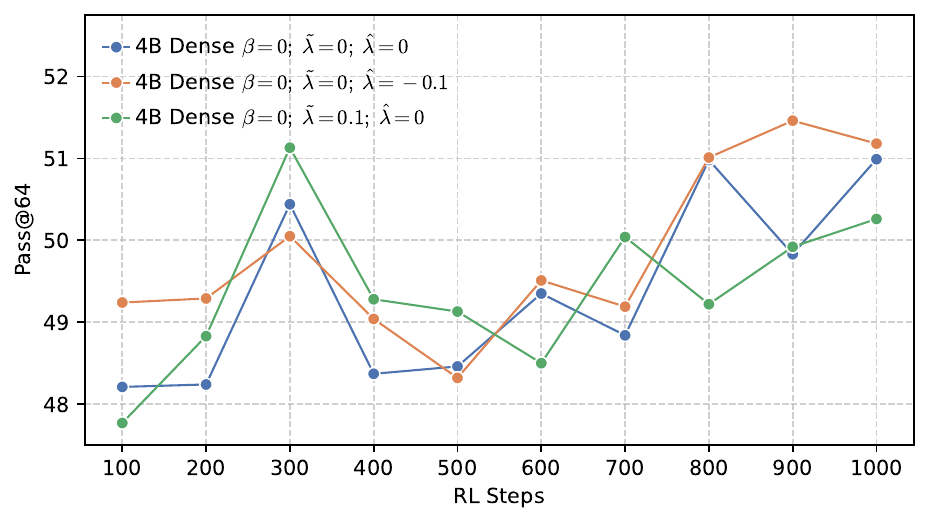}
    \end{subfigure}
    \caption{Changes of performance during RL training across various actor models, developed based on a 4B dense architecture under different configurations.}
    \label{fig:dense_rl_exp}
\end{figure*}

\begin{figure*}[!tb]
    \centering
    \begin{subfigure}{0.315\linewidth}
        \centering
        \includegraphics[width=\linewidth]{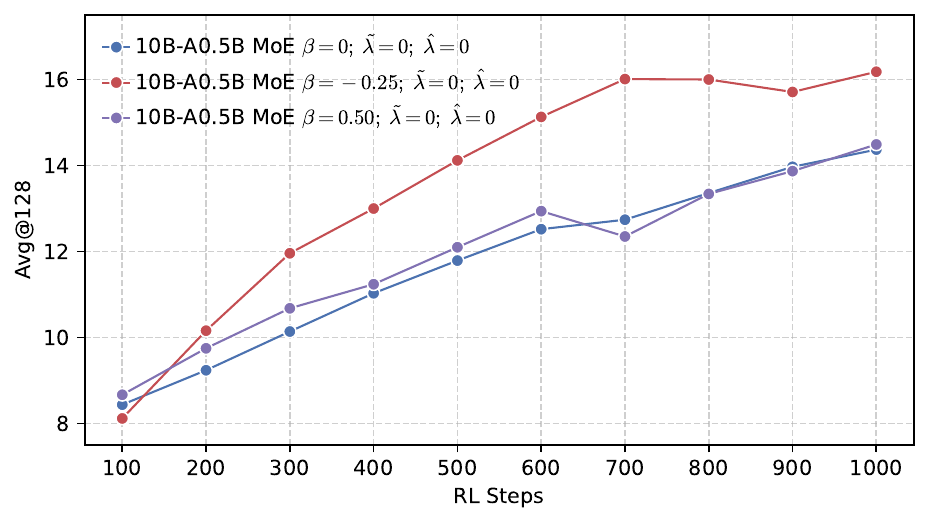}
    \end{subfigure}
    \hfill
    \begin{subfigure}{0.315\linewidth}
        \centering
        \includegraphics[width=\linewidth]{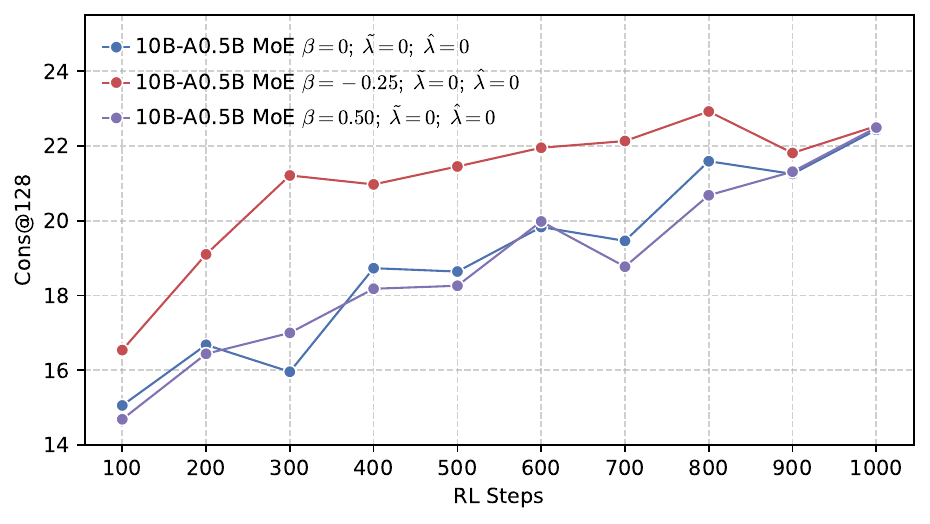}
    \end{subfigure}
    \hfill
    \begin{subfigure}{0.315\linewidth}
        \centering
        \includegraphics[width=\linewidth]{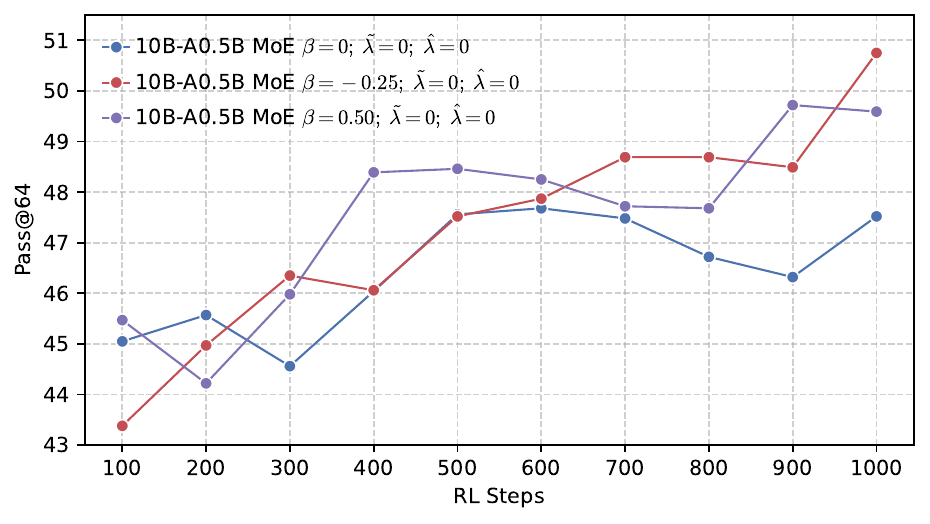}
    \end{subfigure}
    
    \begin{subfigure}{0.315\linewidth}
        \centering
        \includegraphics[width=\linewidth]{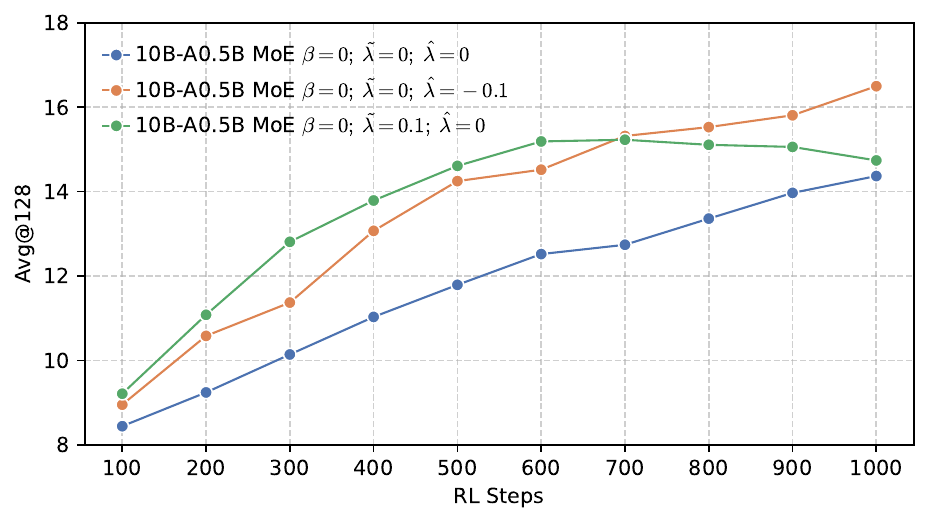}
    \end{subfigure}
    \hfill
    \begin{subfigure}{0.315\linewidth}
        \centering
        \includegraphics[width=\linewidth]{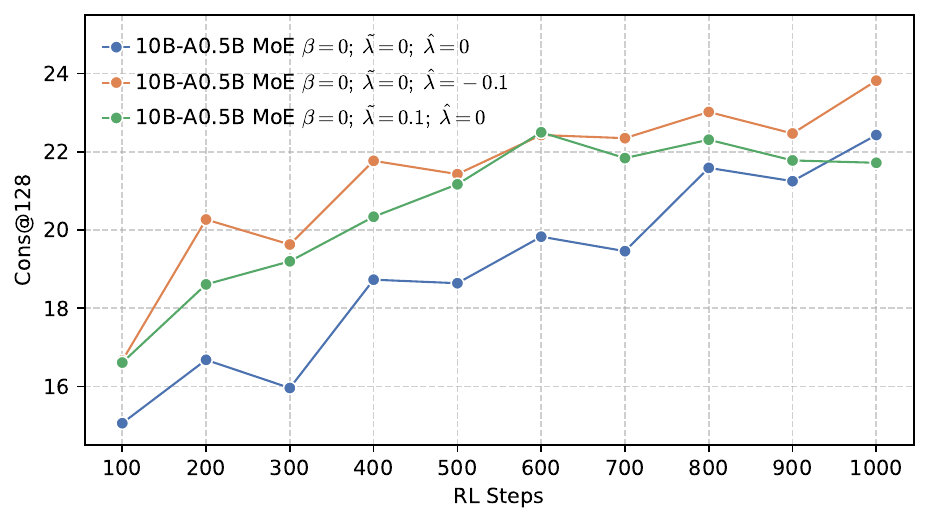}
    \end{subfigure}
    \hfill
    \begin{subfigure}{0.315\linewidth}
        \centering
        \includegraphics[width=\linewidth]{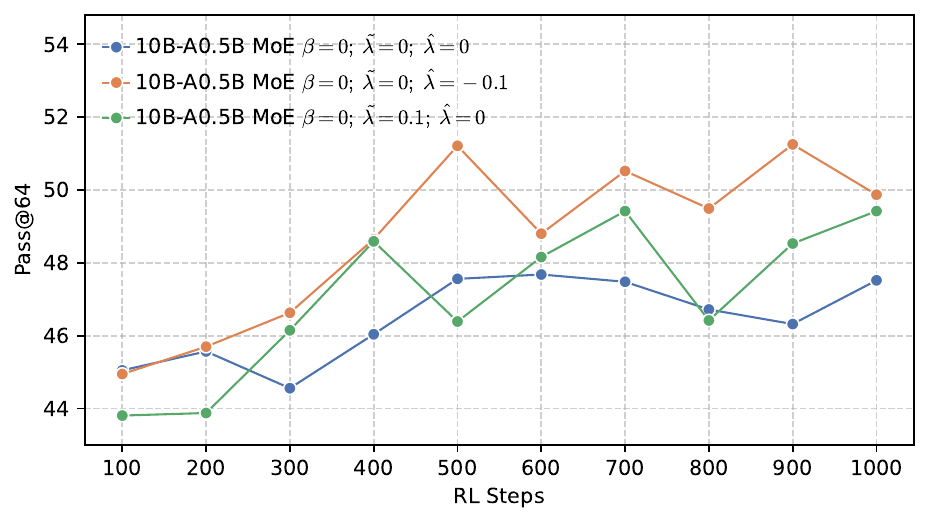}
    \end{subfigure}
    \caption{Changes of performance during RL training across various actor models, developed based on a 10B-A0.5B MoE architecture under different configurations.}
    \label{fig:moe_rl_exp}
\end{figure*}

\begin{figure*}[!tb]
    \centering
    \begin{subfigure}{0.315\linewidth}
        \centering
        \includegraphics[width=\linewidth]{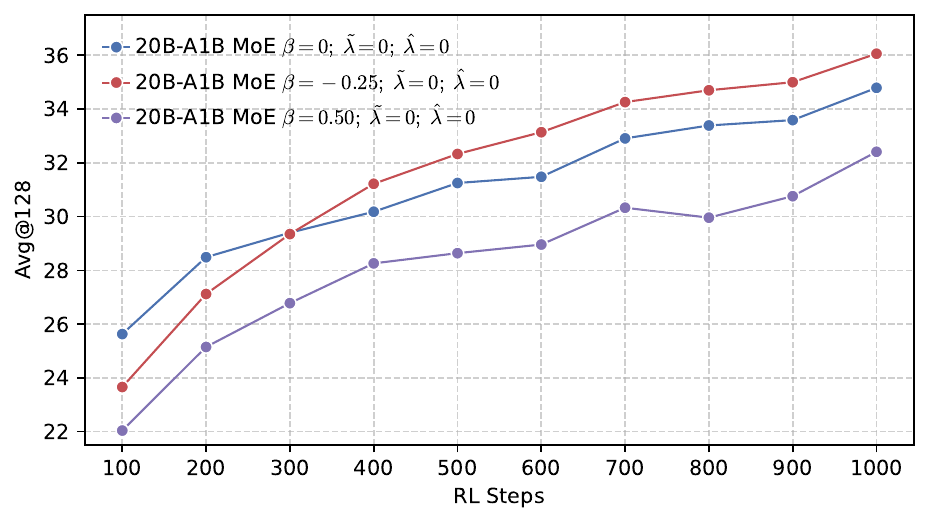}
    \end{subfigure}
    \hfill
    \begin{subfigure}{0.315\linewidth}
        \centering
        \includegraphics[width=\linewidth]{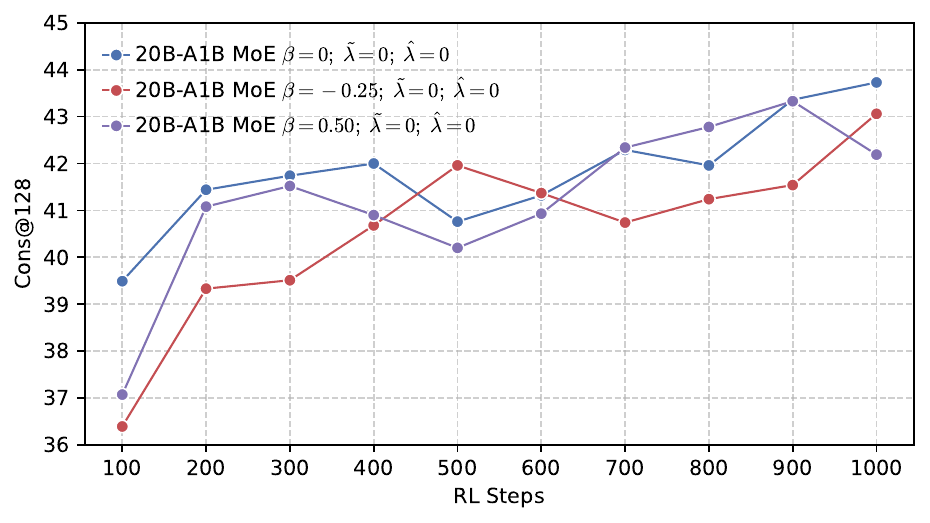}
    \end{subfigure}
    \hfill
    \begin{subfigure}{0.315\linewidth}
        \centering
        \includegraphics[width=\linewidth]{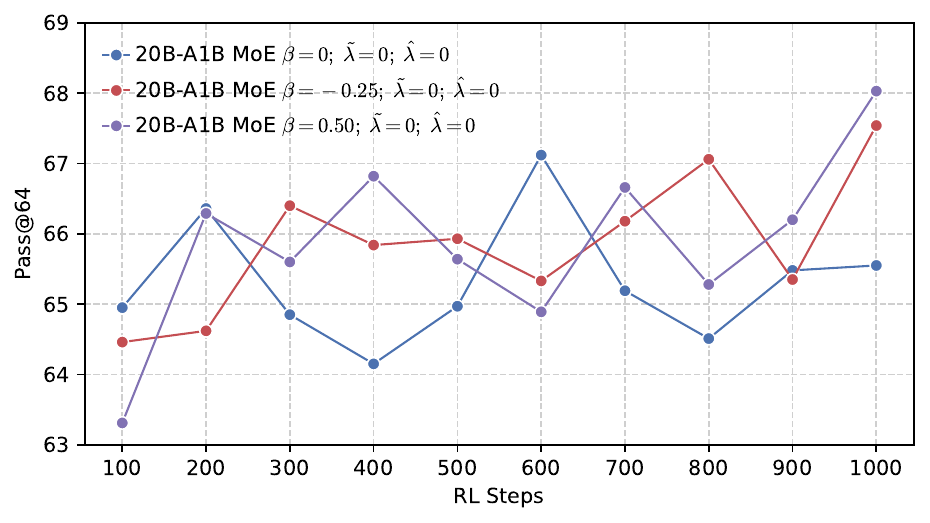}
    \end{subfigure}
    
    \begin{subfigure}{0.315\linewidth}
        \centering
        \includegraphics[width=\linewidth]{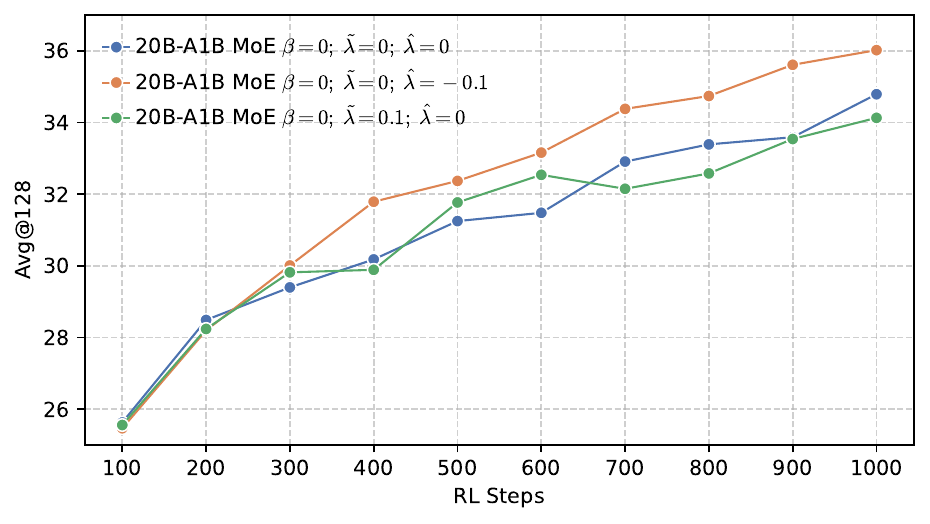}
    \end{subfigure}
    \hfill
    \begin{subfigure}{0.315\linewidth}
        \centering
        \includegraphics[width=\linewidth]{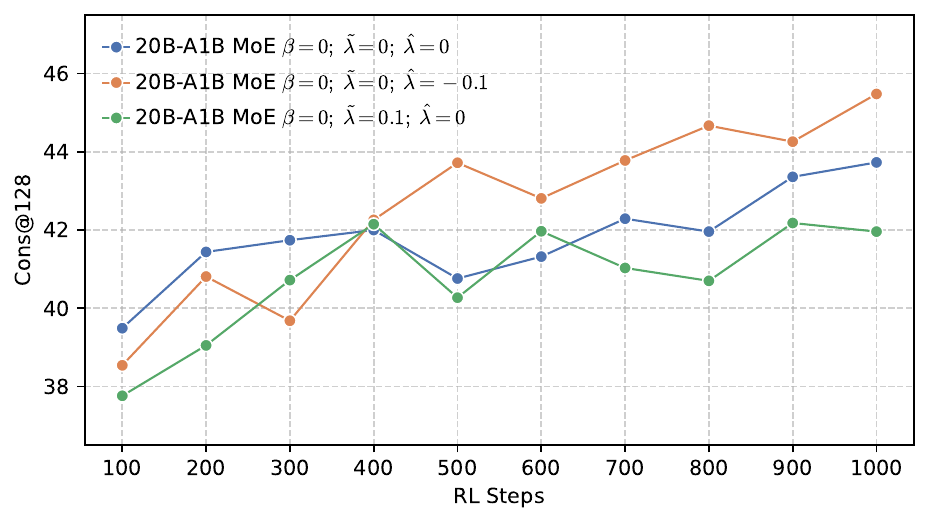}
    \end{subfigure}
    \hfill
    \begin{subfigure}{0.315\linewidth}
        \centering
        \includegraphics[width=\linewidth]{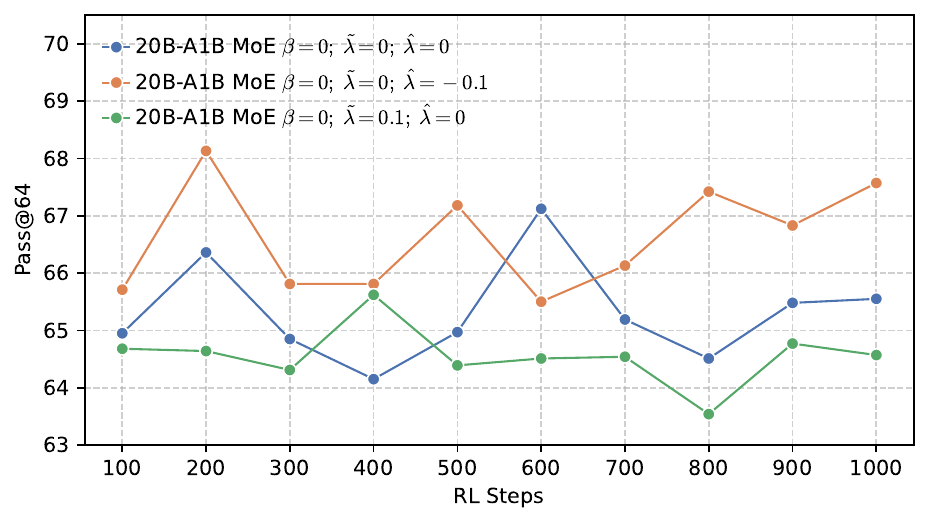}
    \end{subfigure}
    \caption{Changes of performance during RL training across various actor models, developed based on a 20B-A1B MoE architecture under different configurations.}
    \label{fig:flag_rl_exp}
\end{figure*}

\subsection{Reinforcement Learning}

Finally, we investigate the performance dynamics during the RL training stage across various actor models, as illustrated in \Crefrange{fig:dense_rl_exp}{fig:flag_rl_exp}. 
Pre-trained models derived from different reward configurations exhibit distinct output distributions, leading to significant variations in subsequent RL and end-to-end reasoning performance. 
Regarding the global entropy regulator $\beta$, we observe a consistent and robust trend across both the 4B dense and 10B-A0.5B MoE models. 
Specifically, the global low entropy setting ($\beta = -0.25$) yields superior performance trajectories. 
This configuration consistently outperforms the global high entropy setting across all evaluated metrics, including $\text{Avg@}$128, $\text{Cons@}$128, and $\text{Pass@}$64. 
Furthermore, the configuration $\hat{\lambda}=-0.1, \tilde{\lambda}=0, k=100$ demonstrates a significant advantage, consistently achieving the highest performance and notably surpassing the baseline on the 10B-A0.5B MoE model. 
For the 4B dense model, maintaining local high entropy exhibits a superior scaling trend compared to the baseline. 
In conclusion, strategies that promote precision, either globally via generously rewarding positive tokens or locally by aggressively penalizing tail negative tokens, enables the model to converge to higher-quality solutions, potentially providing a better exploration space for RL.

\begin{figure*}[!tb]
    \centering
    \begin{subfigure}{0.230\linewidth}
        \centering
        \includegraphics[width=\linewidth]{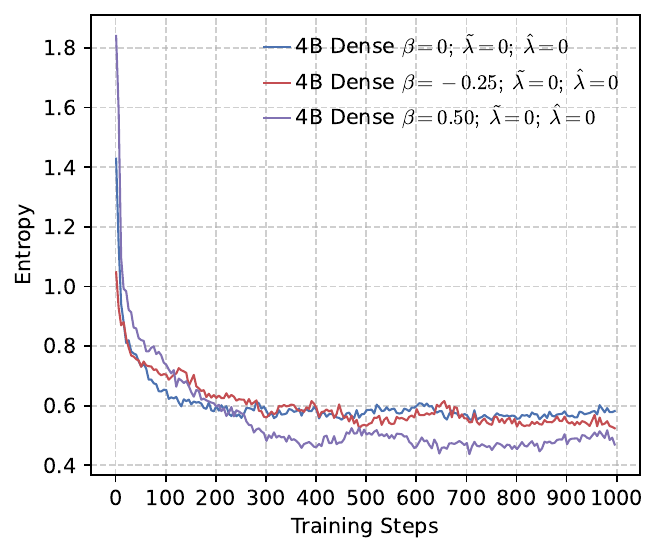}
    \end{subfigure}
    \hfill
    \begin{subfigure}{0.230\linewidth}
        \centering
        \includegraphics[width=\linewidth]{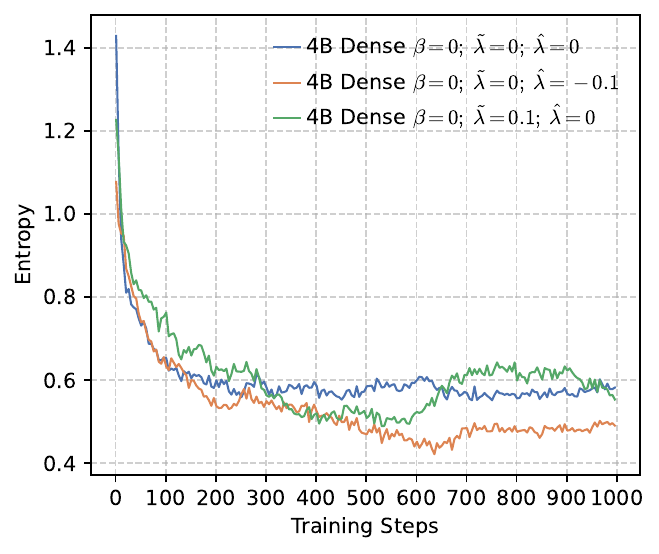}
    \end{subfigure}
    \hfill
    \begin{subfigure}{0.230\linewidth}
        \centering
        \includegraphics[width=\linewidth]{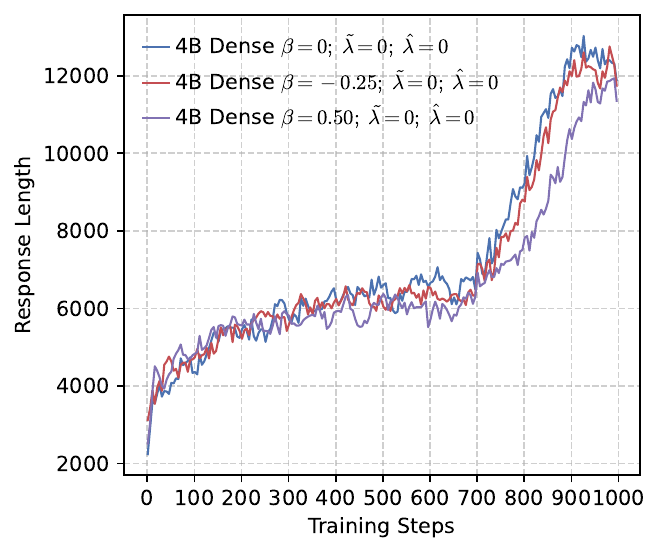}
    \end{subfigure}
    \hfill
    \begin{subfigure}{0.230\linewidth}
        \centering
        \includegraphics[width=\linewidth]{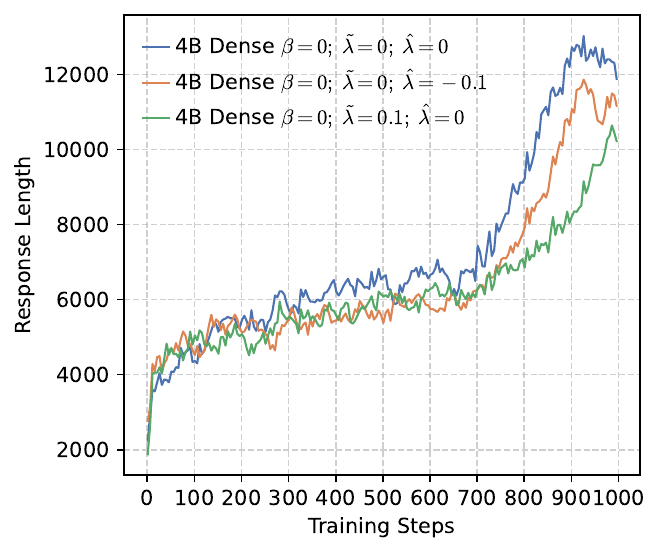}
    \end{subfigure}
    
    \begin{subfigure}{0.230\linewidth}
        \centering
        \includegraphics[width=\linewidth]{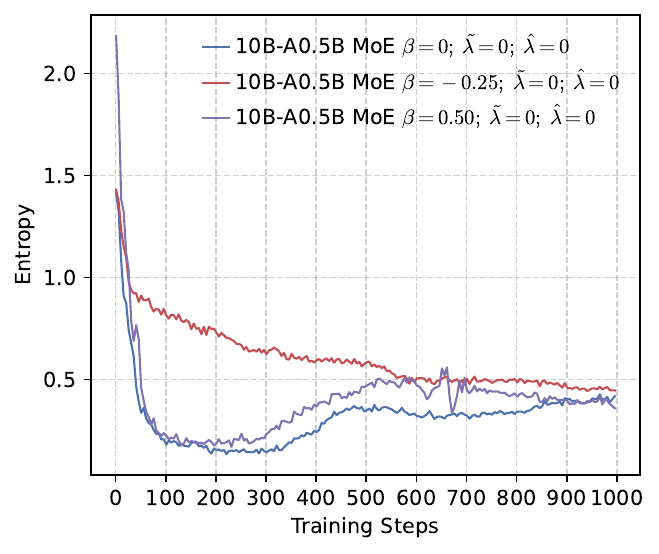}
    \end{subfigure}
    \hfill
    \begin{subfigure}{0.230\linewidth}
        \centering
        \includegraphics[width=\linewidth]{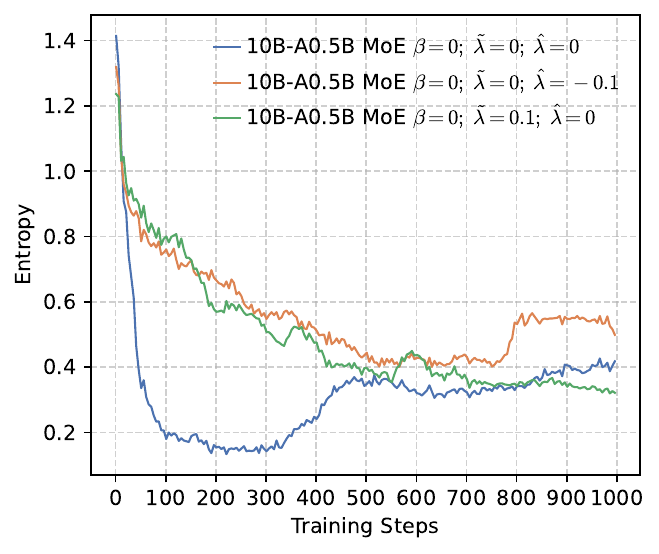}
    \end{subfigure}    
    \hfill
    \begin{subfigure}{0.230\linewidth}
        \centering
        \includegraphics[width=\linewidth]{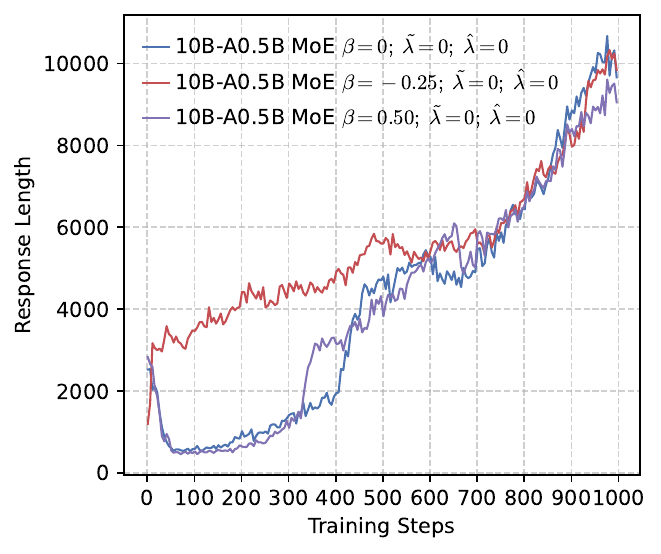}
    \end{subfigure}
    \hfill
    \begin{subfigure}{0.230\linewidth}
        \centering
        \includegraphics[width=\linewidth]{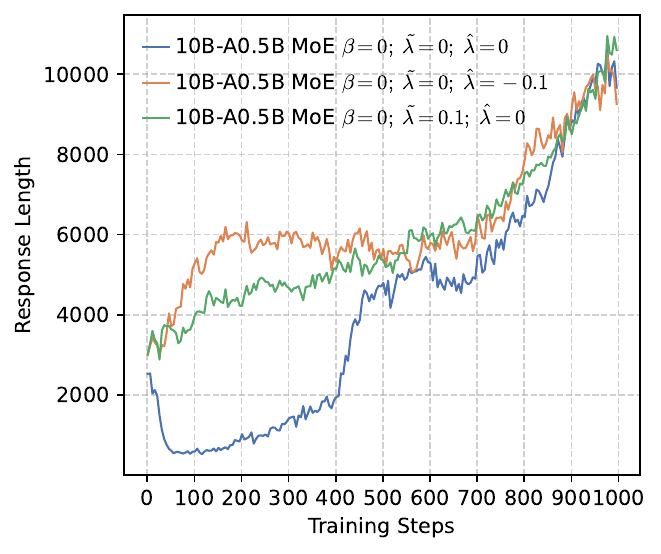}
    \end{subfigure}
    
    \begin{subfigure}{0.230\linewidth}
        \centering
        \includegraphics[width=\linewidth]{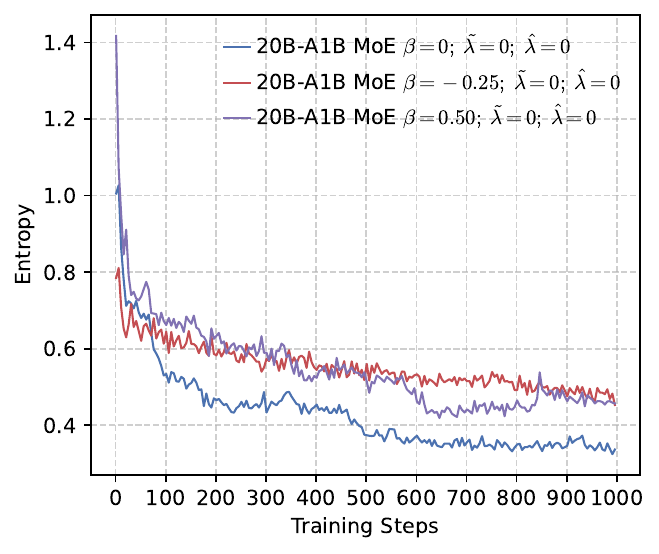}
    \end{subfigure}
    \hfill
    \begin{subfigure}{0.230\linewidth}
        \centering
        \includegraphics[width=\linewidth]{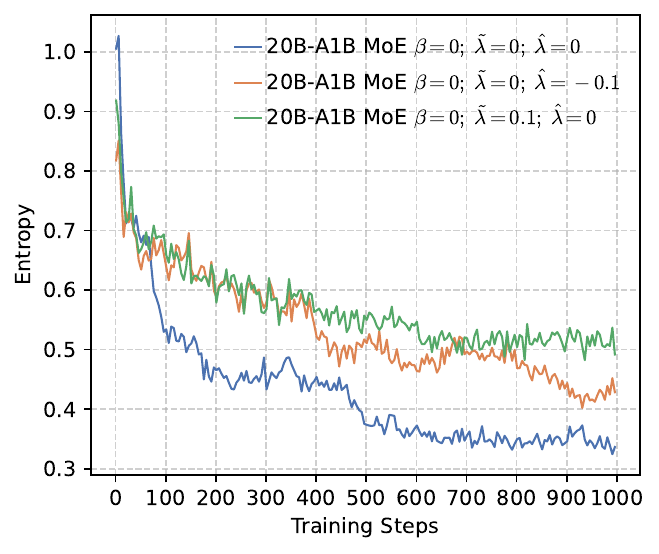}
    \end{subfigure}    
    \hfill
    \begin{subfigure}{0.230\linewidth}
        \centering
        \includegraphics[width=\linewidth]{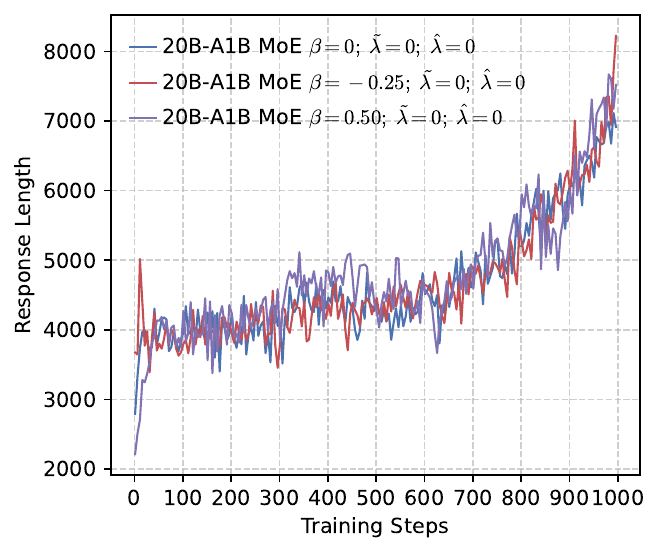}
    \end{subfigure}
    \hfill
    \begin{subfigure}{0.230\linewidth}
        \centering
        \includegraphics[width=\linewidth]{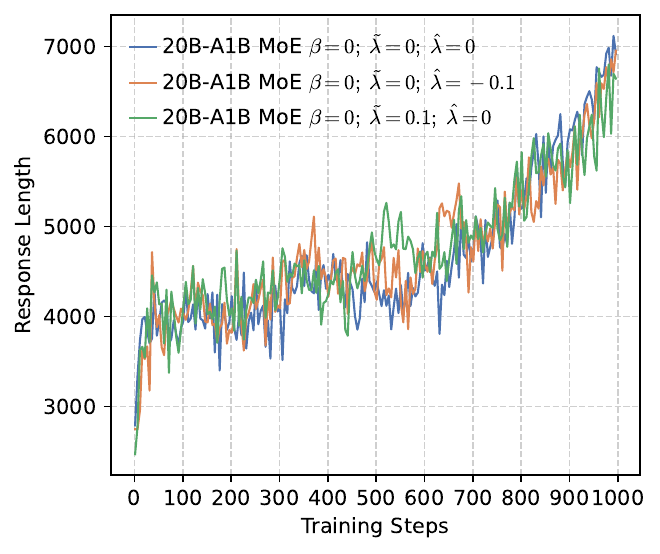}
    \end{subfigure}    
    
    \caption{Changes of entropy and response length during RL training across various actor models, developed based on 4B dense, 10B-A0.5B MoE, and 20B-A1B MoE architectures under different configurations.}
    \label{fig:rl_exp_supp}
\end{figure*}

To better understand the performance divergence observed during RL, we analyze the evolution of policy entropy and response length throughout the training process, as illustrated in \Cref{fig:rl_exp_supp}. 
Contrary to the expectation that higher entropy maintains diversity, setting a higher $\beta$ leads to rapid entropy collapse during the early stages of training.
Coinciding with this collapse, the response length decreases drastically, indicating a suppression of the reasoning capability.
In contrast, local high-entropy configurations exhibit greater stability. 
These settings effectively prevent entropy collapse, maintaining a robust policy distribution from the onset. 
They demonstrate a smooth and continuous activation of long reasoning capabilities, allowing for a steady increase in generation length and reasoning depth without the recovery lag observed in global high entropy settings.

\subsection{$\text{Pass@}k$ Analysis of Base Models}

\begin{figure*}[!tb]
    \centering
    \begin{subfigure}{0.230\linewidth}
        \centering
        \includegraphics[width=\linewidth]{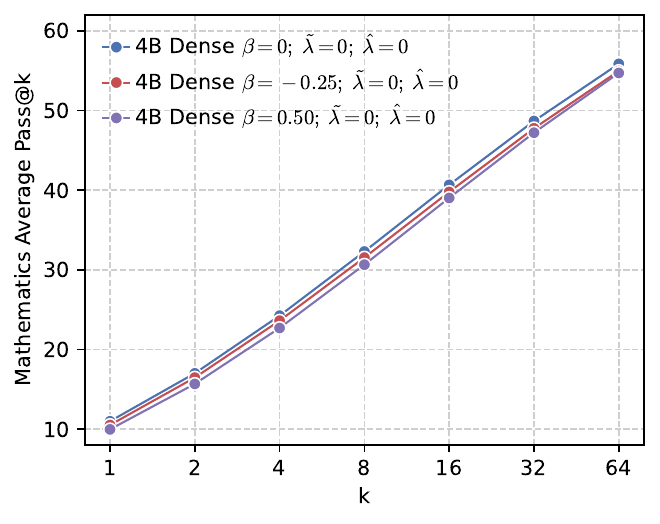}
    \end{subfigure}
    \hfill
    \begin{subfigure}{0.230\linewidth}
        \centering
        \includegraphics[width=\linewidth]{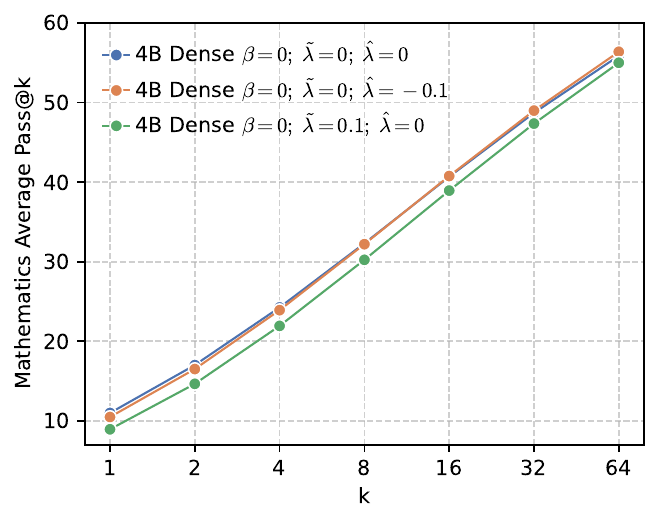}
    \end{subfigure}
    \hfill    
    \begin{subfigure}{0.230\linewidth}
        \centering
        \includegraphics[width=\linewidth]{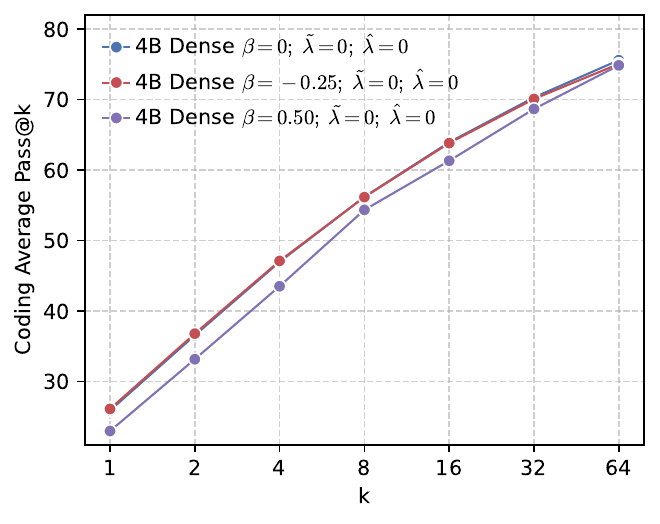}
    \end{subfigure}
    \hfill
    \begin{subfigure}{0.230\linewidth}
        \centering
        \includegraphics[width=\linewidth]{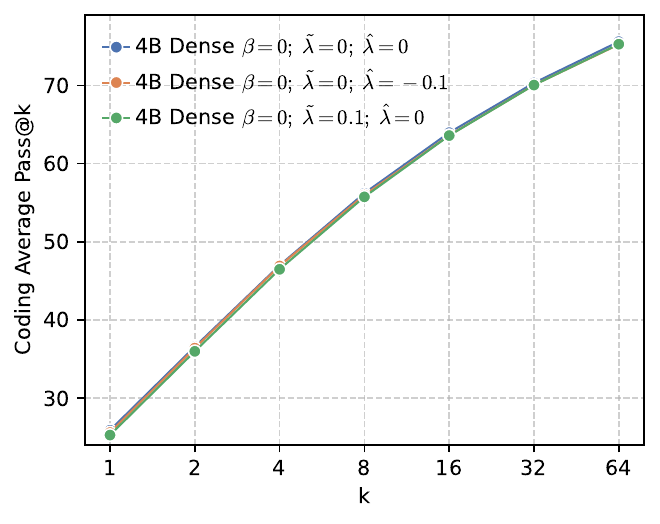}
    \end{subfigure}
    
    \begin{subfigure}{0.230\linewidth}
        \centering
        \includegraphics[width=\linewidth]{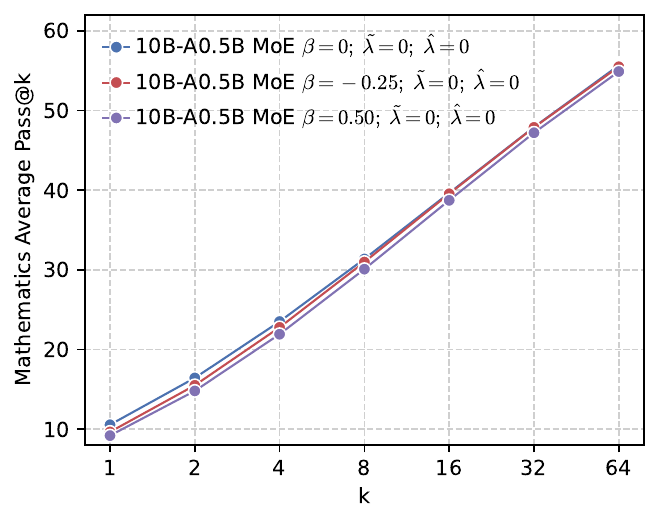}
    \end{subfigure}
    \hfill
    \begin{subfigure}{0.230\linewidth}
        \centering
        \includegraphics[width=\linewidth]{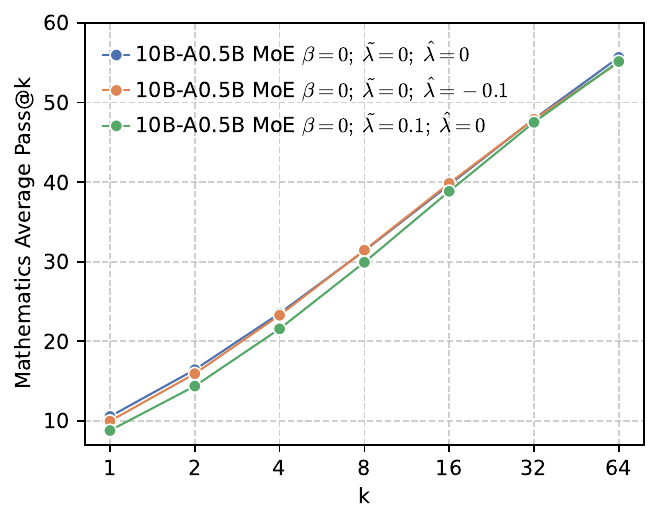}
    \end{subfigure}    
    \hfill
    \begin{subfigure}{0.230\linewidth}
        \centering
        \includegraphics[width=\linewidth]{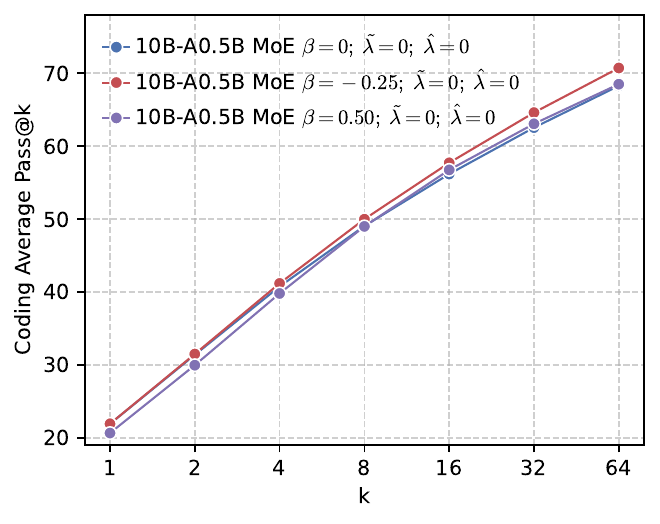}
    \end{subfigure}
    \hfill
    \begin{subfigure}{0.230\linewidth}
        \centering
        \includegraphics[width=\linewidth]{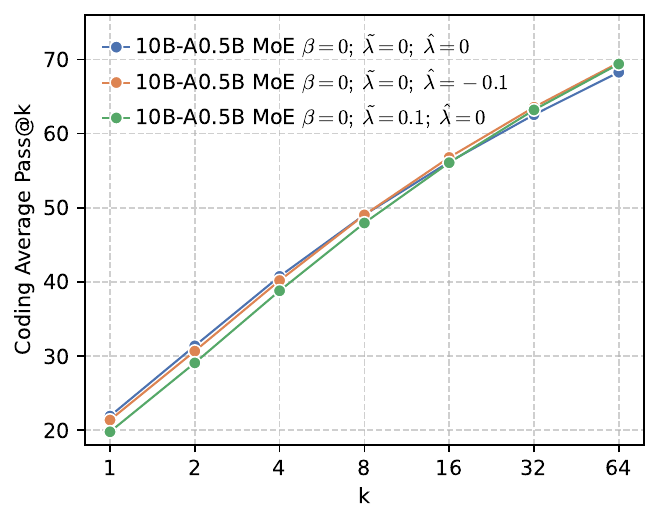}
    \end{subfigure}
    
    \begin{subfigure}{0.230\linewidth}
        \centering
        \includegraphics[width=\linewidth]{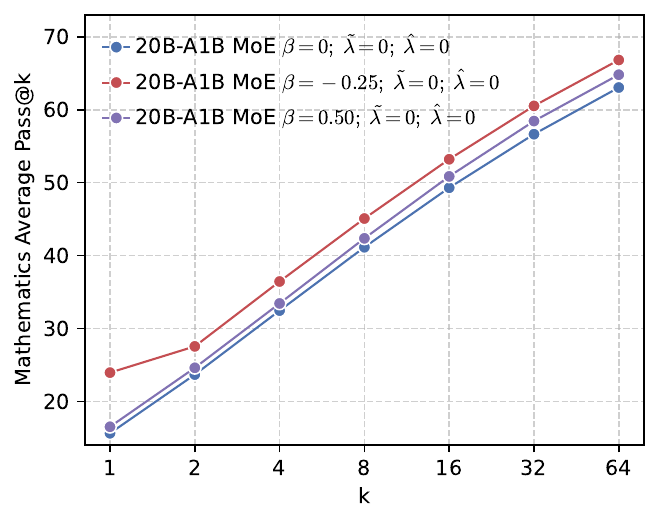}
    \end{subfigure}
    \hfill
    \begin{subfigure}{0.230\linewidth}
        \centering
        \includegraphics[width=\linewidth]{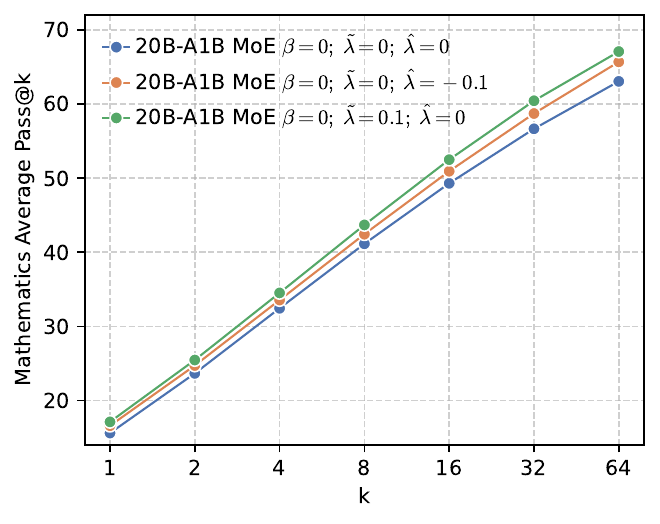}
    \end{subfigure}    
    \hfill
    \begin{subfigure}{0.230\linewidth}
        \centering
        \includegraphics[width=\linewidth]{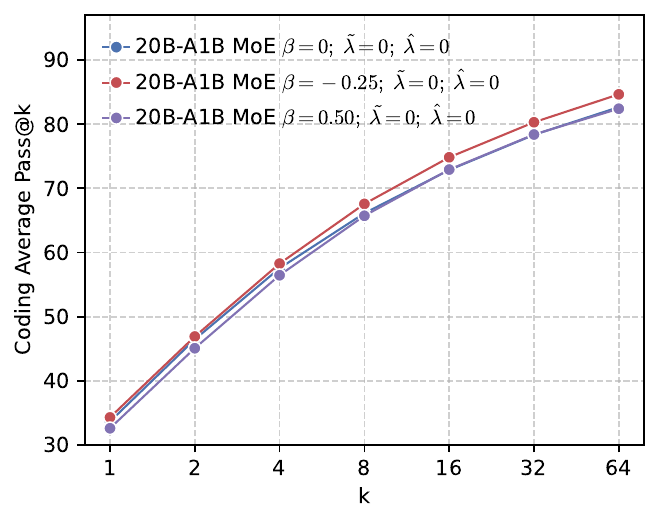}
    \end{subfigure}
    \hfill
    \begin{subfigure}{0.230\linewidth}
        \centering
        \includegraphics[width=\linewidth]{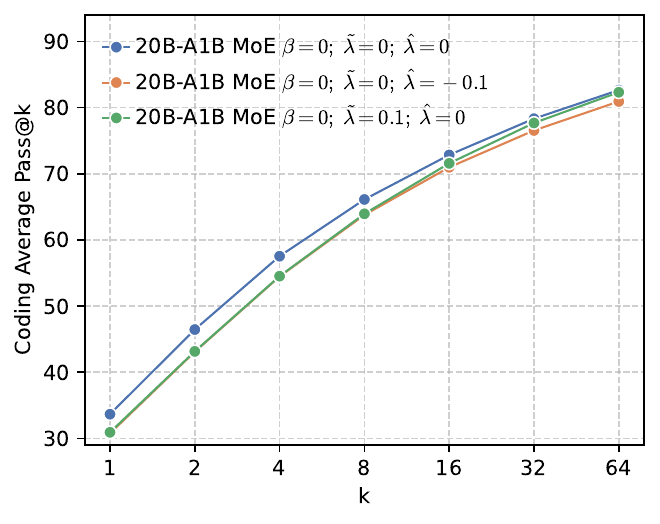}
    \end{subfigure}
    \caption{$\text{Pass@}k$ curve of base models on mathematics reasoning and code generation tasks, developed based on 4B dense, 10B-A0.5B MoE and 20B-A1B MoE models under different configurations.}
    \label{fig:passk_exp}
\end{figure*}

Moreover, we analyze the $\text{Pass@}k$ curves as $k$ increases to estimate the upper bound of the capability of base models. 
This metric relies on a delicate equilibrium between solution precision and diversity. 
As shown in \Cref{fig:passk_exp}, maximizing global diversity (high entropy) does not inherently yield higher $\text{Pass@}k$ curves. 
Instead, superior $\text{Pass@}k$ scores in mathematics and coding tasks are achieved by prioritizing precision. 
Crucially, we observe that this low-entropy setting does not lead to a collapse in output diversity.
Rather, it maintains sufficient variation to cover the solution space. 
Furthermore, the data indicate that promoting local diversity also yields better results. 
This suggests that while models benefit from high precision, they simultaneously benefit from targeted local exploration.

\section{Related Works}

\subsection{Weighted Cross-Entropy Loss}

The standard cross-entropy objective can be generalized within a policy-gradient framework, where it is equivalent to optimizing a sparse reward defined as $r_{\text{CE}}(s_t, a_t) = \mathbbm{1}(a_t = x_t)\pi_\theta(a_t \mid s_t)^{-1}$. 
Existing modifications to this objective include smooth loss (label smoothing), which encourages diversity by allocating a uniform probability mass to all positive tokens, and focal loss~\citep{lin2018focalloss}, which down-weights easy examples via $w_t = (1 - \pi_\theta(x_t \mid s_t))^\gamma$.
Our proposed generalized training objective can also formulate these established variations.
In this paper, we specifically explore two different reward configurations within this framework.
Firstly, we introduce a modified positive reward, which is equivalent to applying a state-dependent weight $w_t = \pi_\theta(x_t \mid s_t)^{1-(1-\pi_\theta(x_t \mid s_t))^{\beta}}$ to the standard cross-entropy. 
In addition, we incorporate TopK-based negative shaping, which explicitly controls local entropy by assigning non-zero rewards to selected actions with $a_t \neq x_t$.

\subsection{Next Token Reasoning}
Treating each token emission as a distinct episode ensures that the reward depends only on the immediate state-action pair $(s_t, a_t)$, thereby preserving unbiased credit assignment.   
The framework is naturally compatible with architectures that perform iterative internal computation prior to token emission, including latent-reasoning models~\citep{zelikman2024quietstar} and loop transformers~\citep{dehghani2019universaltransformers,zhu2025ouro}.
Although each episode terminates at token emission, the state $s_t$ may encode the outcome of internal refinement cycles.
Our reward design can serve as an uncertainty-aware learning signal that can be combined with adaptive computation policies to allocate additional internal processing steps in uncertain contexts. 
By explicitly shaping positive and negative token-level rewards within a single-step policy-gradient framework, we provide a general and controllable mechanism that natively supports reasoning-oriented architectures through principled reward design.
\section{Conclusion}

This study establishes a theoretical bridge between next-token prediction and RL by interpreting cross-entropy loss as a specific instance of policy gradient optimization.
To exploit this connection, we introduce a generalized pre-training objective that utilizes a reward-shaping strategy with positive scaling and rank-aware negative rewards. 
Our experiments across multiple architectures and scales reveal that regulating the diversity-precision trade-off during pre-training modulates token entropy. 
Our findings indicate that precision-focused strategies (e.g., global entropy reduction or tail-token suppression) yield superior scaling for the subsequent RL stage.
These insights provide a novel perspective on optimizing pre-training for long CoT reasoning, suggesting new directions for sophisticated reward shaping in LLM development.

{
\bibliographystyle{iclr2025_conference}
\bibliography{ref/Top, ref/reference}
}

\clearpage
\appendix
\section{Experiment Details for Pre-Training and Mid-Training}
\label{appendix:pretrain}

\subsection{Implementation Details}
For both the pre-training and mid-training phases, we employ the AdamW~\citep{loshchilov2017adamw} optimizer, implementing a weight decay of 0.1 and applying gradient clipping at 1.0. 
Throughout these stages, we utilize a warmup-stable-decay learning rate schedule with a global batch size of 16M.
During the stable pre-training stage, which encompasses 500B tokens, the learning rate warms up over 2000 steps before stabilizing at $3\times 10^{-4}$. 
Subsequently, we perform mid-training on an additional 100B tokens, gradually decaying the learning rate from $3\times 10^{-4}$ to $3\times 10^{-5}$. 
We set the maximum sequence length to 4096 during pre-training and extend it to 16384 for the mid-training stage.
To support long-context modeling during mid-training, we increase the base frequency of RoPE~\citep{su2024roformer} from $1e^4$ to $1e^6$.

\subsection{Model Architecture}
Building upon the Qwen3~\citep{yang2025qwen3} architectures, we perform our experiments utilizing both dense and MoE architectures. 
Notably, we adopt an auxiliary loss free approach~\citep{liu2024deepseekv3} for the training of the MoE models.
Detailed architecture settings are provided in \Cref{table:arch}, where $E$ denotes the total number of experts and $E_a$ denotes the number of active experts.

\begin{table}[h]
\caption{\textbf{Detailed architectures settings of dense and MoE models.}}
\label{table:arch}
\centering
\footnotesize
\setlength{\tabcolsep}{13.2pt} %
\begin{tabular}{cccccccccccc}
\toprule
Model & $n_\text{layer}$ & $d_\text{model}$ & $d_\text{ffn}$ & $d_\text{expert}$ &$n_\text{head}$ & $n_\text{kvhead}$ & $E$ & $E_a$ \\
\midrule
1B Dense & 28 & 1536 & 4608 & - & 16 & 4 & - & -  \\
4B Dense & 36 & 2560 & 9728 & - & 32 & 8 & - & -  \\
5B-A0.3B MoE & 12 & 1024 & - & 320 & 32 & 4 & 384 & 12 \\
10B-A0.5B MoE & 16 & 1536 & - & 320 & 32 & 4 & 384 & 12 \\
20B-A1B MoE & 24 & 1536 & - & 480 & 32 & 4 & 384 & 12 \\
\bottomrule
\end{tabular}
\end{table}

\subsection{Experiment Results}
We report comprehensive evaluation results to demonstrate performance progression throughout the training process.
\Crefrange{table:pt_1b_dense_beta}{table:pt_20b_moe_lambda} present the pre-training results across various models and different training tokens. 
Similarly, \Crefrange{table:mt_4b_dense_beta}{table:mt_20b_moe_lambda} summarize the performance metrics for the mid-training stage.

\section{Experiment Details for RL}
\label{appendix:rl}

\subsection{Implementation Details}
For RLVR on mathematical reasoning tasks, we employ the on-policy GRPO algorithm \citep{shao2024deepseekmath} without KL regularization. 
Following \cite{yu2025dapo}, we incorporate clip-higher and dynamic sampling strategies to stabilize training. 
The process is conducted in two stages: an initial 700 steps with a sequence length of 8K, followed by continued training at a sequence length of 16K. 
We maintain a batch size of 128 and a constant learning rate of $1 \times 10^{-6}$ for two stages. 
During training, we sample 16 outputs per prompt at a temperature of 1.0.

\subsection{Experiment Results}
We provide detailed evaluation results to illustrate performance trajectories during the RL process.
\Crefrange{table:rl_b0_l10_l20_4b_dense}{table:rl_b0_l10.1_l20_20b_moe} display the RL results across different models and training steps.

\newpage

\begin{table}[!tb]
\centering
\setlength\tabcolsep{5.1pt}
\scriptsize
\caption{\textbf{Pre-Training performance comparison across different $\beta$ based on 1B dense models. The highest scores at the final checkpoint across the different configurations are shown in bold.}}
\begin{tabular}{l|l|cccc|cccc|cccc}
\toprule
& Hyperparameters & \multicolumn{4}{c|}{$\beta=-0.25;\; \tilde{\lambda}=0;\; \hat{\lambda}=0$} & \multicolumn{4}{c|}{$\beta=0;\; \tilde{\lambda}=0;\; \hat{\lambda}=0$} & \multicolumn{4}{c}{$\beta=0.50;\; \tilde{\lambda}=0;\; \hat{\lambda}=0$} \\
\cmidrule(l{0.3em}r{0.3em}){3-6}\cmidrule(l{0.3em}r{0.3em}){7-10}\cmidrule(l{0.3em}r{0.3em}){11-14}
\cmidrule(l{0.3em}r{0.3em}){3-6}\cmidrule(l{0.3em}r{0.3em}){7-10}\cmidrule(l{0.3em}r{0.3em}){11-14}
& \# Pre-Trained Tokens & 125B & 250B & 375B & 500B & 125B & 250B & 375B & 500B & 125B & 250B & 375B & 500B \\
\midrule
\multirow{5}{*}{\makecell[l]{General \\ Knowledge}} & MMLU (Acc.) & 20.47 & 23.65 & 24.06 & 25.12 & 20.85 & 23.39 & 24.04 & \textbf{26.62} & 21.71 & 24.26 & 24.37 & 25.51 \\
& MMLU-Pro (Acc.) & 8.01 & 9.25 & 9.54 & \textbf{10.06} & 7.63 & 8.23 & 7.98 & 9.02 & 7.85 & 8.55 & 9.44 & 9.59 \\
& NaturalQuestions (EM) & 3.02 & 4.34 & 4.88 & 5.35 & 2.74 & 4.07 & 4.52 & \textbf{5.79} & 2.69 & 3.77 & 4.18 & 5.21 \\
& TriviaQA (EM) & 8.45 & 12.32 & 14.75 & 16.03 & 8.65 & 12.63 & 13.53 & 16.25 & 8.35 & 12.39 & 14.24 & \textbf{16.49} \\
\rowcolor{green!10} \cellcolor{white} & Average & 9.99 & 12.39 & 13.31 & 14.14 & 9.97 & 12.08 & 12.52 & \textbf{14.42} & 10.15 & 12.24 & 13.06 & 14.20 \\
\midrule
\multirow{8}{*}{\makecell[l]{Commonsense \\ Reasoning}} & Hellaswag (Acc.) & 38.24 & 44.17 & 46.17 & 47.49 & 38.96 & 44.42 & 46.26 & \textbf{48.33} & 38.88 & 43.95 & 46.83 & 48.06 \\
\bottomrule
\end{tabular}
\label{table:pt_1b_dense_beta}
\end{table}

\begin{table}[!tb]
\centering
\setlength\tabcolsep{5.1pt}
\scriptsize
\caption{\textbf{Pre-Training performance comparison across different $\tilde{\lambda}$ and $\hat{\lambda}$ based on 1B dense models. The highest scores at the final checkpoint across the different configurations are shown in bold.}}
\begin{tabular}{l|l|cccc|cccc|cccc}
\toprule
& Hyperparameters & \multicolumn{4}{c|}{$\beta=0;\; \tilde{\lambda}=0;\; \hat{\lambda}=-0.1$} & \multicolumn{4}{c|}{$\beta=0;\; \tilde{\lambda}=0;\; \hat{\lambda}=0$} & \multicolumn{4}{c}{$\beta=0;\; \tilde{\lambda}=0.1;\; \hat{\lambda}=0$} \\
\cmidrule(l{0.3em}r{0.3em}){3-6}\cmidrule(l{0.3em}r{0.3em}){7-10}\cmidrule(l{0.3em}r{0.3em}){11-14}
\cmidrule(l{0.3em}r{0.3em}){3-6}\cmidrule(l{0.3em}r{0.3em}){7-10}\cmidrule(l{0.3em}r{0.3em}){11-14}
& \# Pre-Trained Tokens & 125B & 250B & 375B & 500B & 125B & 250B & 375B & 500B & 125B & 250B & 375B & 500B \\
\midrule
\multirow{5}{*}{\makecell[l]{General \\ Knowledge}} & MMLU (Acc.) & 21.61 & 23.71 & 24.04 & 26.06 & 20.85 & 23.39 & 24.04 & \textbf{26.62} & 23.59 & 22.92 & 24.63 & 25.89 \\
& MMLU-Pro (Acc.) & 8.34 & 9.32 & 8.42 & 8.67 & 7.63 & 8.23 & 7.98 & 9.02 & 9.04 & 8.69 & 8.50 & \textbf{9.23} \\
& NaturalQuestions (EM) & 2.66 & 4.71 & 4.90 & 5.21 & 2.74 & 4.07 & 4.52 & \textbf{5.79} & 3.05 & 4.79 & 4.74 & 5.15 \\
& TriviaQA (EM) & 8.61 & 12.70 & 14.95 & 15.59 & 8.65 & 12.63 & 13.53 & \textbf{16.25} & 8.76 & 11.18 & 14.08 & 15.50 \\
\rowcolor{green!10} \cellcolor{white} & Average & 10.31 & 12.61 & 13.08 & 13.88 & 9.97 & 12.08 & 12.52 & \textbf{14.42} & 11.11 & 11.90 & 12.99 & 13.94 \\
\midrule
\multirow{8}{*}{\makecell[l]{Commonsense \\ Reasoning}} & Hellaswag (Acc.) & 38.61 & 44.06 & 46.40 & 48.17 & 38.96 & 44.42 & 46.26 & \textbf{48.33} & 38.87 & 43.72 & 46.67 & 48.26 \\
\bottomrule
\end{tabular}
\label{table:pt_1b_dense_lambda}
\end{table}

\begin{table}[!tb]
\centering
\setlength\tabcolsep{5.1pt}
\scriptsize
\caption{\textbf{Pre-Training performance comparison across different $\beta$ based on 4B dense models. The highest scores at the final checkpoint across the different configurations are shown in bold.}}
\begin{tabular}{l|l|cccc|cccc|cccc}
\toprule
& Hyperparameters & \multicolumn{4}{c|}{$\beta=-0.25;\; \tilde{\lambda}=0;\; \hat{\lambda}=0$} & \multicolumn{4}{c|}{$\beta=0;\; \tilde{\lambda}=0;\; \hat{\lambda}=0$} & \multicolumn{4}{c}{$\beta=0.50;\; \tilde{\lambda}=0;\; \hat{\lambda}=0$} \\
\cmidrule(l{0.3em}r{0.3em}){3-6}\cmidrule(l{0.3em}r{0.3em}){7-10}\cmidrule(l{0.3em}r{0.3em}){11-14}
\cmidrule(l{0.3em}r{0.3em}){3-6}\cmidrule(l{0.3em}r{0.3em}){7-10}\cmidrule(l{0.3em}r{0.3em}){11-14}
& \# Pre-Trained Tokens & 125B & 250B & 375B & 500B & 125B & 250B & 375B & 500B & 125B & 250B & 375B & 500B \\
\midrule
\multirow{5}{*}{\makecell[l]{General \\ Knowledge}} & MMLU (Acc.) & 24.98 & 30.94 & 34.38 & 36.38 & 25.76 & 29.91 & 33.61 & 36.17 & 24.45 & 30.14 & 32.84 & \textbf{36.55} \\
& MMLU-Pro (Acc.) & 8.81 & 11.11 & 11.99 & \textbf{13.47} & 9.25 & 10.90 & 11.71 & 12.23 & 8.48 & 9.93 & 10.96 & 12.63 \\
& NaturalQuestions (EM) & 6.87 & 9.11 & 10.89 & 12.58 & 6.65 & 9.75 & 10.89 & 12.22 & 6.32 & 8.92 & 10.28 & \textbf{11.69} \\
& TriviaQA (EM) & 17.26 & 25.14 & 30.39 & 33.39 & 17.91 & 25.94 & 27.80 & \textbf{33.83} & 17.19 & 24.26 & 30.10 & 32.90 \\
\rowcolor{green!10} \cellcolor{white} & Average & 14.48 & 19.08 & 21.91 & \textbf{23.96} & 14.89 & 19.13 & 21.00 & 23.61 & 14.11 & 18.31 & 21.05 & 23.44 \\
\midrule
\multirow{8}{*}{\makecell[l]{Commonsense \\ Reasoning}} & Hellaswag (Acc.) & 50.36 & 57.81 & 60.80 & \textbf{63.01} & 50.22 & 57.00 & 60.50 & 62.91 & 39.15 & 57.49 & 60.98 & 62.87 \\
\bottomrule
\end{tabular}
\label{table:pt_4b_dense_beta}
\end{table}

\begin{table}[!tb]
\centering
\setlength\tabcolsep{5.1pt}
\scriptsize
\caption{\textbf{Pre-Training performance comparison across different $\tilde{\lambda}$ and $\hat{\lambda}$ based on 4B dense models. The highest scores at the final checkpoint across the different configurations are shown in bold.}}
\begin{tabular}{l|l|cccc|cccc|cccc}
\toprule
& Hyperparameters & \multicolumn{4}{c|}{$\beta=0;\; \tilde{\lambda}=0;\; \hat{\lambda}=-0.1$} & \multicolumn{4}{c|}{$\beta=0;\; \tilde{\lambda}=0;\; \hat{\lambda}=0$} & \multicolumn{4}{c}{$\beta=0;\; \tilde{\lambda}=0.1;\; \hat{\lambda}=0$} \\
\cmidrule(l{0.3em}r{0.3em}){3-6}\cmidrule(l{0.3em}r{0.3em}){7-10}\cmidrule(l{0.3em}r{0.3em}){11-14}
\cmidrule(l{0.3em}r{0.3em}){3-6}\cmidrule(l{0.3em}r{0.3em}){7-10}\cmidrule(l{0.3em}r{0.3em}){11-14}
& \# Pre-Trained Tokens & 125B & 250B & 375B & 500B & 125B & 250B & 375B & 500B & 125B & 250B & 375B & 500B \\
\midrule
\multirow{5}{*}{\makecell[l]{General \\ Knowledge}} & MMLU (Acc.) & 23.79 & 30.57 & 33.71 & 34.80 & 25.76 & 29.91 & 33.61 & \textbf{36.17} & 23.02 & 31.66 & 33.80 & 35.43 \\
& MMLU-Pro (Acc.) & 9.57 & 10.37 & 11.50 & 11.83 & 9.25 & 10.90 & 11.71 & \textbf{12.23} & 8.44 & 11.37 & 11.87 & 12.20 \\
& NaturalQuestions (EM) & 6.15 & 9.92 & 11.41 & 11.77 & 5.26 & 9.34 & 10.86 & \textbf{12.35} & 6.65 & 9.75 & 10.89 & 12.22 \\
& TriviaQA (EM) & 17.19 & 26.53 & 31.13 & 33.59 & 17.91 & 25.94 & 27.80 & \textbf{33.83} & 16.93 & 25.99 & 30.57 & 33.29 \\
\rowcolor{green!10} \cellcolor{white} & Average & 14.18 & 19.35 & 21.94 & 23.00 & 14.89 & 19.13 & 21.00 & \textbf{23.61} & 13.41 & 19.59 & 21.78 & 23.32 \\
\midrule
\multirow{8}{*}{\makecell[l]{Commonsense \\ Reasoning}} & Hellaswag (Acc.) & 50.56 & 57.22 & 60.25 & 62.27 & 50.22 & 57.00 & 60.50 & \textbf{62.91} & 50.21 & 58.09 & 61.30 & 62.09 \\
\bottomrule
\end{tabular}
\label{table:pt_4b_dense_lambda}
\end{table}

\begin{table}[!tb]
\centering
\setlength\tabcolsep{5.1pt}
\scriptsize
\caption{\textbf{Pre-Training performance comparison across different $\beta$ based on 5B-A0.3B MoE models. The highest scores at the final checkpoint across the different configurations are shown in bold.}}
\begin{tabular}{l|l|cccc|cccc|cccc}
\toprule
& Hyperparameters & \multicolumn{4}{c|}{$\beta=-0.25;\; \tilde{\lambda}=0;\; \hat{\lambda}=0$} & \multicolumn{4}{c|}{$\beta=0;\; \tilde{\lambda}=0;\; \hat{\lambda}=0$} & \multicolumn{4}{c}{$\beta=0.50;\; \tilde{\lambda}=0;\; \hat{\lambda}=0$} \\
\cmidrule(l{0.3em}r{0.3em}){3-6}\cmidrule(l{0.3em}r{0.3em}){7-10}\cmidrule(l{0.3em}r{0.3em}){11-14}
\cmidrule(l{0.3em}r{0.3em}){3-6}\cmidrule(l{0.3em}r{0.3em}){7-10}\cmidrule(l{0.3em}r{0.3em}){11-14}
& \# Pre-Trained Tokens & 125B & 250B & 375B & 500B & 125B & 250B & 375B & 500B & 125B & 250B & 375B & 500B \\
\midrule
\multirow{5}{*}{\makecell[l]{General \\ Knowledge}} & MMLU (Acc.) & 22.70 & 27.21 & 29.36 & 29.77 & 23.10 & 25.12 & 29.17 & \textbf{31.19} & 24.13 & 26.99 & 28.40 & 30.34 \\
& MMLU-Pro (Acc.) & 8.80 & 8.56 & 9.42 & 11.16 & 9.47 & 8.89 & 10.06 & \textbf{11.87} & 8.62 & 9.83 & 9.15 & 10.36 \\
& NaturalQuestions (EM) & 5.21 & 7.45 & 8.45 & 10.06 & 5.15 & 7.51 & 8.73 & \textbf{10.42} & 5.76 & 7.48 & 8.78 & 10.17 \\
& TriviaQA (EM) & 15.05 & 22.23 & 25.51 & 28.08 & 14.93 & 22.03 & 25.27 & \textbf{28.25} & 13.75 & 21.27 & 25.96 & 28.00 \\
\rowcolor{green!10} \cellcolor{white} & Average & 12.94 & 16.36 & 18.19 & 19.77 & 13.16 & 15.89 & 18.31 & \textbf{20.43} & 13.07 & 16.39 & 18.07 & 19.72 \\
\midrule
\multirow{8}{*}{\makecell[l]{Commonsense \\ Reasoning}} & Hellaswag (Acc.) & 47.89 & 54.08 & 56.70 & 57.78 & 48.54 & 54.63 & 56.88 & 57.41 & 48.58 & 54.31 & 56.81 & \textbf{58.13} \\
\bottomrule
\end{tabular}
\label{table:pt_5b_moe_beta}
\end{table}

\begin{table}[!tb]
\centering
\setlength\tabcolsep{5.1pt}
\scriptsize
\caption{\textbf{Pre-Training performance comparison across different $\tilde{\lambda}$ and $\hat{\lambda}$ based on 5B-A0.3B MoE models. The highest scores at the final checkpoint across the different configurations are shown in bold.}}
\begin{tabular}{l|l|cccc|cccc|cccc}
\toprule
& Hyperparameters & \multicolumn{4}{c|}{$\beta=0;\; \tilde{\lambda}=0;\; \hat{\lambda}=-0.1$} & \multicolumn{4}{c|}{$\beta=0;\; \tilde{\lambda}=0;\; \hat{\lambda}=0$} & \multicolumn{4}{c}{$\beta=0;\; \tilde{\lambda}=0.1;\; \hat{\lambda}=0$} \\
\cmidrule(l{0.3em}r{0.3em}){3-6}\cmidrule(l{0.3em}r{0.3em}){7-10}\cmidrule(l{0.3em}r{0.3em}){11-14}
\cmidrule(l{0.3em}r{0.3em}){3-6}\cmidrule(l{0.3em}r{0.3em}){7-10}\cmidrule(l{0.3em}r{0.3em}){11-14}
& \# Pre-Trained Tokens & 125B & 250B & 375B & 500B & 125B & 250B & 375B & 500B & 125B & 250B & 375B & 500B \\
\midrule
\multirow{5}{*}{\makecell[l]{General \\ Knowledge}} & MMLU (Acc.) & 22.24 & 26.11 & 26.41 & 29.99 & 23.10 & 25.12 & 29.17 & 31.19 & 22.20 & 26.45 & 29.44 & \textbf{31.37} \\
& MMLU-Pro (Acc.) & 7.55 & 8.74 & 8.82 & 10.36 & 9.47 & 8.89 & 10.06 & \textbf{11.87} & 8.18 & 9.63 & 10.11 & 11.02 \\
& NaturalQuestions (EM) & 5.60 & 7.34 & 7.37 & 8.84 & 5.15 & 7.51 & 8.73 & \textbf{10.42} & 5.21 & 7.95 & 8.31 & 9.81 \\
& TriviaQA (EM) & 14.54 & 21.20 & 25.70 & 27.26 & 14.93 & 22.03 & 25.27 & 28.25 & 15.09 & 21.84 & 25.85 & \textbf{28.91} \\
\rowcolor{green!10} \cellcolor{white} & Average & 12.48 & 15.85 & 17.08 & 19.11 & 13.16 & 15.89 & 18.31 & \textbf{20.43} & 12.67 & 16.47 & 18.43 & 20.28 \\
\midrule
\multirow{8}{*}{\makecell[l]{Commonsense \\ Reasoning}} & Hellaswag (Acc.) & 48.44 & 53.64 & 56.72 & \textbf{57.58} & 48.54 & 54.63 & 56.88 & 57.41 & 48.51 & 53.89 & 56.42 & 57.33 \\
\bottomrule
\end{tabular}
\label{table:pt_5b_moe_lambda}
\end{table}

\begin{table}[!tb]
\centering
\setlength\tabcolsep{5.1pt}
\scriptsize
\caption{\textbf{Pre-Training performance comparison across different $\beta$ based on 10B-A0.5B MoE models. The highest scores at the final checkpoint across the different configurations are shown in bold.}}
\begin{tabular}{l|l|cccc|cccc|cccc}
\toprule
& Hyperparameters & \multicolumn{4}{c|}{$\beta=-0.25;\; \tilde{\lambda}=0;\; \hat{\lambda}=0$} & \multicolumn{4}{c|}{$\beta=0;\; \tilde{\lambda}=0;\; \hat{\lambda}=0$} & \multicolumn{4}{c}{$\beta=0.50;\; \tilde{\lambda}=0;\; \hat{\lambda}=0$} \\
\cmidrule(l{0.3em}r{0.3em}){3-6}\cmidrule(l{0.3em}r{0.3em}){7-10}\cmidrule(l{0.3em}r{0.3em}){11-14}
\cmidrule(l{0.3em}r{0.3em}){3-6}\cmidrule(l{0.3em}r{0.3em}){7-10}\cmidrule(l{0.3em}r{0.3em}){11-14}
& \# Pre-Trained Tokens & 125B & 250B & 375B & 500B & 125B & 250B & 375B & 500B & 125B & 250B & 375B & 500B \\
\midrule
\multirow{5}{*}{\makecell[l]{General \\ Knowledge}} & MMLU (Acc.) & 26.13 & 32.74 & 35.09 & \textbf{36.60} & 25.35 & 29.14 & 33.50 & 35.12 & 26.05 & 30.71 & 34.03 & 35.48 \\
& MMLU-Pro (Acc.) & 9.44 & 10.21 & 11.38 & \textbf{12.26} & 8.24 & 10.01 & 11.64 & 12.15 & 8.81 & 11.32 & 10.97 & 11.90 \\
& NaturalQuestions (EM) & 6.79 & 10.36 & 12.47 & \textbf{13.99} & 7.06 & 10.47 & 11.41 & 11.96 & 7.84 & 10.58 & 11.36 & 13.77 \\
& TriviaQA (EM) & 20.69 & 29.78 & 35.25 & 37.95 & 19.94 & 29.86 & 33.89 & 37.31 & 20.73 & 29.64 & 35.31 & \textbf{39.13} \\
\rowcolor{green!10} \cellcolor{white} & Average & 15.76 & 20.77 & 23.55 & \textbf{25.20} & 15.15 & 19.87 & 22.61 & 24.14 & 15.86 & 20.56 & 22.92 & 25.07 \\
\midrule
\multirow{8}{*}{\makecell[l]{Commonsense \\ Reasoning}} & Hellaswag (Acc.) & 52.86 & 59.47 & 61.65 & 63.26 & 53.00 & 59.27 & 62.00 & \textbf{63.71} & 53.09 & 59.17 & 61.96 & 63.48 \\
\bottomrule
\end{tabular}
\label{table:pt_10b_moe_beta}
\end{table}

\begin{table}[!tb]
\centering
\setlength\tabcolsep{5.1pt}
\scriptsize
\caption{\textbf{Pre-Training performance comparison across different $\tilde{\lambda}$ and $\hat{\lambda}$ based on 10B-A0.5B MoE models. The highest scores at the final checkpoint across the different configurations are shown in bold.}}
\begin{tabular}{l|l|cccc|cccc|cccc}
\toprule
& Hyperparameters & \multicolumn{4}{c|}{$\beta=0;\; \tilde{\lambda}=0;\; \hat{\lambda}=-0.1$} & \multicolumn{4}{c|}{$\beta=0;\; \tilde{\lambda}=0;\; \hat{\lambda}=0$} & \multicolumn{4}{c}{$\beta=0;\; \tilde{\lambda}=0.1;\; \hat{\lambda}=0$} \\
\cmidrule(l{0.3em}r{0.3em}){3-6}\cmidrule(l{0.3em}r{0.3em}){7-10}\cmidrule(l{0.3em}r{0.3em}){11-14}
\cmidrule(l{0.3em}r{0.3em}){3-6}\cmidrule(l{0.3em}r{0.3em}){7-10}\cmidrule(l{0.3em}r{0.3em}){11-14}
& \# Pre-Trained Tokens & 125B & 250B & 375B & 500B & 125B & 250B & 375B & 500B & 125B & 250B & 375B & 500B \\
\midrule
\multirow{5}{*}{\makecell[l]{General \\ Knowledge}} & MMLU (Acc.) & 25.99 & 31.87 & 35.24 & 36.13 & 25.35 & 29.14 & 33.50 & 35.12 & 25.38 & 32.48 & 34.47 & \textbf{36.75} \\
& MMLU-Pro (Acc.) & 9.40 & 11.10 & 12.17 & \textbf{13.19} & 8.24 & 10.01 & 11.64 & 12.15 & 8.96 & 11.19 & 12.64 & 12.34 \\
& NaturalQuestions (EM) & 7.51 & 11.22 & 11.94 & 12.60 & 7.06 & 10.47 & 11.41 & 11.96 & 7.06 & 10.44 & 11.25 & \textbf{13.52} \\
& TriviaQA (EM) & 20.55 & 29.22 & 33.57 & 37.30 & 19.94 & 29.86 & 33.89 & 37.31 & 21.12 & 29.48 & 33.93 & \textbf{37.61} \\
\rowcolor{green!10} \cellcolor{white} & Average & 15.86 & 20.85 & 23.23 & 24.81 & 15.15 & 19.87 & 22.61 & 24.14 & 15.63 & 20.90 & 23.07 & \textbf{25.06} \\
\midrule
\multirow{8}{*}{\makecell[l]{Commonsense \\ Reasoning}} & Hellaswag (Acc.) & 52.70 & 59.23 & 63.30 & 63.68 & 53.00 & 59.27 & 62.00 & \textbf{63.71} & 53.06 & 59.75 & 63.67 & 63.38 \\
\bottomrule
\end{tabular}
\label{table:pt_10b_moe_lambda}
\end{table}

\begin{table}[!tb]
\centering
\setlength\tabcolsep{5.1pt}
\scriptsize
\caption{\textbf{Pre-Training performance comparison across different $\beta$ based on 20B-A1B MoE models. The highest scores at the final checkpoint across the different configurations are shown in bold.}}
\begin{tabular}{l|l|cccc|cccc|cccc}
\toprule
& Hyperparameters & \multicolumn{4}{c|}{$\beta=-0.25;\; \tilde{\lambda}=0;\; \hat{\lambda}=0$} & \multicolumn{4}{c|}{$\beta=0;\; \tilde{\lambda}=0;\; \hat{\lambda}=0$} & \multicolumn{4}{c}{$\beta=0.50;\; \tilde{\lambda}=0;\; \hat{\lambda}=0$} \\
\cmidrule(l{0.3em}r{0.3em}){3-6}\cmidrule(l{0.3em}r{0.3em}){7-10}\cmidrule(l{0.3em}r{0.3em}){11-14}
\cmidrule(l{0.3em}r{0.3em}){3-6}\cmidrule(l{0.3em}r{0.3em}){7-10}\cmidrule(l{0.3em}r{0.3em}){11-14}
& \# Pre-Trained Tokens & 125B & 250B & 375B & 500B & 125B & 250B & 375B & 500B & 125B & 250B & 375B & 500B \\
\midrule
\multirow{5}{*}{\makecell[l]{General \\ Knowledge}} & MMLU (Acc.) & 28.19 & 35.81 & 39.35 & \textbf{41.67} & 29.77 & 35.09 & 39.40 & 41.53 & 27.33 & 35.44 & 38.75 & 40.78 \\
& MMLU-Pro (Acc.) & 9.11 & 11.98 & 13.45 & \textbf{16.70} & 10.41 & 12.15 & 14.46 & 15.32 & 9.40 & 13.21 & 13.64 & 14.94 \\
& NaturalQuestions (EM) & 8.83 & 13.57 & 15.56 & 18.39 & 10.00 & 14.60 & 16.18 & 17.15 & 9.78 & 14.32 & 16.01 & \textbf{18.61} \\
& TriviaQA (EM) & 26.81 & 37.23 & 43.54 & \textbf{47.66} & 27.31 & 38.02 & 44.07 & 46.75 & 27.37 & 38.54 & 43.38 & 46.17 \\
\rowcolor{green!10} \cellcolor{white} & Average & 18.24 & 24.65 & 27.98 & \textbf{31.11} & 19.37 & 24.97 & 28.53 & 30.19 & 18.47 & 25.38 & 27.95 & 30.13 \\
\midrule
\multirow{8}{*}{\makecell[l]{Commonsense \\ Reasoning}} & Hellaswag (Acc.) & 58.17 & 64.88 & 67.68 & 69.05 & 58.52 & 65.31 & 67.88 & 69.29 & 58.35 & 65.43 & 67.98 & \textbf{69.47} \\
\bottomrule
\end{tabular}
\label{table:pt_20b_moe_beta}
\end{table}

\begin{table}[!tb]
\centering
\setlength\tabcolsep{5.1pt}
\scriptsize
\caption{\textbf{Pre-Training performance comparison across different $\tilde{\lambda}$ and $\hat{\lambda}$ based on 20B-A1B MoE models. The highest scores at the final checkpoint across the different configurations are shown in bold.}}
\begin{tabular}{l|l|cccc|cccc|cccc}
\toprule
& Hyperparameters & \multicolumn{4}{c|}{$\beta=0;\; \tilde{\lambda}=0;\; \hat{\lambda}=-0.1$} & \multicolumn{4}{c|}{$\beta=0;\; \tilde{\lambda}=0;\; \hat{\lambda}=0$} & \multicolumn{4}{c}{$\beta=0;\; \tilde{\lambda}=0.1;\; \hat{\lambda}=0$} \\
\cmidrule(l{0.3em}r{0.3em}){3-6}\cmidrule(l{0.3em}r{0.3em}){7-10}\cmidrule(l{0.3em}r{0.3em}){11-14}
\cmidrule(l{0.3em}r{0.3em}){3-6}\cmidrule(l{0.3em}r{0.3em}){7-10}\cmidrule(l{0.3em}r{0.3em}){11-14}
& \# Pre-Trained Tokens & 125B & 250B & 375B & 500B & 125B & 250B & 375B & 500B & 125B & 250B & 375B & 500B \\
\midrule
\multirow{5}{*}{\makecell[l]{General \\ Knowledge}} & MMLU (Acc.) & 28.84 & 36.46 & 39.82 & \textbf{42.04} & 29.77 & 35.09 & 39.40 & 41.53 & 25.92 & 35.71 & 39.21 & 41.47 \\
& MMLU-Pro (Acc.) & 9.77 & 12.92 & 15.24 & \textbf{16.20} & 10.41 & 12.15 & 14.46 & 15.32 & 9.53 & 12.26 & 13.68 & 15.35 \\
& NaturalQuestions (EM) & 9.83 & 14.40 & 16.18 & 18.39 & 10.00 & 14.60 & 16.18 & 17.15 & 9.06 & 14.46 & 15.65 & \textbf{18.95} \\
& TriviaQA (EM) & 27.44 & 37.48 & 42.76 & 46.19 & 27.31 & 38.02 & 44.07 & 46.75 & 26.42 & 38.38 & 43.95 & \textbf{47.16} \\
\rowcolor{green!10} \cellcolor{white} & Average & 18.97 & 25.32 & 28.50 & 30.71 & 19.37 & 24.97 & 28.53 & 30.19 & 17.73 & 25.20 & 28.12 & \textbf{30.73} \\
\midrule
\multirow{8}{*}{\makecell[l]{Commonsense \\ Reasoning}} & Hellaswag (Acc.) & 57.75 & 64.03 & 67.51 & 69.16 & 58.18 & 64.73 & 68.13 & 69.10 & 58.52 & 65.31 & 67.88 & \textbf{69.29} \\
\bottomrule
\end{tabular}
\label{table:pt_20b_moe_lambda}
\end{table}

\begin{table}[!tb]
\centering
\setlength\tabcolsep{5.1pt}
\scriptsize
\caption{\textbf{Mid-Training performance comparison across different $\beta$ based on 4B dense models. The highest scores at the final checkpoint across the different configurations are shown in bold.}}
\begin{tabular}{l|l|cccc|cccc|cccc}
\toprule
& Hyperparameters & \multicolumn{4}{c|}{$\beta=-0.25;\; \tilde{\lambda}=0;\; \hat{\lambda}=0$} & \multicolumn{4}{c|}{$\beta=0;\; \tilde{\lambda}=0;\; \hat{\lambda}=0$} & \multicolumn{4}{c}{$\beta=0.50;\; \tilde{\lambda}=0;\; \hat{\lambda}=0$} \\
\cmidrule(l{0.3em}r{0.3em}){3-6}\cmidrule(l{0.3em}r{0.3em}){7-10}\cmidrule(l{0.3em}r{0.3em}){11-14}
\cmidrule(l{0.3em}r{0.3em}){3-6}\cmidrule(l{0.3em}r{0.3em}){7-10}\cmidrule(l{0.3em}r{0.3em}){11-14}
& \# Mid-Trained Tokens & 25B & 50B & 75B & 100B & 25B & 50B & 75B & 100B & 25B & 50B & 75B & 100B \\
\midrule
\multirow{5}{*}{\makecell[l]{General \\ Knowledge}} & MMLU (Acc.) & 37.99 & 39.51 & 39.69 & 39.92 & 37.27 & 39.34 & 39.23 & \textbf{40.47} & 36.53 & 38.45 & 39.00 & 39.48 \\
& MMLU-Pro (Acc.) & 16.51 & 18.44 & 19.69 & \textbf{19.70} & 15.50 & 17.27 & 17.77 & 18.68 & 14.97 & 16.44 & 17.99 & 18.53 \\
& NaturalQuestions (EM) & 11.69 & 11.75 & 12.16 & 12.33 & 12.16 & 11.91 & 12.33 & \textbf{12.55} & 11.75 & 11.14 & 12.08 & 11.88 \\
& TriviaQA (EM) & 32.60 & 32.75 & 34.05 & 34.50 & 31.90 & 32.54 & 33.50 & 34.17 & 32.55 & 32.78 & 34.09 & \textbf{34.67} \\
\rowcolor{green!10} \cellcolor{white} & Average & 24.70 & 25.61 & 26.40 & \textbf{26.61} & 24.21 & 25.27 & 25.71 & 26.47 & 23.95 & 24.70 & 25.79 & 26.14 \\
\midrule
\multirow{8}{*}{\makecell[l]{Commonsense \\ Reasoning}} & Hellaswag (Acc.) & 61.59 & 61.70 & 61.95 & 62.41 & 61.25 & 61.72 & 62.14 & 62.09 & 61.68 & 62.01 & 62.54 & \textbf{62.51} \\
\bottomrule
\end{tabular}
\label{table:mt_4b_dense_beta}
\end{table}

\begin{table}[!tb]
\centering
\setlength\tabcolsep{5.1pt}
\scriptsize
\caption{\textbf{Mid-Training performance comparison across different $\tilde{\lambda}$ and $\hat{\lambda}$ based on 4B dense models. The highest scores at the final checkpoint across the different configurations are shown in bold.}}
\begin{tabular}{l|l|cccc|cccc|cccc}
\toprule
& Hyperparameters & \multicolumn{4}{c|}{$\beta=0;\; \tilde{\lambda}=0;\; \hat{\lambda}=-0.1$} & \multicolumn{4}{c|}{$\beta=0;\; \tilde{\lambda}=0;\; \hat{\lambda}=0$} & \multicolumn{4}{c}{$\beta=0;\; \tilde{\lambda}=0.1;\; \hat{\lambda}=0$} \\
\cmidrule(l{0.3em}r{0.3em}){3-6}\cmidrule(l{0.3em}r{0.3em}){7-10}\cmidrule(l{0.3em}r{0.3em}){11-14}
\cmidrule(l{0.3em}r{0.3em}){3-6}\cmidrule(l{0.3em}r{0.3em}){7-10}\cmidrule(l{0.3em}r{0.3em}){11-14}
& \# Mid-Trained Tokens & 25B & 50B & 75B & 100B & 25B & 50B & 75B & 100B & 25B & 50B & 75B & 100B \\
\midrule
\multirow{5}{*}{\makecell[l]{General \\ Knowledge}} & MMLU (Acc.) & 37.33 & 38.73 & 38.83 & 39.13 & 37.27 & 39.34 & 39.23 & \textbf{40.47} & 37.15 & 38.06 & 38.76 & 39.43 \\
& MMLU-Pro (Acc.) & 14.17 & 16.61 & 17.57 & 18.23 & 15.50 & 17.27 & 17.77 & \textbf{18.68} & 14.17 & 15.58 & 16.41 & 17.50 \\
& NaturalQuestions (EM) & 11.25 & 11.99 & 12.30 & 12.66 & 12.16 & 11.91 & 12.33 & 12.55 & 11.14 & 12.49 & 12.58 & \textbf{12.74}  \\
& TriviaQA (EM) & 33.29 & 33.89 & 34.80 & \textbf{35.32} & 31.90 & 32.54 & 33.50 & 34.17 & 32.73 & 33.50 & 34.17 & 34.59 \\
\rowcolor{green!10} \cellcolor{white} & Average & 24.01 & 25.31 & 25.88 & 26.34 & 24.21 & 25.27 & 25.71 & \textbf{26.47} & 23.80 & 24.91 & 25.48 & 26.07 \\
\midrule
\multirow{8}{*}{\makecell[l]{Commonsense \\ Reasoning}} & Hellaswag (Acc.) & 61.11 & 61.25 & 61.76 & \textbf{62.29} & 61.25 & 61.72 & 62.14 & 62.09 & 61.59 & 61.46 & 61.98 & 62.16 \\
\bottomrule
\end{tabular}
\label{table:mt_4b_dense_lambda}
\end{table}

\begin{table}[!tb]
\centering
\setlength\tabcolsep{5.1pt}
\scriptsize
\caption{\textbf{Mid-Training performance comparison across different $\beta$ based on 10B-A0.5B MoE models. The highest scores at the final checkpoint across the different configurations are shown in bold.}}
\begin{tabular}{l|l|cccc|cccc|cccc}
\toprule
& Hyperparameters & \multicolumn{4}{c|}{$\beta=-0.25;\; \tilde{\lambda}=0;\; \hat{\lambda}=0$} & \multicolumn{4}{c|}{$\beta=0;\; \tilde{\lambda}=0;\; \hat{\lambda}=0$} & \multicolumn{4}{c}{$\beta=0.50;\; \tilde{\lambda}=0;\; \hat{\lambda}=0$} \\
\cmidrule(l{0.3em}r{0.3em}){3-6}\cmidrule(l{0.3em}r{0.3em}){7-10}\cmidrule(l{0.3em}r{0.3em}){11-14}
\cmidrule(l{0.3em}r{0.3em}){3-6}\cmidrule(l{0.3em}r{0.3em}){7-10}\cmidrule(l{0.3em}r{0.3em}){11-14}
& \# Mid-Trained Tokens & 25B & 50B & 75B & 100B & 25B & 50B & 75B & 100B & 25B & 50B & 75B & 100B \\
\midrule
\multirow{5}{*}{\makecell[l]{General \\ Knowledge}} & MMLU (Acc.) & 36.39 & 37.23 & 37.75 & \textbf{38.11} & 36.15 & 37.27 & 37.54 & 37.82 & 35.04 & 36.49 & 36.45 & 37.10 \\
& MMLU-Pro (Acc.) & 12.31 & 12.88 & 13.69 & 13.91 & 12.98 & 13.91 & 14.41 & \textbf{14.71} & 12.03 & 12.82 & 13.32 & 13.46 \\
& NaturalQuestions (EM) & 12.80 & 13.77 & 14.18 & \textbf{14.35} & 11.30 & 12.11 & 12.44 & 12.85 & 11.30 & 12.11 & 12.44 & 12.85 \\
& TriviaQA (EM) & 36.61 & 37.37 & 38.29 & 38.76 & 35.84 & 37.15 & 37.81 & 38.38 & 37.57 & 38.84 & 39.62 & \textbf{40.21} \\
\rowcolor{green!10} \cellcolor{white} & Average & 24.53 & 25.31 & 25.98 & \textbf{26.28} & 24.07 & 25.11 & 25.55 & 25.94 & 24.40 & 25.41 & 25.87 & 26.18 \\
\midrule
\multirow{8}{*}{\makecell[l]{Commonsense \\ Reasoning}} & Hellaswag (Acc.) & 62.25 & 62.39 & 62.94 & 63.25 & 62.47 & 62.82 & 63.50 & \textbf{63.74} & 62.51 & 62.67 & 63.29 & 63.43 \\
\bottomrule
\end{tabular}
\label{table:mt_10b_moe_beta}
\end{table}

\begin{table}[!tb]
\centering
\setlength\tabcolsep{5.1pt}
\scriptsize
\caption{\textbf{Mid-Training performance comparison across different $\tilde{\lambda}$ and $\hat{\lambda}$ based on 10B-A0.5B MoE models. The highest scores at the final checkpoint across the different configurations are shown in bold.}}
\begin{tabular}{l|l|cccc|cccc|cccc}
\toprule
& Hyperparameters & \multicolumn{4}{c|}{$\beta=0;\; \tilde{\lambda}=0;\; \hat{\lambda}=-0.1$} & \multicolumn{4}{c|}{$\beta=0;\; \tilde{\lambda}=0;\; \hat{\lambda}=0$} & \multicolumn{4}{c}{$\beta=0;\; \tilde{\lambda}=0.1;\; \hat{\lambda}=0$} \\
\cmidrule(l{0.3em}r{0.3em}){3-6}\cmidrule(l{0.3em}r{0.3em}){7-10}\cmidrule(l{0.3em}r{0.3em}){11-14}
\cmidrule(l{0.3em}r{0.3em}){3-6}\cmidrule(l{0.3em}r{0.3em}){7-10}\cmidrule(l{0.3em}r{0.3em}){11-14}
& \# Mid-Trained Tokens & 25B & 50B & 75B & 100B & 25B & 50B & 75B & 100B & 25B & 50B & 75B & 100B \\
\midrule
\multirow{5}{*}{\makecell[l]{General \\ Knowledge}} & MMLU (Acc.) & 36.86 & 38.22 & 38.98 & 39.07 & 36.15 & 37.27 & 37.54 & 37.82 & 36.90 & 37.96 & 38.29 & \textbf{39.30} \\
& MMLU-Pro (Acc.) & 14.10 & 15.20 & 15.93 & \textbf{16.39} & 12.98 & 13.91 & 14.41 & 14.71 & 13.02 & 15.05 & 14.67 & 14.72 \\
& NaturalQuestions (EM) & 12.35 & 13.05 & 13.63 & \textbf{13.60} & 11.30 & 12.11 & 12.44 & 12.85 & 12.52 & 12.63 & 13.30 & 13.49 \\
& TriviaQA (EM) & 35.51 & 37.14 & 37.49 & 38.25 & 35.84 & 37.15 & 37.81 & 38.38 & 36.25 & 36.98 & 38.01 & \textbf{38.44} \\
\rowcolor{green!10} \cellcolor{white} & Average & 24.71 & 25.90 & 26.51 & \textbf{26.83} & 24.07 & 25.11 & 25.55 & 25.94 & 24.67 & 25.66 & 26.07 & 26.49 \\
\midrule
\multirow{8}{*}{\makecell[l]{Commonsense \\ Reasoning}} & Hellaswag (Acc.) & 62.67 & 62.96 & 63.12 & 63.50 & 62.47 & 62.82 & 63.50 & \textbf{63.74} & 62.93 & 63.07 & 63.42 & 63.73 \\
\bottomrule
\end{tabular}
\label{table:mt_10b_moe_lambda}
\end{table}

\begin{table}[!tb]
\centering
\setlength\tabcolsep{5.1pt}
\scriptsize
\caption{\textbf{Mid-Training performance comparison across different $\beta$ based on 20B-A1B MoE models. The highest scores at the final checkpoint across the different configurations are shown in bold.}}
\begin{tabular}{l|l|cccc|cccc|cccc}
\toprule
& Hyperparameters & \multicolumn{4}{c|}{$\beta=-0.25;\; \tilde{\lambda}=0;\; \hat{\lambda}=0$} & \multicolumn{4}{c|}{$\beta=0;\; \tilde{\lambda}=0;\; \hat{\lambda}=0$} & \multicolumn{4}{c}{$\beta=0.50;\; \tilde{\lambda}=0;\; \hat{\lambda}=0$} \\
\cmidrule(l{0.3em}r{0.3em}){3-6}\cmidrule(l{0.3em}r{0.3em}){7-10}\cmidrule(l{0.3em}r{0.3em}){11-14}
\cmidrule(l{0.3em}r{0.3em}){3-6}\cmidrule(l{0.3em}r{0.3em}){7-10}\cmidrule(l{0.3em}r{0.3em}){11-14}
& \# Mid-Trained Tokens & 25B & 50B & 75B & 100B & 25B & 50B & 75B & 100B & 25B & 50B & 75B & 100B \\
\midrule
\multirow{5}{*}{\makecell[l]{General \\ Knowledge}} & MMLU (Acc.) & 43.19 & 44.69 & 45.71 & 46.25 & 43.40 & 45.56 & 46.02 & \textbf{46.55} & 41.97 & 43.58 & 44.97 & 45.42 \\
& MMLU-Pro (Acc.) & 19.97 & 23.11 & 25.02 & \textbf{25.66} & 21.63 & 23.73 & 25.38 & 25.64 & 19.34 & 22.32 & 24.54 & 24.48 \\
& NaturalQuestions (EM) & 16.62 & 18.06 & 17.89 & \textbf{18.61} & 17.09 & 17.59 & 17.81 & 17.92 & 16.90 & 16.81 & 17.42 & 17.89 \\
& TriviaQA (EM) & 45.35 & 46.57 & 47.31 & 47.98 & 45.95 & 46.65 & 47.89 & \textbf{48.31} & 44.76 & 45.73 & 47.02 & 47.10 \\
\rowcolor{green!10} \cellcolor{white} & Average & 31.28 & 33.11 & 33.98 & \textbf{34.63} & 32.02 & 33.38 & 34.28 & 34.60 & 30.74 & 32.11 & 33.49 & 33.72 \\
\midrule
\multirow{8}{*}{\makecell[l]{Commonsense \\ Reasoning}} & Hellaswag (Acc.) & 67.88 & 67.61 & 68.01 & \textbf{68.61} & 67.92 & 68.10 & 68.45 & 68.49 & 67.87 & 68.12 & 68.33 & 68.50 \\
\bottomrule
\end{tabular}
\label{table:mt_20b_moe_beta}
\end{table}

\begin{table}[!tb]
\centering
\setlength\tabcolsep{5.1pt}
\scriptsize
\caption{\textbf{Mid-Training performance comparison across different $\tilde{\lambda}$ and $\hat{\lambda}$ based on 20B-A1B MoE models. The highest scores at the final checkpoint across the different configurations are shown in bold.}}
\begin{tabular}{l|l|cccc|cccc|cccc}
\toprule
& Hyperparameters & \multicolumn{4}{c|}{$\beta=0;\; \tilde{\lambda}=0;\; \hat{\lambda}=-0.1$} & \multicolumn{4}{c|}{$\beta=0;\; \tilde{\lambda}=0;\; \hat{\lambda}=0$} & \multicolumn{4}{c}{$\beta=0;\; \tilde{\lambda}=0.1;\; \hat{\lambda}=0$} \\
\cmidrule(l{0.3em}r{0.3em}){3-6}\cmidrule(l{0.3em}r{0.3em}){7-10}\cmidrule(l{0.3em}r{0.3em}){11-14}
\cmidrule(l{0.3em}r{0.3em}){3-6}\cmidrule(l{0.3em}r{0.3em}){7-10}\cmidrule(l{0.3em}r{0.3em}){11-14}
& \# Mid-Trained Tokens & 25B & 50B & 75B & 100B & 25B & 50B & 75B & 100B & 25B & 50B & 75B & 100B \\
\midrule
\multirow{5}{*}{\makecell[l]{General \\ Knowledge}} & MMLU (Acc.) & 42.79 & 44.69 & 45.86 & \textbf{46.77} & 43.40 & 45.56 & 46.02 & 46.55 & 42.56 & 44.47 & 46.10 & 45.61 \\
& MMLU-Pro (Acc.) & 22.57 & 24.29 & 25.91 & 25.83 & 21.63 & 23.73 & 25.38 & 25.64 & 21.57 & 23.55 & 26.33 & \textbf{26.15} \\
& NaturalQuestions (EM) & 16.51 & 16.54 & 17.73 & 18.01 & 17.09 & 17.59 & 17.81 & 17.92 & 16.98 & 18.14 & 18.53 & \textbf{18.92} \\
& TriviaQA (EM) & 45.43 & 45.85 & 47.29 & 47.82 & 45.95 & 46.65 & 47.89 & 48.31 & 45.97 & 46.90 & 47.72 & \textbf{48.35} \\
\rowcolor{green!10} \cellcolor{white} & Average & 31.82 & 32.84 & 34.20 & 34.61 & 32.02 & 33.38 & 34.28 & 34.60 & 31.77 & 33.27 & 34.67 & \textbf{34.76} \\
\midrule
\multirow{8}{*}{\makecell[l]{Commonsense \\ Reasoning}} & Hellaswag (Acc.) & 67.67 & 68.00 & 68.43 & 68.46 & 67.92 & 68.10 & 68.45 & 68.49 & 68.12 & 67.73 & 68.49 & \textbf{68.55} \\
\bottomrule
\end{tabular}
\label{table:mt_20b_moe_lambda}
\end{table}

\newpage

\begin{table}[!tb]
\centering
\renewcommand{\arraystretch}{0.96}
\setlength\tabcolsep{9pt}
\scriptsize
\caption{\textbf{RL performance of the 4B Dense Model with $\beta=0;\; \tilde{\lambda}=0;\; \hat{\lambda}=0$.}}
\begin{tabular}{l|l|cccccccccc}
\toprule
& \# RL Steps & 100 & 200 & 300 & 400 & 500 & 600 & 700 & 800 & 900 & 1000 \\
\midrule
\multirow{3}{*}{\makecell[l]{AIME24}} & Avg@128 & 0.89 & 0.81 & 1.14 & 1.04 & 1.12 & 1.54 & 1.33 & 1.69 & 1.64 & 1.62 \\
& Cons@128 & 0.00 & 0.00 & 0.00 & 0.00 & 3.33 & 3.33 & 3.33 & 6.67 & 3.33 & 3.33 \\
& Pass@64 & 15.34 & 13.77 & 21.27 & 14.80 & 11.68 & 22.86 & 19.81 & 19.68 & 14.78 & 18.32 \\
\midrule
\multirow{3}{*}{\makecell[l]{AIME25}} & Avg@128 & 0.57 & 0.86 & 0.86 & 1.46 & 1.59 & 2.29 & 1.30 & 2.06 & 2.21 & 2.24 \\
& Cons@128 & 0.00 & 0.00 & 0.00 & 3.33 & 0.00 & 6.67 & 3.33 & 3.33 & 3.33 & 3.33 \\
& Pass@64 & 14.90 & 11.67 & 10.84 & 10.00 & 16.16 & 10.84 & 12.93 & 14.35 & 10.84 & 14.59 \\
\midrule
\multirow{3}{*}{\makecell[l]{AMC23}} & Avg@128 & 18.46 & 23.07 & 24.94 & 27.87 & 28.96 & 28.73 & 27.17 & 28.18 & 29.16 & 29.51 \\
\bottomrule
\end{tabular}
\label{table:rl_b0_l10_l20_4b_dense}
\end{table}

\begin{table}[!tb]
\centering
\renewcommand{\arraystretch}{0.96}
\setlength\tabcolsep{9pt}
\scriptsize
\caption{\textbf{RL performance of the 4B Dense Model with $\beta=-0.25;\; \tilde{\lambda}=0;\; \hat{\lambda}=0$.}}
\begin{tabular}{l|l|cccccccccc}
\toprule
& \# RL Steps & 100 & 200 & 300 & 400 & 500 & 600 & 700 & 800 & 900 & 1000 \\
\midrule
\multirow{3}{*}{\makecell[l]{AIME24}} & Avg@128 & 0.76 & 0.96 & 1.02 & 0.55 & 1.09 & 1.43 & 1.82 & 2.03 & 2.47 & 2.57 \\
& Cons@128 & 0.00 & 3.33 & 3.33 & 0.00 & 0.00 & 3.33 & 3.33 & 6.67 & 6.67 & 6.67 \\
& Pass@64 & 14.17 & 17.11 & 13.31 & 12.10 & 17.09 & 19.99 & 20.64 & 19.17 & 24.12 & 20.34 \\
\midrule
\multirow{3}{*}{\makecell[l]{AIME25}} & Avg@128 & 0.83 & 1.46 & 2.11 & 1.72 & 2.27 & 2.11 & 1.98 & 2.19 & 2.40 & 2.63 \\
& Cons@128 & 0.00 & 0.00 & 3.33 & 3.33 & 3.33 & 3.33 & 3.33 & 3.33 & 6.67 & 3.33 \\
& Pass@64 & 17.46 & 13.33 & 21.90 & 19.12 & 19.99 & 15.64 & 14.15 & 16.51 & 11.67 & 15.82 \\
\midrule
\multirow{3}{*}{\makecell[l]{AMC23}} & Avg@128 & 19.55 & 25.47 & 29.08 & 30.96 & 31.70 & 31.72 & 32.38 & 32.85 & 33.01 & 33.09 \\
\bottomrule
\end{tabular}
\label{table:rl_b-0.25_l10_l20_4b_dense}
\end{table}

\begin{table}[!tb]
\centering
\renewcommand{\arraystretch}{0.96}
\setlength\tabcolsep{9pt}
\scriptsize
\caption{\textbf{RL performance of the 4B Dense Model with $\beta=0.50;\; \tilde{\lambda}=0;\; \hat{\lambda}=0$.}}
\begin{tabular}{l|l|cccccccccc}
\toprule
& \# RL Steps & 100 & 200 & 300 & 400 & 500 & 600 & 700 & 800 & 900 & 1000 \\
\midrule
\multirow{3}{*}{\makecell[l]{AIME24}} & Avg@128 & 0.94 & 1.38 & 1.43 & 1.46 & 2.29 & 1.95 & 1.74 & 2.19 & 2.53 & 2.42 \\
& Cons@128 & 3.33 & 3.33 & 3.33 & 3.33 & 6.67 & 3.33 & 3.33 & 3.33 & 3.33 & 3.33 \\
& Pass@64 & 18.77 & 23.78 & 16.58 & 20.87 & 23.65 & 19.99 & 18.98 & 24.56 & 23.62 & 22.08 \\
\midrule
\multirow{3}{*}{\makecell[l]{AIME25}} & Avg@128 & 0.83 & 1.88 & 1.95 & 2.03 & 3.39 & 3.57 & 2.76 & 3.67 & 4.14 & 4.66 \\
& Cons@128 & 0.00 & 6.67 & 3.33 & 3.33 & 10.00 & 10.00 & 3.33 & 6.67 & 10.00 & 10.00 \\
& Pass@64 & 17.05 & 15.84 & 17.51 & 13.33 & 15.00 & 19.59 & 15.84 & 20.01 & 18.77 & 15.00 \\
\midrule
\multirow{3}{*}{\makecell[l]{AMC23}} & Avg@128 & 19.34 & 22.52 & 23.09 & 26.35 & 26.60 & 25.46 & 28.28 & 29.45 & 29.49 & 30.10 \\
\bottomrule
\end{tabular}
\label{table:rl_b0.50_l10_l20_4b_dense}
\end{table}

\begin{table}[!tb]
\centering
\renewcommand{\arraystretch}{0.96}
\setlength\tabcolsep{9pt}
\scriptsize
\caption{\textbf{RL performance of the 4B Dense Model with $\beta=0;\; \tilde{\lambda}=0;\; \hat{\lambda}=-0.1$.}}
\begin{tabular}{l|l|cccccccccc}
\toprule
& \# RL Steps & 100 & 200 & 300 & 400 & 500 & 600 & 700 & 800 & 900 & 1000 \\
\midrule
\multirow{3}{*}{\makecell[l]{AIME24}} & Avg@128 & 0.70 & 0.70 & 0.76 & 0.89 & 1.07 & 0.81 & 0.86 & 1.15 & 1.38 & 1.46 \\
& Cons@128 & 0.00 & 3.33 & 3.33 & 3.33 & 3.33 & 3.33 & 3.33 & 3.33 & 3.33 & 3.33\\
& Pass@64 & 22.51 & 12.51 & 15.01 & 14.19 & 13.77 & 15.01 & 20.87 & 21.03 & 22.93 & 23.70 \\
\midrule
\multirow{3}{*}{\makecell[l]{AIME25}} & Avg@128 & 0.76 & 1.09 & 1.46 & 1.90 & 1.67 & 2.32 & 2.11 & 3.02 & 3.72 & 3.18 \\
& Cons@128 & 0.00 & 0.00 & 0.00 & 3.33 & 3.33 & 6.67 & 3.33 & 3.33 & 6.67 & 3.33 \\
& Pass@64 & 13.97 & 15.85 & 17.50 & 19.59 & 18.29 & 16.68 & 15.00 & 21.90 & 23.58 & 18.33 \\
\midrule
\multirow{3}{*}{\makecell[l]{AMC23}} & Avg@128 & 18.50 & 20.76 & 21.84 & 24.67 & 21.82 & 25.35 & 27.55 & 28.73 & 30.47 & 31.13 \\
\bottomrule
\end{tabular}
\label{table:rl_b0_l10_l2-0.1_4b_dense}
\end{table}

\begin{table}[!tb]
\centering
\renewcommand{\arraystretch}{0.96}
\setlength\tabcolsep{9pt}
\scriptsize
\caption{\textbf{RL performance of the 4B Dense Model with $\beta=0;\; \tilde{\lambda}=0.1;\; \hat{\lambda}=0$.}}
\begin{tabular}{l|l|cccccccccc}
\toprule
& \# RL Steps & 100 & 200 & 300 & 400 & 500 & 600 & 700 & 800 & 900 & 1000 \\
\midrule
\multirow{3}{*}{\makecell[l]{AIME24}} & Avg@128 & 0.49 & 0.55 & 0.91 & 0.73 & 0.68 & 0.63 & 0.68 & 0.76 & 0.91 & 1.28 \\
& Cons@128 & 0.00 & 0.00 & 0.00 & 0.00 & 0.00 & 0.00 & 0.00 & 0.00 & 0.00 & 3.33 \\
& Pass@64 & 9.17 & 13.24 & 23.37 & 8.13 & 18.97 & 17.10 & 20.65 & 17.94 & 16.37 & 20.32 \\
\midrule
\multirow{3}{*}{\makecell[l]{AIME25}} & Avg@128 & 0.68 & 0.83 & 1.35 & 1.46 & 1.28 & 1.25 & 1.51 & 1.93 & 2.29 & 2.29 \\
& Cons@128 & 0.00 & 0.00 & 3.33 & 3.33 & 3.33 & 3.33 & 3.33 & 3.33 & 3.33 & 6.67 \\
& Pass@64 & 21.79 & 22.09 & 17.83 & 20.01 & 16.61 & 15.81 & 17.48 & 15.84 & 16.26 & 15.00 \\
\midrule
\multirow{3}{*}{\makecell[l]{AMC23}} & Avg@128 & 18.24 & 20.59 & 24.16 & 25.16 & 23.89 & 28.38 & 29.20 & 29.57 & 31.43 & 32.68 \\
\bottomrule
\end{tabular}
\label{table:rl_b0_l10.1_l20_4b_dense}
\end{table}

\clearpage

\begin{table}[!tb]
\centering
\renewcommand{\arraystretch}{0.96}
\setlength\tabcolsep{9pt}
\scriptsize
\caption{\textbf{RL performance of the 10B-A0.5B MoE Model with $\beta=0;\; \tilde{\lambda}=0;\; \hat{\lambda}=0$.}}
\begin{tabular}{l|l|cccccccccc}
\toprule
& \# RL Steps & 100 & 200 & 300 & 400 & 500 & 600 & 700 & 800 & 900 & 1000 \\
\midrule
\multirow{3}{*}{\makecell[l]{AIME24}} & Avg@128 & 0.34 & 0.47 & 0.34 & 0.44 & 0.63 & 0.57 & 0.55 & 0.55 & 0.65 & 0.65 \\
& Cons@128 & 0.00 & 0.00 & 0.00 & 0.00 & 0.00 & 0.00 & 0.00 & 3.33 & 0.00 & 0.00 \\
& Pass@64 & 12.88 & 12.94 & 11.42 & 11.68 & 12.93 & 11.47 & 9.59 & 9.17 & 10.00 & 10.84 \\
\midrule
\multirow{3}{*}{\makecell[l]{AIME25}} & Avg@128 & 0.13 & 0.23 & 0.16 & 0.29 & 0.36 & 0.47 & 0.60 & 0.44 & 0.63 & 0.57 \\
& Cons@128 & 0.00 & 0.00 & 0.00 & 0.00 & 0.00 & 0.00 & 0.00 & 0.00 & 0.00 & 0.00 \\
& Pass@64 & 6.68 & 11.27 & 8.35 & 14.80 & 16.58 & 19.10 & 18.67 & 14.10 & 16.66 & 16.59 \\
\midrule
\multirow{3}{*}{\makecell[l]{AMC23}} & Avg@128 & 10.64 & 11.82 & 11.95 & 13.07 & 14.26 & 16.11 & 16.21 & 17.79 & 18.38 & 19.06 \\
\bottomrule
\end{tabular}
\label{table:rl_b0_l10_l20_10b_moe}
\end{table}

\begin{table}[!tb]
\centering
\renewcommand{\arraystretch}{0.96}
\setlength\tabcolsep{9pt}
\scriptsize
\caption{\textbf{RL performance of the 10B-A0.5B MoE Model with $\beta=-0.25;\; \tilde{\lambda}=0;\; \hat{\lambda}=0$.}}
\begin{tabular}{l|l|cccccccccc}
\toprule
& \# RL Steps & 100 & 200 & 300 & 400 & 500 & 600 & 700 & 800 & 900 & 1000 \\
\midrule
\multirow{3}{*}{\makecell[l]{AIME24}} & Avg@128 & 0.29 & 0.49 & 0.86 & 0.49 & 0.49 & 0.86 & 0.86 & 0.78 & 0.86 & 1.15 \\
& Cons@128 & 0.00 & 0.00 & 3.33 & 0.00 & 0.00 & 3.33 & 0.00 & 3.33 & 0.00 & 3.33 \\
& Pass@64 & 9.15 & 7.93 & 8.13 & 10.84 & 11.68 & 14.17 & 14.59 & 19.19 & 15.01 & 21.27 \\
\midrule
\multirow{3}{*}{\makecell[l]{AIME25}} & Avg@128 & 0.16 & 0.29 & 0.36 & 0.36 & 0.63 & 0.63 & 0.89 & 0.70 & 0.81 & 1.04 \\
& Cons@128 & 0.00 & 0.00 & 0.00 & 0.00 & 0.00 & 0.00 & 0.00 & 0.00 & 0.00 & 0.00 \\
& Pass@64 & 8.35 & 13.78 & 15.46 & 7.92 & 19.81 & 17.11 & 19.20 & 13.12 & 14.17 & 17.21 \\
\midrule
\multirow{3}{*}{\makecell[l]{AMC23}} & Avg@128 & 11.52 & 14.02 & 16.88 & 18.48 & 19.47 & 21.05 & 22.44 & 22.66 & 21.25 & 22.40 \\
\bottomrule
\end{tabular}
\label{table:rl_b-0.25_l10_l20_10b_moe}
\end{table}

\begin{table}[!tb]
\centering
\renewcommand{\arraystretch}{0.96}
\setlength\tabcolsep{9pt}
\scriptsize
\caption{\textbf{RL performance of the 10B-A0.5B MoE Model with $\beta=0.50;\; \tilde{\lambda}=0;\; \hat{\lambda}=0$.}}
\begin{tabular}{l|l|cccccccccc}
\toprule
& \# RL Steps & 100 & 200 & 300 & 400 & 500 & 600 & 700 & 800 & 900 & 1000 \\
\midrule
\multirow{3}{*}{\makecell[l]{AIME24}} & Avg@128 & 0.42 & 0.29 & 0.65 & 0.70 & 0.68 & 0.78 & 0.68 & 1.02 & 0.94 & 1.04 \\
& Cons@128 & 0.00 & 0.00 & 3.33 & 3.33 & 0.00 & 3.33 & 0.00 & 3.33 & 3.33 & 3.33 \\
& Pass@64 & 18.31 & 8.32 & 16.69 & 19.19 & 18.77 & 14.17 & 12.51 & 17.53 & 20.66 & 16.58 \\
\midrule
\multirow{3}{*}{\makecell[l]{AIME25}} & Avg@128 & 0.26 & 0.16 & 0.18 & 0.34 & 0.26 & 0.36 & 0.29 & 0.26 & 0.34 & 0.68 \\
& Cons@128 & 0.00 & 0.00 & 0.00 & 0.00 & 0.00 & 0.00 & 0.00 & 0.00 & 0.00 & 0.00 \\
& Pass@64 & 14.19 & 9.17 & 10.01 & 14.41 & 13.77 & 14.97 & 12.11 & 11.06 & 13.67 & 19.02 \\
\midrule
\multirow{3}{*}{\makecell[l]{AMC23}} & Avg@128 & 11.17 & 13.46 & 13.54 & 14.14 & 15.90 & 17.27 & 16.15 & 17.60 & 18.42 & 19.67 \\
\bottomrule
\end{tabular}
\label{table:rl_b0.50_l10_l20_10b_moe}
\end{table}

\begin{table}[!tb]
\centering
\renewcommand{\arraystretch}{0.96}
\setlength\tabcolsep{9pt}
\scriptsize
\caption{\textbf{RL performance of the 10B-A0.5B MoE Model with $\beta=0;\; \tilde{\lambda}=0;\; \hat{\lambda}=-0.1$.}}
\begin{tabular}{l|l|cccccccccc}
\toprule
& \# RL Steps & 100 & 200 & 300 & 400 & 500 & 600 & 700 & 800 & 900 & 1000 \\
\midrule
\multirow{3}{*}{\makecell[l]{AIME24}} & Avg@128 & 0.37 & 0.34 & 0.50 & 0.70 & 0.78 & 0.70 & 0.76 & 0.81 & 0.78 & 1.22 \\
& Cons@128 & 0.00 & 0.00 & 0.00 & 0.00 & 0.00 & 0.00 & 0.00 & 0.00 & 0.00 & 0.00 \\
& Pass@64 & 11.45 & 11.66 & 13.77 & 18.58 & 26.30 & 17.94 & 19.83 & 23.30 & 26.39 & 25.86 \\
\midrule
\multirow{3}{*}{\makecell[l]{AIME25}} & Avg@128 & 0.39 & 0.18 & 0.31 & 0.31 & 0.60 & 0.52 & 0.55 & 0.60 & 0.55 & 0.76 \\
& Cons@128 & 0.00 & 0.00 & 0.00 & 0.00 & 0.00 & 0.00 & 0.00 & 0.00 & 0.00 & 0.00 \\
& Pass@64 & 12.50 & 10.84 & 12.83 & 13.25 & 19.91 & 15.01 & 23.10 & 15.03 & 14.19 & 13.24 \\
\midrule
\multirow{3}{*}{\makecell[l]{AMC23}} & Avg@128 & 11.91 & 14.30 & 15.12 & 17.87 & 19.71 & 20.27 & 21.72 & 22.68 & 22.87 & 24.32 \\
\bottomrule
\end{tabular}
\label{table:rl_b0_l10_l2-0.1_10b_moe}
\end{table}

\begin{table}[!tb]
\centering
\renewcommand{\arraystretch}{0.96}
\setlength\tabcolsep{9pt}
\scriptsize
\caption{\textbf{RL performance of the 10B-A0.5B MoE Model with $\beta=0;\; \tilde{\lambda}=0.1;\; \hat{\lambda}=0$.}}
\begin{tabular}{l|l|cccccccccc}
\toprule
& \# RL Steps & 100 & 200 & 300 & 400 & 500 & 600 & 700 & 800 & 900 & 1000 \\
\midrule
\multirow{3}{*}{\makecell[l]{AIME24}} & Avg@128 & 0.32 & 0.50 & 0.65 & 0.86 & 0.83 & 0.96 & 0.81 & 0.68 & 0.83 & 0.99 \\
& Cons@128 & 0.00 & 0.00 & 0.00 & 0.00 & 0.00 & 0.00 & 0.00 & 0.00 & 0.00 & 0.00 \\
& Pass@64 & 9.57 & 10.84 & 9.17 & 21.66 & 10.84 & 13.96 & 16.48 & 10.85 & 13.67 & 20.77 \\
\midrule
\multirow{3}{*}{\makecell[l]{AIME25}} & Avg@128 & 0.21 & 0.26 & 0.34 & 0.44 & 0.42 & 0.76 & 0.83 & 0.99 & 1.17 & 1.02 \\
& Cons@128 & 0.00 & 0.00 & 0.00 & 0.00 & 0.00 & 0.00 & 0.00 & 0.00 & 3.33 & 3.33 \\
& Pass@64 & 8.97 & 10.65 & 14.62 & 16.81 & 11.14 & 18.23 & 18.34 & 16.27 & 17.94 & 17.51 \\
\midrule
\multirow{3}{*}{\makecell[l]{AMC23}} & Avg@128 & 12.30 & 14.57 & 17.58 & 19.16 & 19.90 & 20.88 & 20.55 & 21.35 & 20.63 & 19.86 \\
\bottomrule
\end{tabular}
\label{table:rl_b0_l10.1_l20_10b_moe}
\end{table}

\clearpage

\begin{table}[!tb]
\centering
\renewcommand{\arraystretch}{0.96}
\setlength\tabcolsep{9pt}
\scriptsize
\caption{\textbf{RL performance of the 20B-A1B MoE Model with $\beta=0;\; \tilde{\lambda}=0;\; \hat{\lambda}=0$.}}
\begin{tabular}{l|l|cccccccccc}
\toprule
& \# RL Steps & 100 & 200 & 300 & 400 & 500 & 600 & 700 & 800 & 900 & 1000 \\
\midrule
\multirow{3}{*}{\makecell[l]{AIME24}} & Avg@128 & 6.35 & 8.15 & 8.98 & 9.40 & 11.07 & 10.55 & 12.73 & 13.22 & 14.01 & 15.23 \\
& Cons@128 & 20.00 & 20.00 & 23.33 & 23.33 & 23.30 & 20.00 & 23.33 & 23.33 & 23.33 & 23.33 \\
& Pass@64 & 38.19 & 37.48 & 34.86 & 30.84 & 35.43 & 37.50 & 35.00 & 32.51 & 30.00 & 34.59 \\
\midrule
\multirow{3}{*}{\makecell[l]{AIME25}} & Avg@128 & 7.81 & 8.75 & 8.83 & 9.69 & 9.77 & 10.76 & 11.22 & 12.08 & 12.06 & 12.97 \\
& Cons@128 & 16.67 & 16.67 & 20.00 & 20.00 & 16.67 & 20.00 & 20.00 & 16.67 & 23.33 & 20.00 \\
& Pass@64 & 89.90 & 94.10 & 92.47 & 92.94 & 89.63 & 94.31 & 92.81 & 91.25 & 95.40 & 94.69 \\
\midrule
\multirow{3}{*}{\makecell[l]{AMC23}} & Avg@128 & 36.97 & 41.91 & 42.70 & 43.14 & 44.34 & 44.22 & 46.07 & 46.60 & 46.52 & 48.67 \\
\bottomrule
\end{tabular}
\label{table:rl_b0_l10_l20_20b_moe}
\end{table}

\begin{table}[!tb]
\centering
\renewcommand{\arraystretch}{0.96}
\setlength\tabcolsep{9pt}
\scriptsize
\caption{\textbf{RL performance of the 20B-A1B MoE Model with $\beta=-0.25;\; \tilde{\lambda}=0;\; \hat{\lambda}=0$.}}
\begin{tabular}{l|l|cccccccccc}
\toprule
& \# RL Steps & 100 & 200 & 300 & 400 & 500 & 600 & 700 & 800 & 900 & 1000 \\
\midrule
\multirow{3}{*}{\makecell[l]{AIME24}} & Avg@128 & 5.08 & 6.25 & 7.11 & 9.51 & 10.68 & 10.91 & 11.67 & 12.06 & 12.03 & 13.38 \\
& Cons@128 & 16.67 & 20.00 & 16.67 & 20.00 & 23.33 & 20.00 & 20.00 & 20.00 & 20.00 & 20.00 \\
& Pass@64 & 35.42 & 35.22 & 40.77 & 42.21 & 39.60 & 39.68 & 40.81 & 45.54 & 37.95 & 46.13 \\
\midrule
\multirow{3}{*}{\makecell[l]{AIME25}} & Avg@128 & 34.88 & 39.82 & 42.85 & 45.37 & 46.80 & 48.18 & 49.28 & 50.45 & 50.76 & 52.32 \\
& Cons@128 & 47.50 & 52.50 & 55.00 & 55.00 & 57.50 & 60.00 & 52.50 & 55.00 & 55.00 & 62.50 \\
& Pass@64 & 91.83 & 93.13 & 89.55 & 89.03 & 89.12 & 87.25 & 89.98 & 92.31 & 89.64 & 93.71 \\
\midrule
\multirow{3}{*}{\makecell[l]{AMC23}} & Avg@128 & 11.52 & 14.02 & 16.88 & 18.48 & 19.47 & 21.05 & 22.44 & 22.66 & 21.25 & 22.40 \\
\bottomrule
\end{tabular}
\label{table:rl_b-0.25_l10_l20_20b_moe}
\end{table}

\begin{table}[!tb]
\centering
\renewcommand{\arraystretch}{0.96}
\setlength\tabcolsep{9pt}
\scriptsize
\caption{\textbf{RL performance of the 20B-A1B MoE Model with $\beta=0.50;\; \tilde{\lambda}=0;\; \hat{\lambda}=0$.}}
\begin{tabular}{l|l|cccccccccc}
\toprule
& \# RL Steps & 100 & 200 & 300 & 400 & 500 & 600 & 700 & 800 & 900 & 1000 \\
\midrule
\multirow{3}{*}{\makecell[l]{AIME24}} & Avg@128 & 4.77 & 5.55 & 6.64 & 7.50 & 7.76 & 8.20 & 9.38 & 9.03 & 10.18 & 12.14 \\
& Cons@128 & 13.33 & 20.00 & 20.00 & 20.00 & 16.67 & 20.00 & 20.00 & 23.33 & 20.00 & 23.33 \\
& Pass@64 & 34.76 & 39.16 & 36.18 & 40.52 & 37.66 & 38.13 & 41.45 & 39.57 & 39.32 & 40.99 \\
\midrule
\multirow{3}{*}{\makecell[l]{AIME25}} & Avg@128 & 5.83 & 7.42 & 7.94 & 9.61 & 9.64 & 9.06 & 9.74 & 9.77 & 10.18 & 11.25 \\
& Cons@128 & 16.67 & 20.00 & 16.67 & 16.67 & 16.67 & 16.67 & 20.00 & 23.33 & 23.33 & 16.67 \\
& Pass@64 & 32.93 & 38.56 & 36.81 & 39.17 & 37.86 & 36.26 & 37.45 & 32.88 & 35.82 & 42.89 \\
\midrule
\multirow{3}{*}{\makecell[l]{AMC23}} & Avg@128 & 30.41 & 35.27 & 39.39 & 40.25 & 40.66 & 40.51 & 43.50 & 41.13 & 42.44 & 45.43 \\
\bottomrule
\end{tabular}
\label{table:rl_b0.50_l10_l20_20b_moe}
\end{table}

\begin{table}[!tb]
\centering
\renewcommand{\arraystretch}{0.96}
\setlength\tabcolsep{9pt}
\scriptsize
\caption{\textbf{RL performance of the 20B-A1B MoE Model with $\beta=0;\; \tilde{\lambda}=0;\; \hat{\lambda}=-0.1$.}}
\begin{tabular}{l|l|cccccccccc}
\toprule
& \# RL Steps & 100 & 200 & 300 & 400 & 500 & 600 & 700 & 800 & 900 & 1000 \\
\midrule
\multirow{3}{*}{\makecell[l]{AIME24}} & Avg@128 & 6.72 & 8.67 & 9.77 & 11.85 & 11.12 & 12.29 & 12.81 & 13.49 & 14.14 & 15.03 \\
& Cons@128 & 20.00 & 20.00 & 20.00 & 23.33 & 23.33 & 20.00 & 20.00 & 23.33 & 20.00 & 23.33 \\
& Pass@64 & 38.01 & 42.74 & 35.32 & 38.36 & 38.76 & 39.20 & 40.77 & 41.44 & 40.38 & 43.95 \\
\midrule
\multirow{3}{*}{\makecell[l]{AIME25}} & Avg@128 & 7.97 & 9.58 & 10.39 & 10.52 & 11.64 & 11.77 & 12.47 & 12.66 & 13.62 & 13.98 \\
& Cons@128 & 13.33 & 16.67 & 13.33 & 23.33 & 23.33 & 23.33 & 23.33 & 23.33 & 23.33 & 23.33 \\
& Pass@64 & 37.42 & 42.73 & 39.13 & 41.91 & 41.47 & 35.01 & 37.31 & 45.65 & 41.49 & 42.49 \\
\midrule
\multirow{3}{*}{\makecell[l]{AMC23}} & Avg@128 & 36.45 & 40.68 & 42.64 & 45.55 & 45.29 & 46.68 & 49.82 & 49.71 & 50.70 & 51.29 \\
\bottomrule
\end{tabular}
\label{table:rl_b0_l10_l2-0.1_20b_moe}
\end{table}

\begin{table}[!tb]
\centering
\renewcommand{\arraystretch}{0.96}
\setlength\tabcolsep{9pt}
\scriptsize
\caption{\textbf{RL performance of the 20B-A1B MoE Model with $\beta=0;\; \tilde{\lambda}=0.1;\; \hat{\lambda}=0$.}}
\begin{tabular}{l|l|cccccccccc}
\toprule
& \# RL Steps & 100 & 200 & 300 & 400 & 500 & 600 & 700 & 800 & 900 & 1000 \\
\midrule
\multirow{3}{*}{\makecell[l]{AIME24}} & Avg@128 & 5.76 & 6.77 & 8.33 & 9.27 & 9.11 & 9.89 & 10.21 & 10.13 & 11.07 & 11.93 \\
& Cons@128 & 13.33 & 13.33 & 23.33 & 20.00 & 20.00 & 23.33 & 23.33 & 20.00 & 20.00 & 23.33 \\
& Pass@64 & 37.39 & 32.44 & 33.13 & 35.41 & 31.26 & 37.10 & 30.85 & 33.15 & 31.27 & 30.43 \\
\midrule
\multirow{3}{*}{\makecell[l]{AIME25}} & Avg@128 & 8.02 & 9.09 & 9.77 & 11.28 & 11.20 & 11.35 & 11.07 & 11.82 & 12.06 & 12.19 \\
& Cons@128 & 13.33 & 16.67 & 16.67 & 23.33 & 16.67 & 16.67 & 16.67 & 20.00 & 16.67 & 20.00 \\
& Pass@64 & 37.39 & 38.82 & 39.48 & 39.60 & 40.46 & 34.03 & 39.78 & 34.90 & 39.92 & 38.34 \\
\midrule
\multirow{3}{*}{\makecell[l]{AMC23}} & Avg@128 & 37.64 & 41.50 & 42.64 & 45.20 & 46.04 & 47.40 & 45.72 & 46.13 & 47.93 & 48.59 \\
\bottomrule
\end{tabular}
\label{table:rl_b0_l10.1_l20_20b_moe}
\end{table}

\end{document}